\pdfoutput=1

\documentclass[11pt]{article}

\usepackage[preprint]{acl}
\usepackage{times}
\usepackage{latexsym}
\usepackage{float}
\usepackage{paracol} 
\usepackage{amsmath}
\usepackage{amssymb}
\usepackage{breqn}
\usepackage{wrapfig}
\usepackage{dblfloatfix}
\usepackage{cuted} 
\usepackage{qrcode} 
\usepackage{ai-usage-card}
\usepackage{annotation}

\usepackage{booktabs}
\usepackage{tablefootnote}
\usepackage{cleveref}
\usepackage[T1]{fontenc}

\usepackage[utf8]{inputenc}

\usepackage{microtype}
\usepackage{inconsolata}
\usepackage{picinpar}

\usepackage{graphicx}
\usepackage{multicol}
\usepackage{placeins}

\usepackage[nolist]{acronym}
\usepackage{caption}
\usepackage{subcaption}

\usepackage{enumitem}

\usepackage{colortbl}
\usepackage{xcolor} %
    \definecolor{UBlau}{HTML}{153268}
    \definecolor{LBlau}{HTML}{005f9b}
    \definecolor{LMBlau}{HTML}{0091c8}
    \definecolor{LHBlau}{HTML}{50a5d2}
    \definecolor{WiWi}{HTML}{2b7ab3}
    \definecolor{Grau60}{HTML}{878786}
\usepackage{pgfplots}
\usepackage{collcell}

\usepackage{soul}

\definecolor{lightred}{RGB}{255, 182, 193} %
\colorlet{transparentlightred}{lightred!50} %

\sethlcolor{transparentlightred}

\newcommand{\gradientcell}[3]{%
  \pgfmathsetmacro{\percent}{(#1-#2)/(#3-#2)*100} %
  \edef\tempcellcolor{\noexpand\cellcolor{green!\percent!red!30}}%
  \tempcellcolor #1%
}

\usepackage{tcolorbox} %
\newtcolorbox{combinedprompt}{
    colframe=black,
    colback=white,
    boxrule=0.6mm,
    width=\linewidth,
    fonttitle=\bfseries,
    rounded corners,
    coltitle=black
}

\usepackage{listings} %
\lstnewenvironment{configpython}[2][] %
{
    \lstset{
        commentstyle=\color{UBlau},
        keywordstyle=\color{WiWi},
        numberstyle=\tiny\color{Grau60},
        basicstyle=\ttfamily\footnotesize,
        language=python,
        numbers=none,
        frame=tblr,           %
        tabsize=4,
        captionpos=b,         %
        caption={#1},
        label={#2},
        breaklines=true,      %
        breakatwhitespace=false,
        breakautoindent=true,
        prebreak=\mbox{\textcolor{red}{$\hookrightarrow$}} %
    }
}{}

\lstnewenvironment{configjson}[2][] %
{
    \lstset{
        commentstyle=\color{UBlau},
        keywordstyle=\color{WiWi},
        numberstyle=\tiny\color{Grau60},
        basicstyle=\ttfamily\footnotesize,
        numbers=none,
        frame=tblr,           %
        tabsize=4,
        captionpos=b,         %
        caption={#1},
        label={#2},
        breaklines=true,      %
        breakatwhitespace=false,
        breakautoindent=true,
        prebreak=\mbox{\textcolor{red}{$\hookrightarrow$}} %
    }
}{}

\definecolor{systemcolor}{rgb}{0.9,0.9,1} %
\definecolor{usercolor}{rgb}{1,0.9,0.9} %

\usepackage{adjustbox} %

\newcommand{\dataset}{SPaRC} %

\tcbset{
  aibox/.style={
    width=474.18663pt,
    top=10pt,
    colback=white,
    colframe=black,
    colbacktitle=black,
    enhanced,
    center,
    attach boxed title to top left={yshift=-0.1in,xshift=0.15in},
    boxed title style={boxrule=0pt,colframe=white,},
  }
}
\newtcolorbox{AIbox}[2][]{aibox,title=#2,#1}

\title{\dataset{}: A Spatial Pathfinding Reasoning Challenge}

\author{
  Lars Benedikt Kaesberg$^*$, Jan Philip Wahle$^*$, Terry Ruas, Bela Gipp\\
  University of Göttingen, Germany\\
  $^*$\texttt{\{l.kaesberg, wahle\}@uni-goettingen.de}\\[1em] %
  \hspace*{0.15cm}
  \begin{tabular}{l@{\hskip 0.5em}l@{\hskip 1em}l}
    \adjustbox{valign=c}{\includegraphics[height=1em]{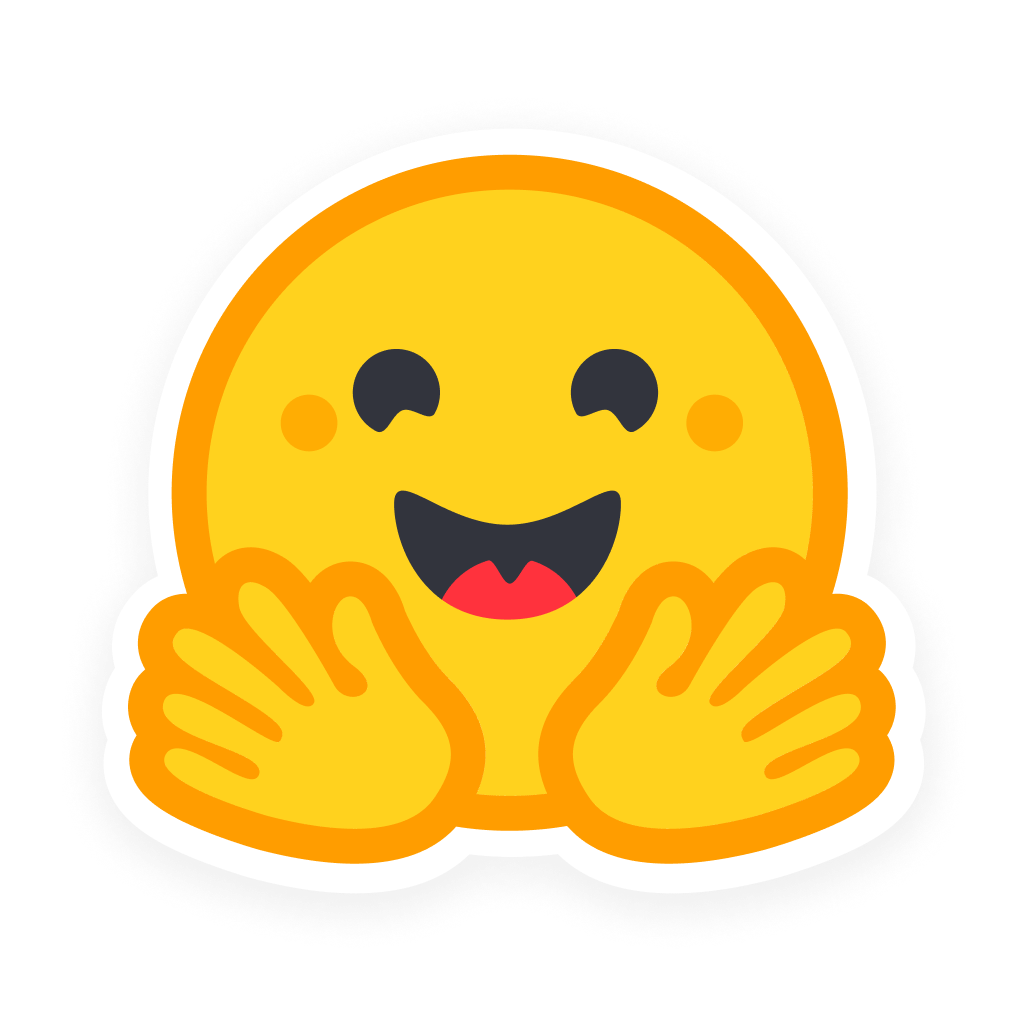}} & Dataset & \href{https://huggingface.co/datasets/lkaesberg/SPaRC}{hf.co/datasets/lkaesberg/SPaRC} \\
    \adjustbox{valign=c}{\includegraphics[height=1em]{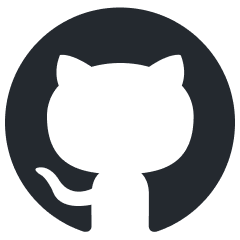}} & Code & \href{https://github.com/lkaesberg/SPaRC}{github.com/lkaesberg/SPaRC} \\
    \adjustbox{valign=c}{\includegraphics[height=1em]{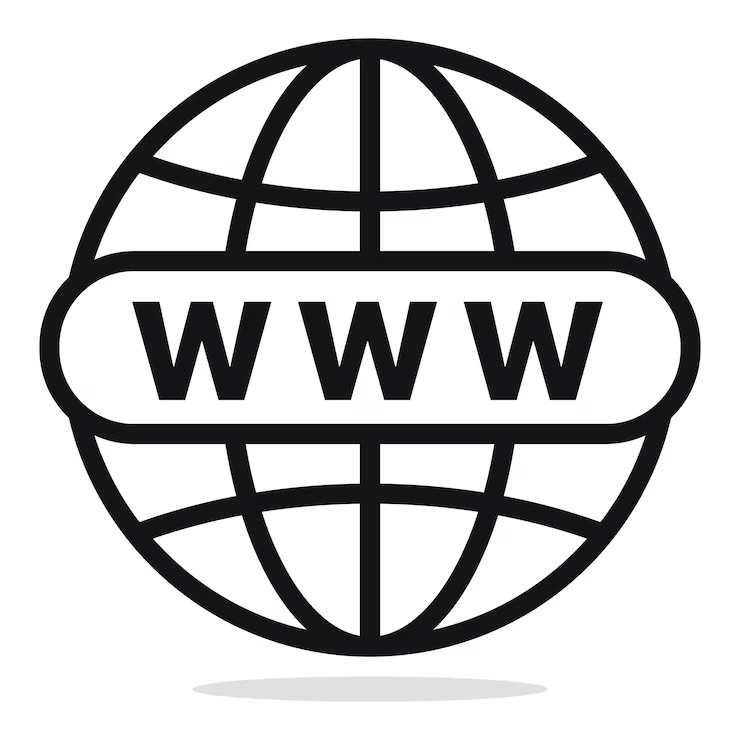}} & Webpage & \href{https://sparc.gipplab.org}{sparc.gipplab.org} \\
  \end{tabular}
}

\begin{acronym}
\end{acronym}

\begin{document}
\maketitle
\AddAnnotationRef
\begin{abstract}
Existing reasoning datasets saturate and fail to test abstract, multi-step problems, especially pathfinding and complex rule constraint satisfaction. We introduce \textbf{\dataset{}} (\textbf{S}patial \textbf{Pa}thfinding \textbf{R}easoning \textbf{C}hallenge), a dataset of 1,000 2D grid pathfinding puzzles to evaluate spatial and rule-based reasoning, requiring step-by-step planning with arithmetic and geometric rules. 
Humans achieve near-perfect accuracy (98.0\%; 94.5\% on hard puzzles), while the best reasoning models, such as o4-mini, struggle (15.8\%; 1.1\% on hard puzzles). 
Models often generate invalid paths (>50\% of puzzles for o4-mini), and reasoning tokens reveal they make errors in navigation and spatial logic. 
Unlike humans, who take longer on hard puzzles, models fail to scale test-time compute with difficulty.
Allowing models to make multiple solution attempts improves accuracy, suggesting potential for better spatial reasoning with improved training and efficient test-time scaling methods.
\dataset{} can be used as a window into models' spatial reasoning limitations and drive research toward new methods that excel in abstract, multi-step problem-solving.

\end{abstract}

\section{Introduction}
\label{sec:introduction}

Reasoning models made stark progress to solve complex mathematical \cite{hendrycks_measuring_2021}, software-engineering \cite{jimenez2024swebenchlanguagemodelsresolve,quan2025codeelobenchmarkingcompetitionlevelcode}, and knowledge tasks \cite{Hendrycks2020Measuring}.
With more capable models comes the question of how to measure their progress in reasoning, and how they compare to humans.
As reasoning benchmarks test specific tasks with priors (e.g., (MMLU-Pro \cite{wang_mmlu-pro_2024}, GPQA \cite{rein_gpqa_2023}), models started to achieve (super-) human scores, leading to rapid dataset saturation.
Thus, datasets probing abstract reasoning with minimal priors have become increasingly important as they are more robust to scaling training data and pattern-matching. 
Notably, ARC-AGI \cite{Chollet2019Measure} and related works \cite{song2025visualpuzzlesdecouplingmultimodalreasoning,wang2024pictureworththousandwords} challenge models with spatial few-shot grid problems.
However, they often do not require a combination of step-by-step planning, pathfinding, and logic skills; abilities most human possesses \cite{Chollet2019Measure}.
Spatial reasoning is an important component for solving tasks like navigation and manipulation in robotics, scene understanding in computer vision, and augmented reality.

\begin{figure}
    \centering
    \includegraphics[width=0.8\linewidth]{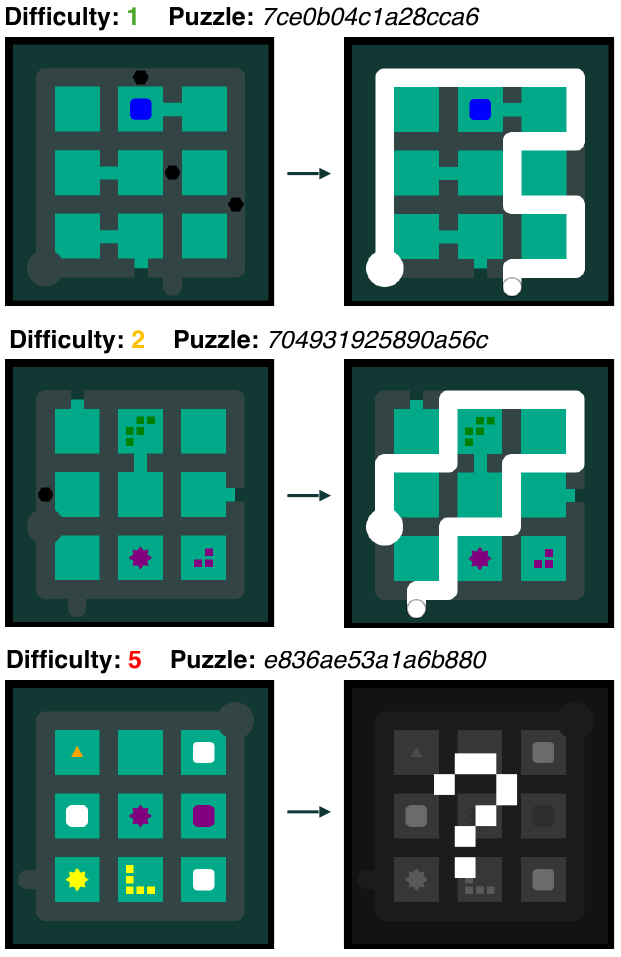}
    \caption{Example puzzles from \dataset{}.} %
    \label{fig:example-puzzle}
\end{figure}

We propose \dataset{}, a new dataset to overcome limitations of current datasets, primarily focusing on pathfinding and the combination of arithmetic and geometric rules, such as counting, segregation, and shape logic, by presenting multi-step constraint problems.
Our proposed task consists of 2D grid puzzles through which a line must be drawn from start to end while fulfilling various rules, such as collecting dots along the way or separating colored elements  (see \Cref{fig:example-puzzle} for an example).
These rules can be combined in various non-trivial ways and involve deep, abstract, rule-based reasoning within a constrained spatial pathfinding environment.
We use text-based grids rather than images to avoid testing perception and instead isolate pure spatial reasoning and planning ability.
Solving these puzzles requires an understanding of the individual rules and their connections, and long-term planning to meet all rules simultaneously. This often involves revising previous hypotheses where a single wrong step can irrevocably lead to the wrong path.
We provide 500 train and 500 test puzzles of different sizes and difficulty degrees from 1 (very easy) to 5 (very hard).

\begin{figure}
    \centering
    \includegraphics[width=\linewidth]{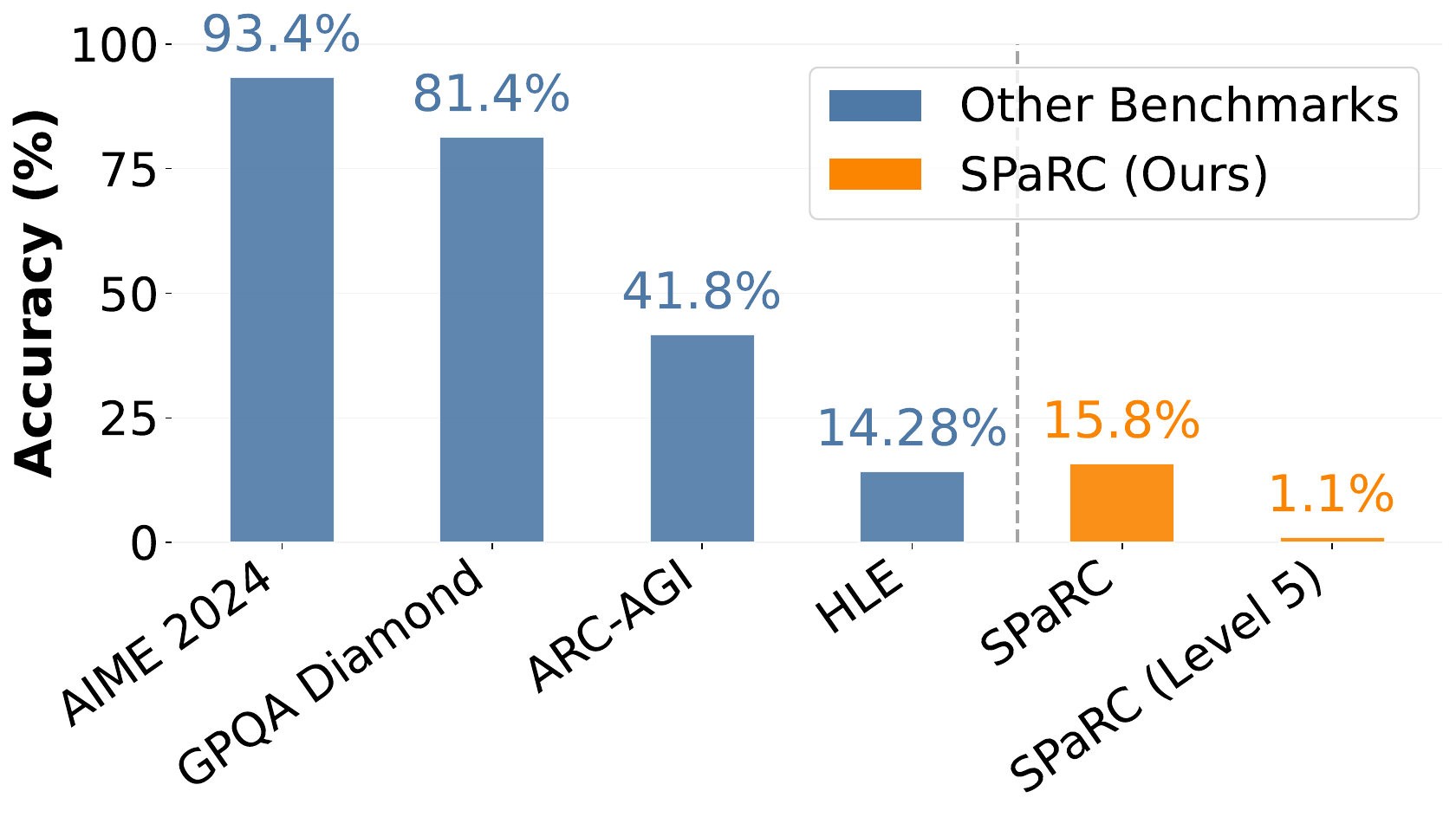}
    \caption{\textbf{Accuracy (\%)} of o4-mini on existing benchmarks and on \dataset{}, as well as on only hard puzzles from \dataset{} with difficulty 5.}
    \label{fig:dataset-comparison}
\end{figure}

Experiments with three instruction-tuned models, four reasoning models, and six human annotators show puzzles are solved easily by humans at 98\% accuracy (94.5\% for difficulty 5 puzzles) but challenge the best reasoning model, o4-mini, at 15.8\% accuracy (1.1\% for difficulty 5 puzzles).
\Cref{fig:dataset-comparison} compares the accuracy of the best reasoning model we tested (o4-mini) on existing reasoning benchmarks with our proposed \dataset{}, showing that it poses a new challenge for models.
Models often fail to generate valid paths, and reasoning tokens reveal issues with grid navigation, spatial logic, and careless mistakes that lead to irreversible errors.
Humans take up to 13 times longer on harder puzzles. 
Instruction-tuned models increase test-time tokens by $\sim40\%$, and reasoning models only by $\sim5\%$ with higher difficulty. Multiple attempts per puzzle raise accuracy (e.g., 15.8\% to 35.0\% for o4-mini), indicating inefficient solution-finding and potential for improved spatial reasoning training.
Ablations show prompt design (15.8\% to 21.0\%) and few-shot examples (12.6\% to 15.8\%) have modest effects, and multimodal prompting (i.e., puzzle screenshots) does not improve performance over text (12.6\% vs. 5.6\% for o4-mini).

\dataset{} provides a new challenge to evaluate spatial and rule-based reasoning in large language models (LLMs), addressing limitations of existing saturated benchmarks. 

\medskip
\noindent\textbf{Key Contributions:}
\begin{itemize}[leftmargin=*,label={\color{teal}$\blacktriangleright$},itemsep=0.15em]
    \item We propose \textbf{\dataset{}}, a new challenging benchmark of 1,000 examples to test spatial and rule-based reasoning on 2D pathfinding tasks. (\S \ref{section:dataset})
    \item We conduct extensive manual and automated evaluation with six human annotators, three state-of-the-art instruction-tuned (Qwen 2.5, GPT-4.1, Gemma 3) and four reasoning models (o4-mini, o3-mini, QwQ, R1) on \dataset. (\S \ref{sec:main_results})
    \item We analyze why models fail to solve puzzles (e.g., rule cell crossing), causes for reasoning mistakes (e.g., logical fallacies), and upper bounds for reasoning when increasing test-time compute by using pass@k sampling. (\S \ref{sec:path_errors}) %
    \item We perform various ablation studies on puzzle representation (e.g., prompt design, visual representation), and prompting (e.g., few-shot examples), underlining our results' robustness. (\S \ref{sec:ablations})
\end{itemize}

\section{Related Work}
\label{sec:related_work}

Benchmarking language models has shifted from core NLP tasks like question answering \cite{rajpurkar2016squad} and paraphrasing \cite{dolan2005automatically,wahle-paraphrase-2023, wahle-etal-2024-prompt} in GLUE \cite{wang2019gluemultitaskbenchmarkanalysis} to more complex evaluations, as these tasks have saturated. 
Long-horizon reasoning datasets, including MATH \cite{hendrycks_measuring_2021}, AIME \cite{AoPS_AIME}, BBH \cite{suzgun2022challengingbigbenchtaskschainofthought}, and MUSR \cite{sprague_musr_2024}, challenge models on multi-step problem-solving.
However, these benchmarks rely on data priors, knowledge recall, or pattern matching, enabling reasoning models like DeepSeek's R1 to saturate them, showing a gap in evaluating spatial reasoning and complex planning.

Specifically related to our proposal are rule-based and spatial benchmarks that use novel task representations or underrepresented ones in LLM training data (e.g., topological reasoning from text, abstract diagrams).
Notably, ARC-AGI \cite{Chollet2019Measure} tests abstract pattern recognition and inductive reasoning from few-shot 2D grid examples, showing that even in simple scenarios, the most advanced reasoning models fail.
However, ARC-AGI does not require step‑by‑step planning or following discrete rules.
VisualPuzzles \cite{song2025visualpuzzlesdecouplingmultimodalreasoning} presents algorithmic, analogical, and spatial riddles, but every task is multiple choice, so the model is not constructing individual solutions.
SpatialEval \cite{wang2024pictureworththousandwords} covers navigation, relation, and counting on images, 2D grids, and text.
However, SpatialEval mazes span only a few moves, and the counting or relation questions appear independently, not within one combined task.
PPNL \cite{Aghzal2023CanLL} tests spatial-temporal reasoning via 2D grid-based path planning.
It focuses on obstacle avoidance within the grid and does not incorporate complex, interacting rules.
Another related task is EnigmaEval \citep{wang2025enigmaevalbenchmarklongmultimodal}, but it does not focus on pathfinding.

\dataset{} addresses these limitations by requiring long-term, step-by-step path planning, where early errors in the reasoning chain can significantly impact later steps.
\dataset{} requires path-finding, counting, segregation, and logic involving colors and shapes in a single task, and on different-sized puzzle grids with complex, interacting rules.
Unlike other benchmarks, we also support problems with multiple correct solutions, allowing for testing different path-finding strategies and not relying on a single solution per example.

\section{Dataset}
\label{section:dataset}
The primary goal of \dataset{} is to test new pathfinding capabilities not represented in current benchmarks, specifically spatial navigation, rule understanding, constraint satisfaction, and multi-step planning, and also combinations in new ways, such as counting, segregation, and color or shape logic.
The design of the dataset is inspired by the puzzle mechanics of the video game \textit{The Witness} \citep{wiki:The_Witness_(2016_video_game)}, adapted into a format suitable for LLM assessment.

\subsection{Puzzle Rules}
\label{sec:game_rules}

Each puzzle in \dataset{} is a 2D grid of $m \times n$ \textbf{rule cells} with $(x,y) =(0,0) \in (m,n)$ being the top-left corner of the grid, and $x$ increases to the right, and $y$ downwards.
Rule cells are surrounded by \textbf{edges} that can be used to draw a path. 
There exists one \textbf{start point} on the edges (large circle) and one \textbf{end point} on the edges (extension of the edge). %
The goal of solving a puzzle is to move from the start point to the end point along the edges around the rule cells to fulfill all rule cell conditions. 
The path must be a single, continuous sequence of edges from the start to the end point, without crossing or overlapping itself at any edge segment.
Central to each puzzle are the rule cells, which we describe together with what it means to fulfill the rule cell condition. 
\Cref{sec:puzzle_subtypes_visual} contains puzzle examples to illustrate the components of our dataset.

\long\def\figwindownonum[#1,#2,#3,#4] {%
  \begin{window}[#1,#2,{#3},{\centering#4\par}] }
\def\endfigwindownonum{\end{window}}%

\newcommand{\iconheight}{5ex}

\newcommand{\puzzleitem}[3]{%
  \par\smallskip                    %
  \begin{figwindownonum}[0,l,{
    \includegraphics[width=\iconheight]{#1}},{}]
    \noindent\textbf{#2:} #3%
  \end{figwindownonum}%
  \par\medskip                     %
}

\puzzleitem{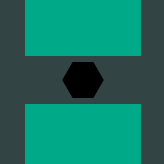}{Item Collection (Dots)}{%
The solution path needs to pass through every dot.
}

\puzzleitem{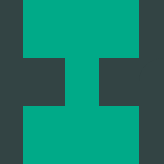}{Path Breaks (Gaps)}{%
The solution path cannot go through any edge segment containing a gap. Gaps act as local barriers.
}

\puzzleitem{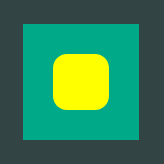}{Color Separation (Stones)}{%
The solution path must be drawn to separate stones of different colors. All stones located within any single enclosed region must be of the same color.
}

\puzzleitem{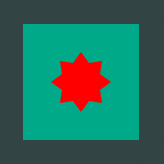}{Pairing (Stars)}{%
Each star must share its region with exactly one other symbol of the same color. No unpaired stars are allowed.
}

\puzzleitem{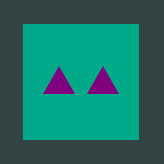}{Edge Count (Triangles)}{%
The solution path must touch the number of edges shown by the triangles in the cell, e.g., two triangles mean the path must touch exactly two edges of that cell.
}

\puzzleitem{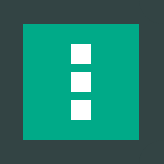}{Shape Fitting (Polyominoes)}{%
If a cell contains a polyomino (poly), the solution path must enclose a region that matches its exact shape and area. The region must not rotate or mirror the poly. Multiple polys can share a region if their shapes fit without overlapping.
}

\puzzleitem{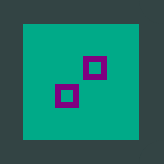}{Shape Subtraction (Ylop)}{%
A ylop must be enclosed in the same region as one or more polys. Its shape and area subtract from the total required by the polys. If a ylop cancels out a poly exactly, that pair imposes no constraint.
}

\subsection{Dataset Creation}

\paragraph{Generation.} %
Our process starts with randomly creating an $x$ by $y$ grid, where $x$ and $y$ range from 2 to 6 (e.g., $3 \times 5$). \Cref{fig:app_puzzle_index} in \Cref{ap:grid_indexing} provides an indexing example. %
We then randomly fill half of the grid with rule cells. The rule cell to grid cell percentage is termed \textit{rule density} and set a random start and end point. %

To solve puzzles automatically, we implement a generation-validation loop. 
First, we generate an initial puzzle and solve it using brute-force by exhaustively testing all valid paths from start to end.\footnote{Brute-force is necessary because many puzzles fall into NP or NP-Complete complexity classes \cite{abel2019witnesseswitnessfindingwitnesses}.} 
If the initial puzzle fails to produce a solvable puzzle, we decrease the rule density and regenerate the puzzle. 
Conversely, if the solver finds over $k$ distinct solutions (indicating the puzzle might be too unconstrained), rule density is increased, and the puzzle is regenerated. 
We found 50 solutions to be a reasonable hyperparameter choice for $k$ empirically by testing different generation setups.

We generate the \dataset\ dataset containing 500 training and 500 testing examples.
The distributions of different rules in \dataset{} are shown in \Cref{tab:dataset_summary_sections}. 
When sampling puzzles, we aim for an approximately equal distribution between rules. However, puzzles tend to have fewer stars (color pairing rule) and triangles (edge counting rule) than other rules. 
Observe how in \Cref{tab:dataset_summary_sections}, we only generated 25 puzzles containing ylops.
This is for two reasons: they can only exist if polys are available, and they are the hardest rule, as judged by humans.
For later tests on specific rules, we also created single-rule splits (more on this in \Cref{sec:main_results}).

\paragraph{Difficulty Estimation.} To quantify puzzle complexity, we created a difficulty metric that weights individual spatial reasoning tasks.
More specifically, the number of distinct rules, the total number of rule cells, the rule cell density, and an estimate of potential complex rule interactions. 
Each element contributes via a weighted sum to a raw score, which we then statistically normalize onto a standardized 1 (easiest) to 5 (hardest) scale (see \Cref{app:metric_details} for calculation specifics). 
As our later experiments with humans and reasoning models will demonstrate, this difficulty estimate is robust.
The distribution of difficulties of \dataset{} is detailed in \Cref{tab:dataset_summary_sections}.
We sample with an approximately uniform distribution between puzzles, ending up with slightly more level 3 (121) and level 2 puzzles (118) than level 1 (86), level 4 (86), and level 5 (89).

\begin{table}
\centering
\small
\begin{tabular}{lr}
\toprule
Statistics & Count \\
\midrule
\multicolumn{2}{l}{\textbf{Puzzles with Rule Type}} \\
\quad Gaps & 313 \\
\quad Dots & 292  \\
\quad Stones & 355  \\
\quad Stars & 210  \\
\quad Triangles & 233 \\
\quad Polygons & 305\\
\quad Ylops & 25\\
\multicolumn{2}{l}{\textbf{Puzzles with Difficulty Level}} \\
\quad Level 1 & 86 \\
\quad Level 2 & 118\\
\quad Level 3 & 121\\
\quad Level 4 & 86 \\
\quad Level 5 & 89\\
\bottomrule
\end{tabular}
\caption{\textbf{Counts} of puzzles for \dataset for different difficulties and rules based on the test set.} %
\label{tab:dataset_summary_sections}
\end{table}

\section{Experiments}
\label{sec:evaluation}
We assess \dataset{} through automated and manual studies.
In the automated evaluation, we consider instruction-based models - Gemma 3 27B \citep{gemmateam2025gemma3technicalreport}, Qwen 2.5 72B \citep{yang_qwen2_2024}, GPT 4.1 \citep{openai2025gpt41}; and reasoning models - QwQ 32B \citep{team_qwq_2024}, DeepSeek R1 Distill Llama 70B \citep{deepseekai2025deepseekr1incentivizingreasoningcapability}, o3-Mini \citep{openai2025o3mini}, and o4-mini \citep{openai2025o3o4mini}. 
We measure model accuracy on solving our puzzles (\Cref{sec:main_results}), performance on specific rule cells, reasoning errors (\Cref{sec:path_errors}), and conduct ablation studies regarding the stability of our findings (\Cref{sec:ablations}).
For the manual inspection, we test human performance and time on the same puzzles (\Cref{sec:human_baseline}). 
We used six annotators (aged 22-27) with CS backgrounds. %

\paragraph{Setup.}

All puzzles are presented to the LLMs using prompts with a human-annotated example solution.
Our textual representation is inspired by the ARC challenge \cite{Chollet2019Measure}.
Extraction occurs using regex after a predefined sequence of ``\#\#\#\#'' as stated in the prompt. %
By default, we provide a one-shot example with a human-annotated path, as it yielded the best results (\Cref{sec:ablations}).
\Cref{app:prompts_examples} contains details about the prompts, examples, and their solution.
Details on models, hardware, and tokens processed are in \Cref{app:model_hardware}.

\subsection{Main Results}
\label{sec:main_results}

We present key baseline evaluations across models and difficulty. 
Scaling test-time compute allows us to identify upper bounds of model capabilities.

\begin{figure}[t]
    \centering
    \includegraphics[width=\linewidth]{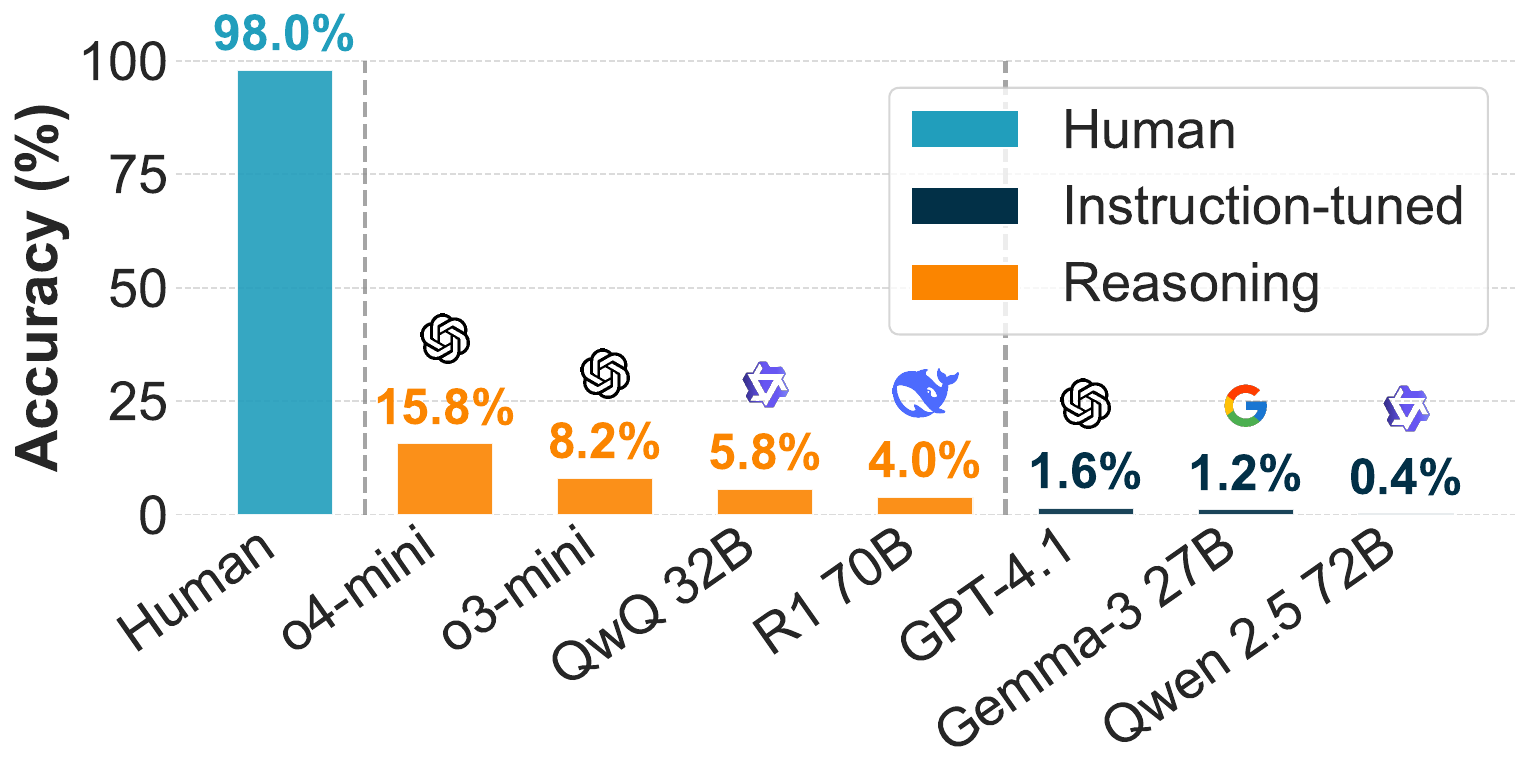}
    \caption{
      \textbf{Accuracy (\%)} of \textit{human} annotators (teal) against different LLMs: reasoning models (orange) and instruction-tuned models (blue). Higher is better.
    }
    \label{fig:llm_human_solve_rates}
\end{figure}

\paragraph{Baselines.}
We want to understand how reasoning- and instruction-tuned LLMs solve spatial multi-step reasoning tasks compared to humans. 
We compute accuracy (\% of solved puzzles) for these models. 
Human baseline results use majority votes from three annotators per puzzle (details on the human evaluation later in \Cref{sec:human_baseline}). 
\Cref{fig:llm_human_solve_rates} shows accuracy for humans and LLMs. 
Humans solve puzzles nearly perfectly at 98.0\% (98/100 puzzles solved).
The top reasoning model, o4-mini, performs much worse at 15.8\% (79/500 puzzles). %
GPT-4.1 is the best instruction-tuned model at 1.6\% (8/500 puzzles). 
Reasoning models perform better overall (avg. 8.5\%). %
Closed models outperform open ones: o4-mini (15.8\%) and o3-mini (8.2\%) versus R1 70B (4.0\%) and QwQ (5.8\%), with similar trends in instruction-tuned models. 
Results suggest these puzzles are very challenging for LLMs, while relatively easy for humans.
We hypothesize errors arise from models' spatial understanding limitations, such as misunderstanding rules, logical fallacies, and misinterpreting grid representations \cite{huang2023reasoninglargelanguagemodels,turpin2023languagemodelsdontsay}. 

\paragraph{Difficulty Level.}
We decompose the results in \Cref{fig:llm_human_solve_rates} by difficulty.
We compare the best model (o4-mini) against human performance. 
Humans achieve 100\% accuracy at difficulty level 1, while o4-mini reaches 47.7\%, showing it solves nearly half of the simple puzzles. 
At level 2, o4-mini drops to 19.5\%, but humans remain at 100\%. 
For higher difficulties, with larger grids and complex rules, o4-mini’s rate decreases further, reaching 1.2\% at level 4 (solving 1 of 89 puzzles), compared to 94.4\% for humans. 
Level 5 shows similar results to level 4 (similar trends across all models). %
Results for all models are in \Cref{app:detailed_results}.
Overall, LLMs have severe reasoning challenges as puzzle difficulty increases. 
A possible explanation could be that models conclude reasoning prematurely in complex puzzles by ignoring certain rules and running into dead ends. 
Specific rules or combinations of rules may also be particularly challenging.

\begin{figure}[t]
    \centering
    \includegraphics[width=0.95\linewidth]{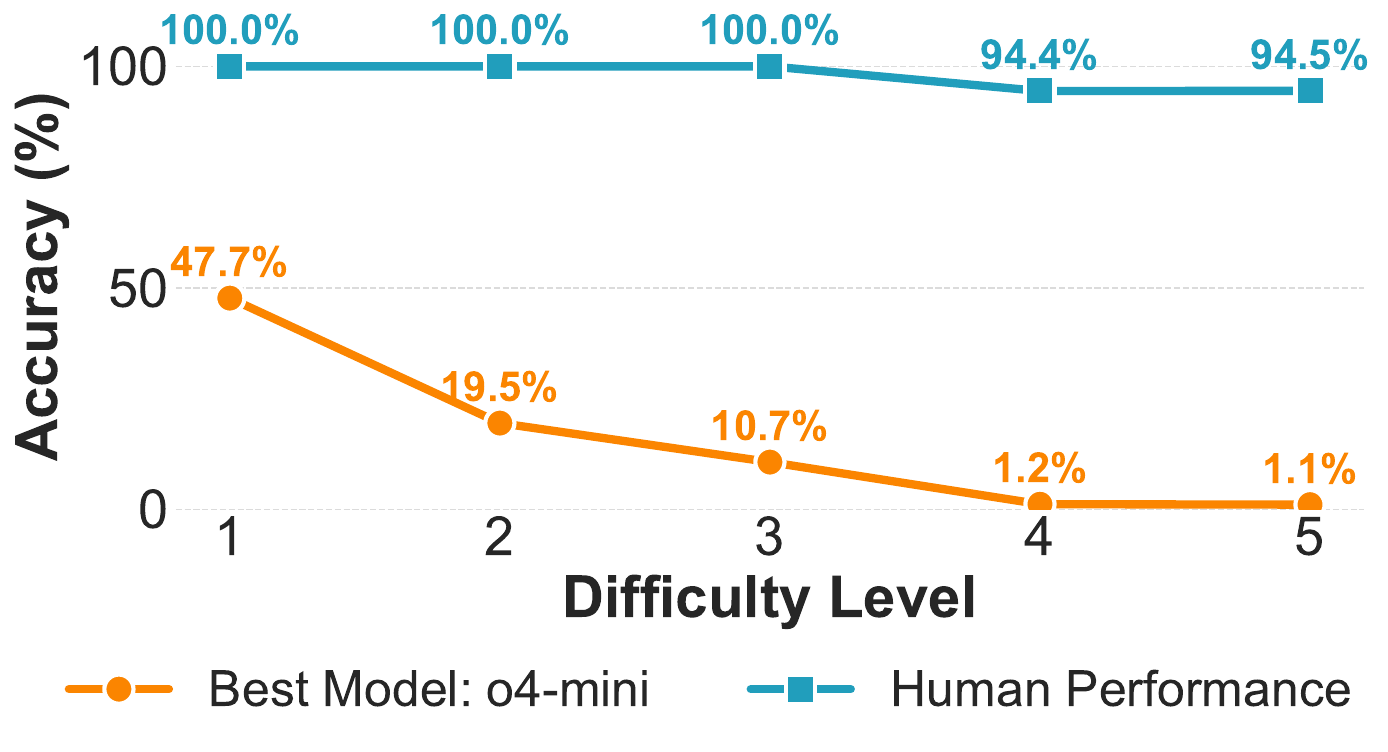}
    \caption{
      \textbf{Accuracy (\%)} at different difficulty (1-5) between o4-mini (orange) and human annotators (teal). Higher is better.
    }
    \label{fig:solve_rate_vs_difficulty}
\end{figure}

\begin{figure*}
    \centering
    \includegraphics[width=\textwidth]{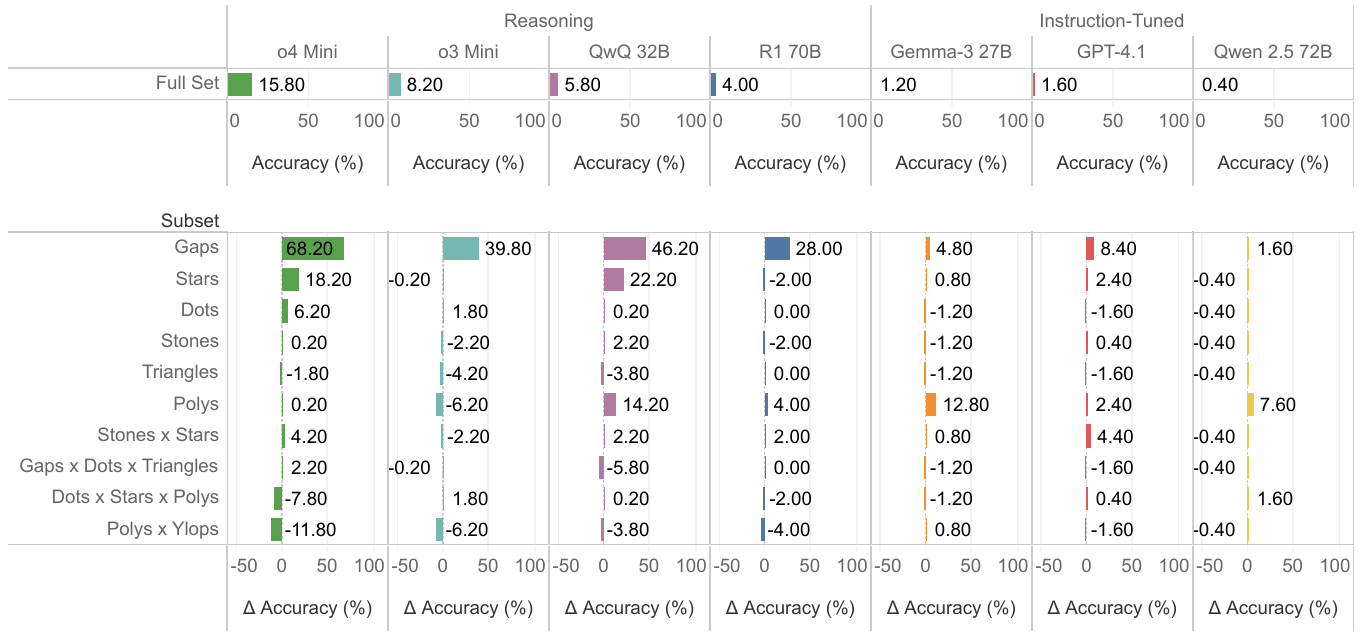}
    \caption{
      Performance of models on puzzles containing only specific rules. 
      Columns represent individual models, for reasoning- and instruction-tuned models. 
      The \textit{Full Set} row shows the \textbf{Accuracy (\%)} per model across all puzzles. 
      The rows below show accuracy on specific rules minus accuracy on the full set (\textbf{$\Delta$ Accuracy (\%)}).
    }
    \label{fig:llm_subset_performance_delta}
\end{figure*}

\paragraph{Rule Specific Analysis.}
We examine the accuracy of models on splits containing individual rules or specific rule combinations to analyze which rules the models succeed or fail on. 
Specifically, we create puzzles only containing \textit{gaps}, \textit{dots}, \textit{stones}, \textit{stars}, \textit{triangles}, \textit{polys}, or \textit{ylops}.
We also create multi-rule combination splits to investigate how models handle the interaction between a few distinct types of rules, \textit{stones x stars}, \textit{gaps x dots x triangles}, and \textit{dots x stars x polys}. 
Because \textit{ylops} can only exist in the presence of polys, this split contains puzzles with ylops and polys. 
Each split contains 50 training and 50 test samples, and we also make them available in our release.

\Cref{fig:llm_subset_performance_delta} shows accuracy for the primary dataset (top row) and the relative performance delta ($\Delta$) of specific splits (e.g., gap accuracy minus full set accuracy; bottom rows).

The gaps split shows superior performance across all models, whereas dots hover near the average model performance on all puzzles. 
Dots and gaps tasks are similar yet differ in performance: gaps explicitly forbid using edges, providing immediate error feedback, whereas dots require edge use, with errors apparent only after path completion. 
Polys produce mixed results; stronger models (o4-mini, o3-mini) show minor performance differences compared to all puzzles, while smaller reasoning and instruction-tuned models markedly improve. 
Polys and ylops lead to substantial performance decreases, which are also the most challenging rules perceived by humans. 
Some weaker models (QwQ, Gemma) markedly outperform their average on polys (improvements of 13.2 and 12.8 points, respectively), suggesting smaller models might solve some puzzles more intuitively, while others tend to ``overthink'' problems, leading to higher success in simpler setups (more details later in \Cref{sec:path_errors}). 
Performance differences may also result from fundamental path construction errors, logical mistakes, or model rule misinterpretations.

\subsection{Path Errors and Reasoning Mistakes}
\label{sec:path_errors}

We analyze model-constructed paths and their reasoning tokens to shed light on why reasoning models fail to solve puzzles.

\begin{figure}[t]
    \centering
    \includegraphics[width=\linewidth]{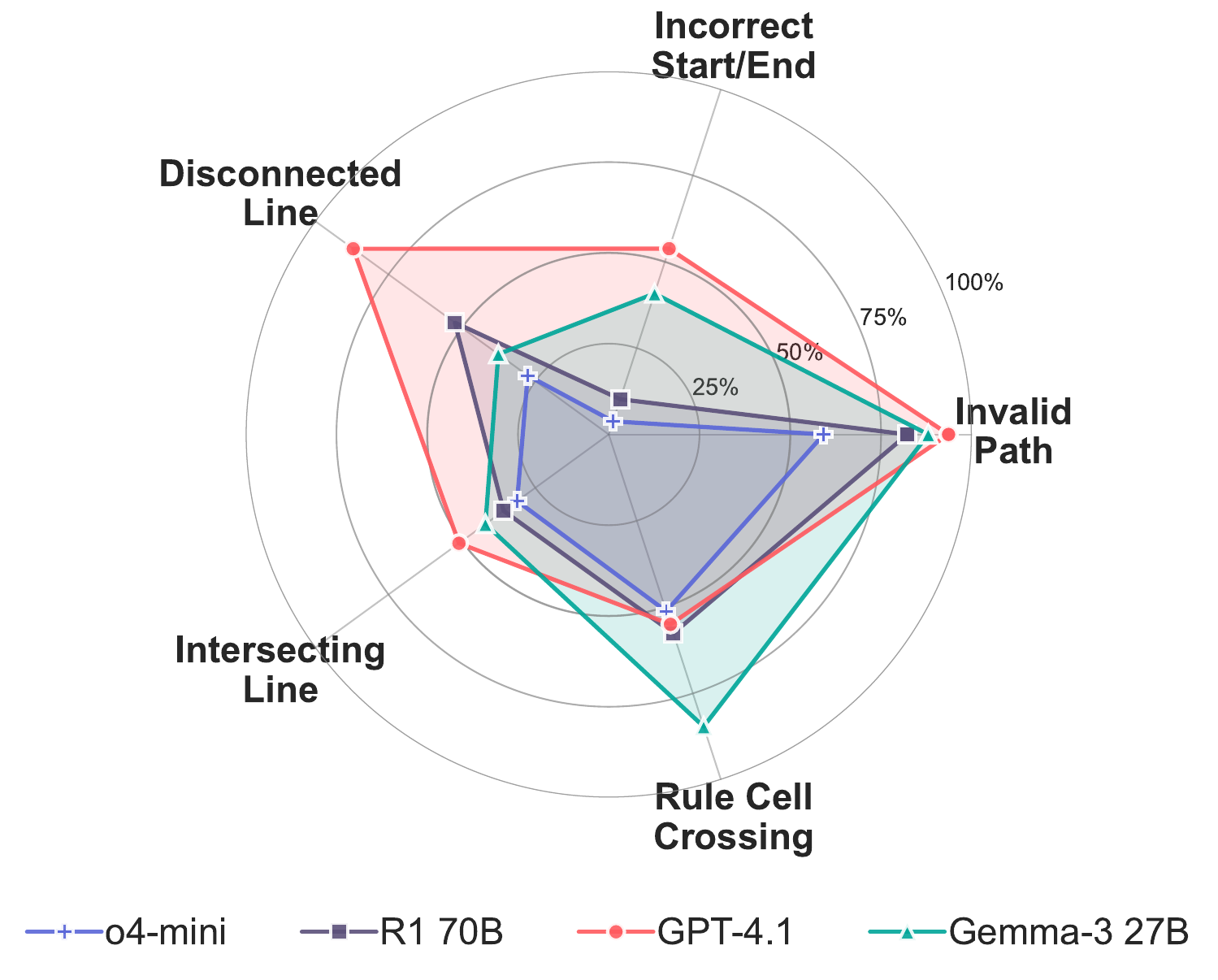}
\caption{
  Analysis of \textbf{Path Errors (\%)} in generated solutions for different LLMs. %
  Each corner shows a specific error, and the distance from the center indicates the \% of generations with that error. Lower is better.
}
\label{fig:llm_path_violations}
\end{figure}

\paragraph{Path Errors} %
We analyze common errors in constructing a valid path (e.g., ignoring rules to solve the game for now). 
We assess five error types for all models: \textit{Incorrect Start/End} (i.e., line starts or ends at the wrong edge), \textit{Disconnected Line} (i.e., line not continuous), \textit{Intersecting Line} (i.e., line crosses an edge multiple times), \textit{Rule Cell Crossing} (i.e., line does not stay on edges but crosses rule cells).
Paths with any such errors are deemed \textit{Invalid Path}.
Examples for each error type can be found in \Cref{sec:appendix_visual_examples}.

\Cref{fig:llm_path_violations} shows the percentage of path rule violations for four selected models.
Results for all models can be found in \Cref{tab:llm_path_violation_rates} in \Cref{app:detailed_results}.
Smaller enclosed areas in the figure imply better adherence to path rules. 
The two reasoning models (o4-mini and R1 70B) have similar violation patterns, but o4-mini performs better overall. 
Notably, over 50\% of puzzles fail because models do not construct valid paths. 
Instruction models (GPT-4.1, Gemma-3 27B) perform worse, showing distinct weaknesses.
GPT-4.1 frequently produces disconnected lines, while Gemma-3 27B commonly crosses rule cells. 
Interestingly, Gemma-3 27B produces fewer disconnected lines than the larger reasoning model R1 70B. 
Reasoning models have higher accuracy despite similar basic path errors, suggesting successful path construction is only the first hurdle. %
Across models, the most common error is \textit{Rule Cell Crossing}, indicating frequent violations by paths moving through rule cells rather than along edges. 
However, up to this point, our explanations of other model failures have been largely hypotheses, and the precise underlying causes require further investigation.

\noindent\paragraph{Causes for Reasoning Mistakes.} To shed light on the ``why'' of reasoning model failures, we manually analyzed R1 70B puzzles (as it openly provides reasoning tokens) with the puzzles containing only single rule types (e.g., only stones). We selected puzzles where models produced valid paths (without path errors) but failed to fulfill all rule cells. This resulted in 48 puzzle solutions for analysis.

Models most commonly failed due to logical fallacies (36/48), grid/index system misinterpretation (26/48), and careless shortcutting of multiple reasoning steps (23/48). Interestingly, R1 often recognized mistakes or dead ends (25/48) before concluding, indicating limited reasoning but awareness of its constraints.

Different splits revealed specific reasoning limitations. With dots, models typically recognized missed ones during path construction but failed to correct their paths accordingly (e.g., \Cref{fig:reasoning_mistakes_puzzle_dots} in \Cref{ap:reasoning_mistakes}). With gaps, models frequently made careless, unvalidated multi-step moves, violating rules by crossing gaps (e.g., \Cref{fig:reasoning_mistakes_puzzle_gaps} in \Cref{ap:reasoning_mistakes}). We provide further examples with highlights of R1's reasoning tokens in \Cref{ap:reasoning_mistakes}.

Mistakes, like unvalidated multi-step moves and grid misinterpretation, highlight ongoing challenges in long-term spatial planning, as even minor shortcuts lead to significant rule violations. However, models' recognition of errors and dead ends points toward opportunities and gives space for future contributions to improve spatial reasoning, e.g., via iterative reasoning or sampling multiple parallel paths with strategies to find correct ones.

\begin{figure}
    \centering
    \includegraphics[width=\linewidth]{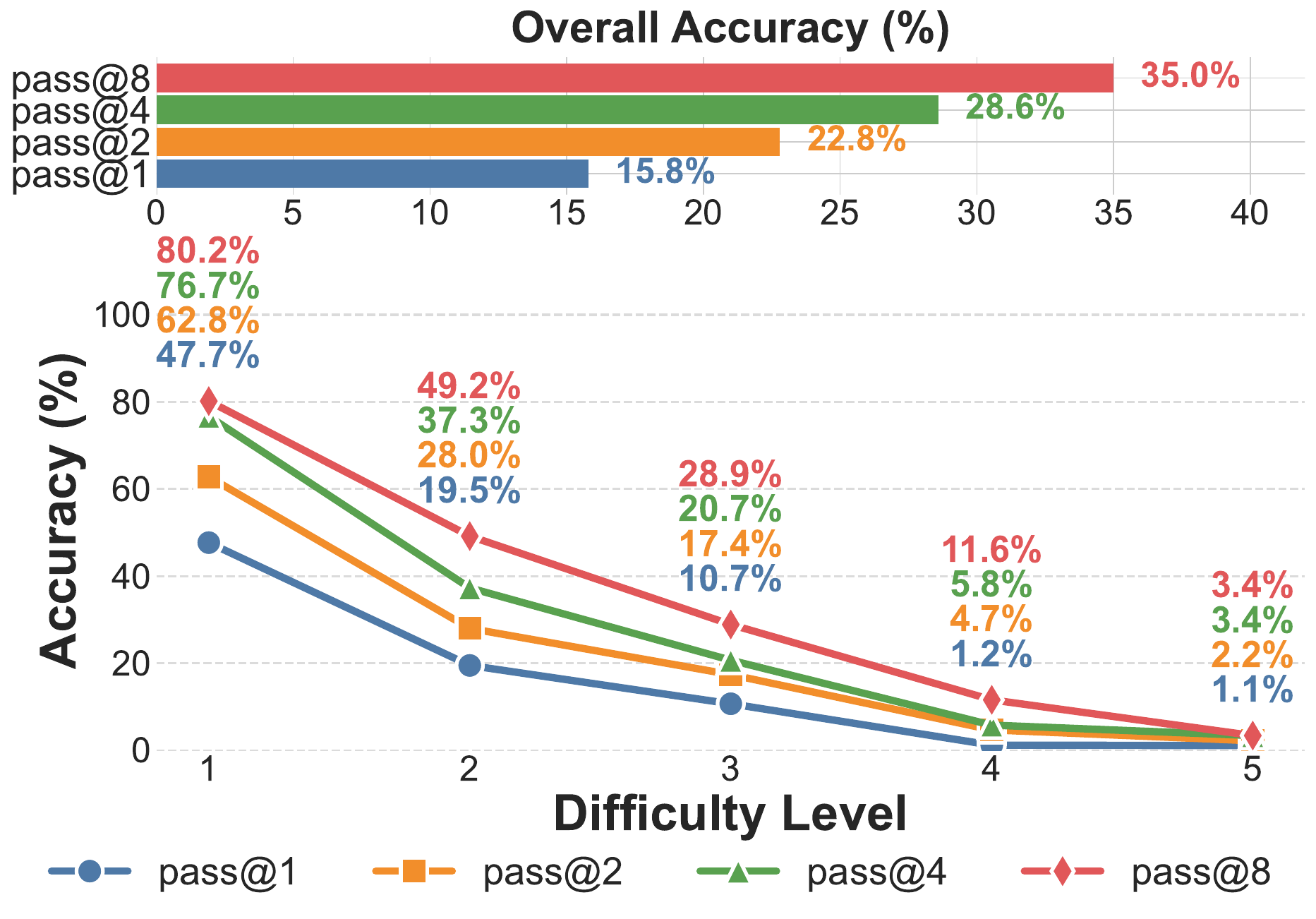}
    \caption{
      \textbf{Accuracy (\%)} for generating $k \in \{1, 2, 4, 8\}$ solutions and evaluating whether the correct path is in one of the $k$ attempts (pass@k) for o4-mini across difficulty (1-5). Higher is better.
    }
    \label{fig:o4mini_top_k_performance}
\end{figure}

\paragraph{Upper Reasoning Bounds.}
A common strategy to improve performance on complex tasks is to scale test-time compute, for instance through multi-agent debate \citep{becker2025mallmmultiagentlargelanguage}.
However, this approach can suffer from performance degradation in discussions requiring longer reasoning chains \citep{becker2025stayfocusedproblemdrift}. 
Given that SPaRC requires long-term, step-by-step planning where early errors can be critical, this makes such a debate-based approach potentially less suitable. 
Therefore, to determine models' upper limits, we purposefully increase test-time compute by generating up to eight attempts per puzzle for each model (i.e., pass@1 to pass@8).

\Cref{fig:o4mini_top_k_performance} shows accuracy rising from 15.8\% (pass@1) to 35.0\% (pass@8) for o4-mini. 
This improvement is expected as we scale computation.
Importantly, this setting is not practical at test-time, as we only verify if the solution appears among the $k$ generations. 
In practice, a decision mechanism like majority voting would be more suitable \cite{kaesberg2025votingconsensusdecisionmakingmultiagent}.

Still, additional attempts are not sufficient to solve complex puzzles. 
Success rates improve by 32.5 points for level 1 puzzles (easy), but only 2.3 points for level 4 and 5 (difficult ones).
This shows that our puzzles cannot be easily solved by just increasing the computation, but the reasoning steps have to get more sophisticated and have to adjust according to the difficulty level.
Higher results for larger $k$ give hope that future work can find better training methods to improve reasoning.

\begin{figure}
    \centering
    \includegraphics[width=\linewidth]{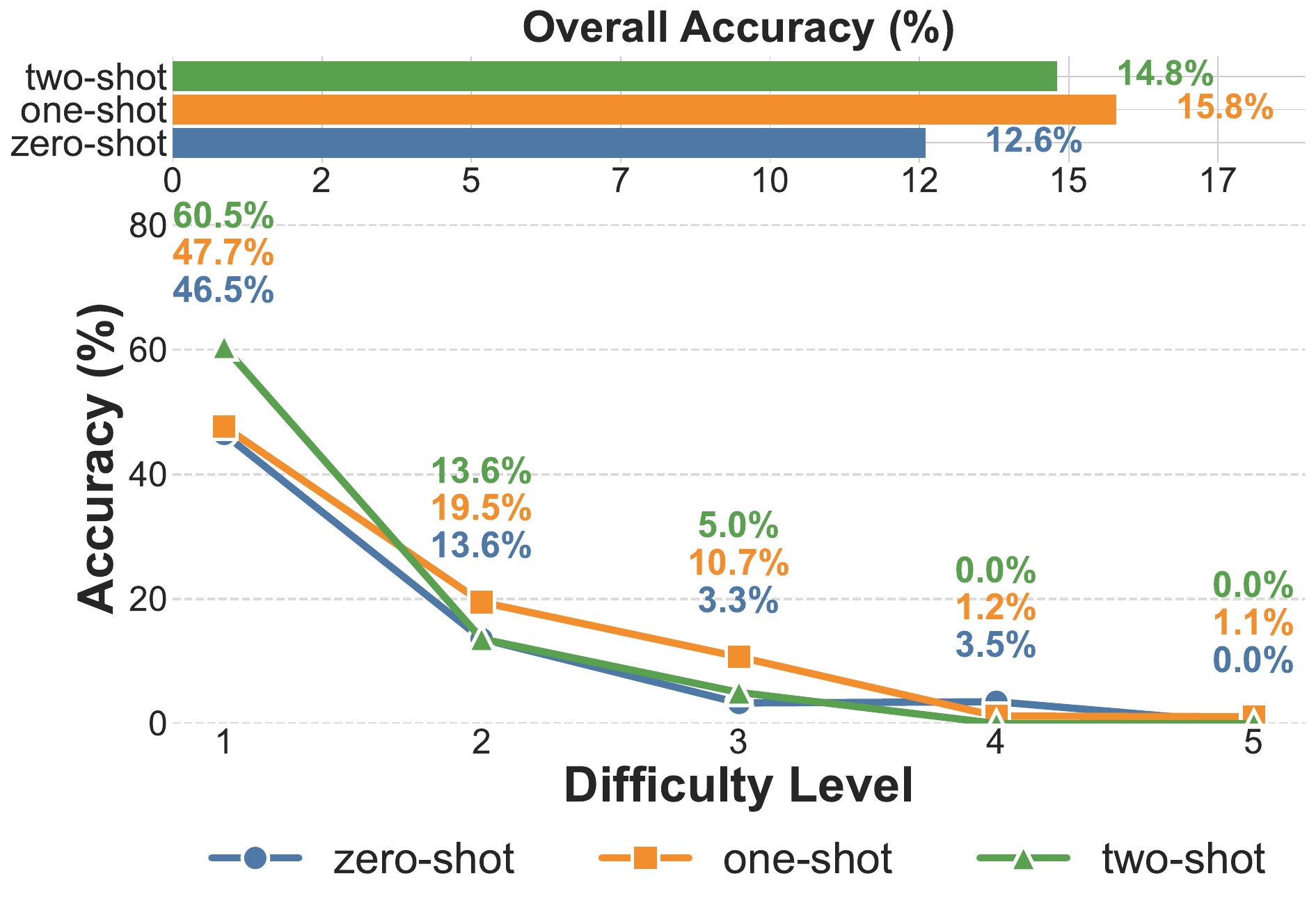}
    \caption{
      \textbf{Accuracy (\%)} for zero-shot (blue), one-shot (orange), and two-shot (green) examples provided to o4-mini across difficulty (1-5). Higher is better.
    }
    \label{fig:o4mini_fewshot_comparison}
\end{figure}

\subsection{Ablations}
\label{sec:ablations}

We investigate how changes to the prompting (e.g., few-shot examples, different prompts) and puzzle representation (e.g., text and images) affect our results through various ablations.

\paragraph{Few-Shot Prompting.}
We investigate the effect of in-context learning by comparing zero-shot, one-shot, and two-shot configurations (see \Cref{app:prompts_examples} for few-shot examples). 
Previous experiments always defaulted to one-shot.

\Cref{fig:o4mini_fewshot_comparison} shows that one-shot has the highest overall accuracy (15.8\%), while zero-shot performs worst (12.6\%) for o4-mini. 
At difficulty 1, two-shot outperforms one-shot, but this reverses at higher levels. 
Examples generally help model comprehension, but too many examples seem to have no benefit, and sometimes negatively impact performance.
Additional analysis in \Cref{ap:few-shot-details} shows that zero-shot has fewer path violations than few-shot.

Improved one-shot over zero-shot performance is expected, but two-shot's slightly lower performance than one-shot is surprising, as more examples should clarify rule interactions; however, given small differences, stochastic variance is possible. 
Similar findings were reported by \citet{Ye2023ACC}, suggesting increased examples do not always help, possibly due to cognitive overload or excessive focus on example analysis instead of task solving.

\begin{figure}[t]
    \centering
    \includegraphics[width=\linewidth]{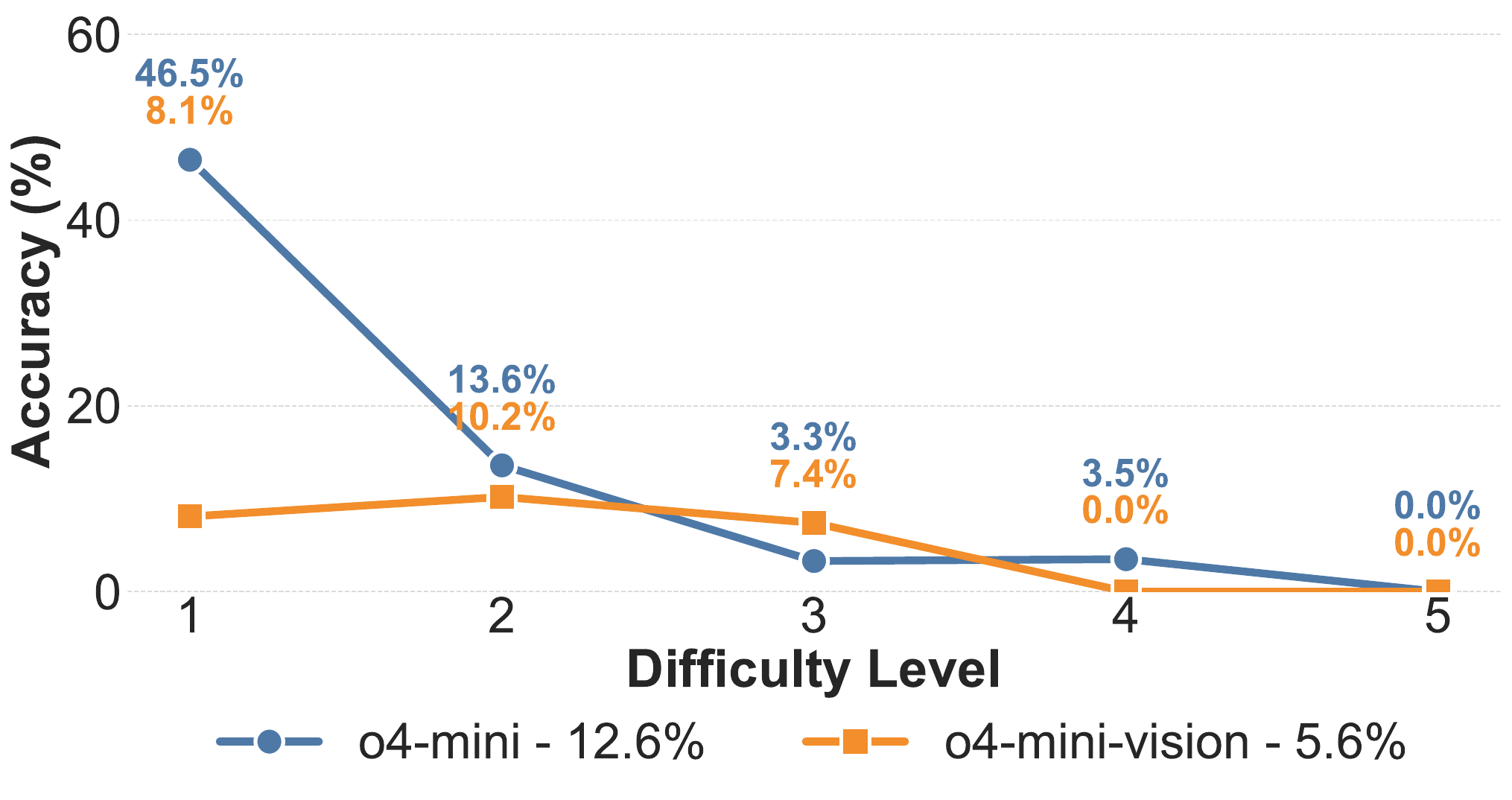}
\caption{
  Comparison of \textbf{Accuracy (\%)} for o4-mini using a textual representation (blue) vs. a puzzle screenshot (orange) across difficulty (1-5). Higher is better.
  }
\label{fig:o4mini_configs_solve_rate}

\end{figure}

\paragraph{Visual Representation.} 
Another factor that might influence our results is the textual 2D representation of the puzzles.
Therefore, we provide screenshots of the puzzle, similar to  \Cref{fig:example-puzzle}, and adjust the prompt accordingly.
We compare visual results to zero-shot textual results, as the visual prompt lacks an example solution.
Details on this configuration are available in \Cref{app:prompts_vision}.

\Cref{fig:o4mini_configs_solve_rate} compares the accuracy of o4-mini using default textual prompts versus visual prompts across difficulty levels. 
The visual representation reduces overall performance from 12.6\% to 5.6\%. 
The gap between text and image prompts is larger at easier difficulty levels, but diminishes at higher difficulty levels. 
Additional analysis in \Cref{ap:vision-details} shows that a main cause for bad results on easy puzzles is invalid path constructions.

These results suggest that current textual representations are easier for multi-modal reasoning models to understand. 
Likely, connecting textual descriptions to visual puzzle elements adds complexity compared to purely textual prompts. 
However, whether the current textual representation is also optimal requires more investigation. %

\paragraph{Alternative Prompt.}
We test if our results are affected by different formulations in our prompts, i.e., prompt engineering \cite{white2023prompt, wahle-etal-2024-prompt}.
Because paths previously failed due to violations of path rules, we adjusted the prompt to emphasize these rules more explicitly. 
This adjustment improved o4-mini’s accuracy from 15.8\% to 21.0\% and reduced path errors, with \textit{Rule Cell Crossing} decreasing from 51.2\% to 29.0\%, and \textit{Intersecting Line} dropping from 31.2\% to 21.2\%. 
However, at higher difficulty (level 5), there was no improvement, with o4-mini still only solving 1 out of 89 puzzles (more details in \Cref{fig:o4mini_prompt_comparison} in \Cref{ap:ablations}). 
Prompt engineering moderately increases performance at lower difficulty levels, but it does not have a marked impact at higher levels.
In additional experiments, long-context reasoning-tuned models provided only modest gains (see \Cref{fig:prompting} in \Cref{app:prompt_repre}).
Markdown-based grids also did not improve over the array format puzzle representation (see \Cref{fig:representation} in \Cref{app:prompt_repre}). 
These results suggest that low task performance is due to limited spatial reasoning skills rather than representation style or prompting.

\begin{table}[t]
\centering
\small
\resizebox{\linewidth}{!}{
\begin{tabular}{lrrrrr}
\toprule
Difficulty Level & 1 & 2 & 3 & 4 & 5 \\
\midrule
Hum. Acc. (\%) & 100 \% & 100 \%  & 100 \% & 94.4 \%  & 94.5 \%  \\
Hum. Avg. (s) & 10.7 & 18.3 & 26.7 & 60.7 & 131.5 \\
Hum. Mdn. (s) & 7.1 & 13.7 & 15.6 & 28.8 & 85.6 \\
\midrule
QwQ Acc. (\%) & 20.9\% & 5.9\% & 2.5\% & 1.2\% & 0.0\% \\
QwQ \#Tokens & 14433 & 14200 & 13983 & 14072 & 13114\\
Qwen 2.5 Acc. (\%) & 0.0\% & 1.7\% & 0.0\% & 0.0\% & 0.0\% \\
Qwen 2.5 \#Tokens & 790 & 888 & 953 & 1037 & 1161 \\
\bottomrule
\end{tabular}}
\caption{
  \textbf{Accuracy}, \textbf{Average} and \textbf{Median} human solve time (seconds), and \textbf{Accuracy} and \textbf{Number of (\#) Generated Tokens} for QwQ 32B and Qwen 2.5 72B over \textbf{Difficulty Level} (1-5). 
}
\label{tab:human_solve_times}

\end{table}

\subsection{Human Evaluation}
\label{sec:human_baseline}

For a human baseline, we asked six annotators aged 22-27, with a background in computer science and data science, to solve 100 i.i.d. drawn puzzles from the dataset, divided into two subsets of 50 samples each.
Each of the 100 puzzles is annotated three times, and a puzzle is marked as solved if the majority found a correct solution. 
Even though we did not test all 500 test samples of \dataset, sampling i.i.d., and using two non-overlapping sets with three annotators each gives us a fair estimate of human performance.
We recorded the accuracy, number of attempts, and solving time. 
Details on annotation instructions are in \Cref{app:human_annotator_instructions}.

\Cref{tab:human_solve_times} shows humans achieve near-perfect performance, with 100\% accuracy at difficulty levels 1–3 and around 95\% at levels 4 and 5. 
Average solve time increases exponentially with difficulty, from 10.7 seconds for difficulty 1 to 26.7 seconds for difficulty 3, then starkly increasing to 60.7 seconds for difficulty 4 and 131.5 seconds for difficulty 5. 
Median solve times are consistently lower than average times, indicating that a few very difficult puzzles significantly increase the average.

Compared to humans, models show two relevant time-scaling aspects.
First, previous pass@k experiments (\Cref{fig:o4mini_top_k_performance} in \Cref{sec:ablations}) showed that multiple attempts to solve puzzles improved performance on easy puzzles but did not increase performance on difficult puzzles. 
Second, analyzing the number of generated tokens (\Cref{tab:human_solve_times}), instruction-tuned models such as Qwen 2.5, increase token counts with puzzle difficulty (from 790 to 1161), while reasoning models, such as QwQ maintain relatively constant token counts across difficulties (14433 to 13114). See \Cref{tab:tokens_by_difficulty} in \Cref{app:tokens_by_difficulty} for all models. 
This suggests models do not effectively scale spatial reasoning at test-time.

\section{Conclusion}
\label{sec:epilogue}

We introduced \dataset{}, a dataset of 1,000 examples designed to evaluate spatial and rule-based reasoning capabilities on 2D grid pathfinding puzzles.
This dataset tests reasoning skills not evaluated by existing benchmarks, focusing specifically on multi-step constraint satisfaction problems requiring spatial and rule-based reasoning.

We evaluated puzzles with six human annotators, three instruction-tuned models (GPT-4.1, Gemma 3, Qwen 2.5), and four reasoning models (o4-mini, o3-mini, QwQ, R1). 
Humans achieved a 98\% accuracy. The best reasoning model, o4-mini, reached only 16\%. 
Performance was drastically affected by puzzle difficulty, with models solving 48\% at level 1, 20\% at level 2, and just 1.1\% at level 5. 
Humans consistently solved puzzles across levels, including 95\% at level 5.
Our error analysis revealed that path errors and reasoning mistakes stemmed from logical fallacies, grid misunderstandings, and performing too many reasoning steps at once.
Generating up to eight attempts per puzzle improved accuracy up to 30\% for difficulty 1 puzzles and 2\% for difficulty 5. %
Humans needed up to 13 times more time to solve hard puzzles than easy ones, and instruction-tuned models scaled token usage with difficulty by $\sim40\%$. 
Reasoning models showed only a $\sim5\%$ increase for harder difficulties. %
Ablation studies on visual puzzle representation, prompting, and few-shot examples show only mild variations and support the robustness of our results. 

Empirically, \dataset{} reveals critical limitations in current reasoning models regarding spatial reasoning, rule-based reasoning, multi-step planning, and constraint satisfaction. 
Existing methods, including enhanced prompting and increased computational sampling, offer only partial improvements.
Fundamental advances in model reasoning capabilities are still needed to reach human-level results.

\section*{Limitations}
\label{sec:limitations}

Our evaluation depends on a fixed delimiter (``\#\#\#\#'') and a regex that collects the following coordinate list. 
When a model omits the delimiter, writes several delimiter lines, or inserts natural language text between coordinates, extraction can fail, producing false negatives. 
These events are rare in practice, and we stress the required format in every prompt, but complete robustness is unattainable when testing many different models.

OpenAI models (i.e., o4-mini, o3-mini) return only final coordinates with a small explanation, but redact intermediate reasoning tokens.
Consequently, detailed failure analysis is restricted to open models like R1 70B. 
Intermediate reasoning can differ from final answers in models of any scale, as previously documented by \citet{turpin2023languagemodelsdontsay,chen2025reasoningmodelsdontsay}, thus potentially limiting generalization from trace-based analyses.

The dataset covers single-rule puzzles and a limited set of two- and three-rule combinations but does not exhaustively represent all possible interactions among the seven rule types. 
Future releases could introduce underrepresented combinations (e.g., \textit{stars × triangles × polys × ylops}) to probe generalization more comprehensively. 
However, as models fail on most easy tasks already and current splits reveal clear error patterns and support comparative ranking of the different rule types, we leave this to future work when models become more capable.

The poly set in \Cref{fig:llm_subset_performance_delta} in \Cref{sec:main_results} shows improvements for weaker but not stronger models. 
The poly rule sometimes fills the entire grid with poly shapes, necessitating a path along the grid's edge. 
This condition impacts only the poly subset, explaining performance spikes. 
Smaller models find this shortcut more frequently, likely because simpler solutions emerge when overwhelmed by many complex poly shapes.

\section*{Acknowledgements}
This work was partially supported by the Lower Saxony Ministry of Science and Culture and the VW Foundation.
Many thanks to Andreas Stephan, Tianyu Yang, Zeinab Taghavi, and Annika Schulte-Hürmann for their thoughtful discussions and feedback.

\bibliography{custom}
\clearpage
\appendix

\section*{\Large Appendix}
\section{Models \& Hardware}
\label{app:model_hardware}

This section details the large language models (LLM) used in our experiments, the hardware on which they were run, and the approximate number of tokens processed for each model. 

For open models we used \href{https://huggingface.co/google/gemma-3-27b-it}{Gemma 3 27B} \citep{gemmateam2025gemma3technicalreport}, \href{https://huggingface.co/Qwen/QwQ-32B}{QwQ 32B} \citep{team_qwq_2024}, \href{https://huggingface.co/Qwen/Qwen2.5-72B-Instruct}{Qwen 2.5 72B} \citep{yang_qwen2_2024} and \href{https://huggingface.co/deepseek-ai/DeepSeek-R1-Distill-Llama-70B}{DeepSeek R1 70B} \citep{deepseekai2025deepseekr1incentivizingreasoningcapability}.

For propietary models we used \href{https://platform.openai.com/docs/models/gpt-4.1}{GPT-4.1} \citep{openai2025gpt41}, \href{https://platform.openai.com/docs/models/o3-mini}{o3-mini} \citep{openai2025o3mini} and \href{https://platform.openai.com/docs/models/o4-mini}{o4-mini} \citep{openai2025o3o4mini}. For both OpenAI reasoning models, the default medium effort reasoning mode was used.

\Cref{tab:model_hardware_tokens} shows the details regarding model size, tokens processed, and hardware used.

\begin{table}[h!]
\centering

\resizebox{\linewidth}{!}{\begin{tabular}{l l r l}
\toprule
Model Name & Size & Tokens Processed & Hardware \\
\midrule
\multicolumn{4}{l}{\textbf{Open Models}}\\
\quad Gemma 3  & 27B   & 875,711        & 4x Nvidia A100 \\
\quad QwQ          & 32B   & 13,863,364     & 4x Nvidia A100 \\
\quad Qwen 2.5    & 72B   & 955,167        & 8x Nvidia A100 \\
\quad DeepSeek R1  & 70B   & 9,136,467      & 8x Nvidia A100 \\
\multicolumn{4}{l}{\textbf{Propietary Models}}\\
\quad GPT-4.1     & N/A   & 5,057,588      & OpenAI API \\
\quad o3-mini      & N/A   & 19,776,881     & OpenAI API \\
\quad o4-mini      & N/A   & 59,192,466     & OpenAI API \\
\bottomrule
\end{tabular}}
\caption{Overview of models, hardware, and token counts. Token counts are approximate.}
\label{tab:model_hardware_tokens}
\end{table}
For all ablations and the main study, o4-mini was analyzed on 6500 puzzles overall. For a comparable 1000 puzzles, this would equate to approximately 9,865,411 tokens. Both OpenAI reasoning models were used with the medium reasoning effort.

\section{Grid Indexing}
\label{ap:grid_indexing}
\begin{figure}[H]
    \centering
    \includegraphics[width=0.9\linewidth]{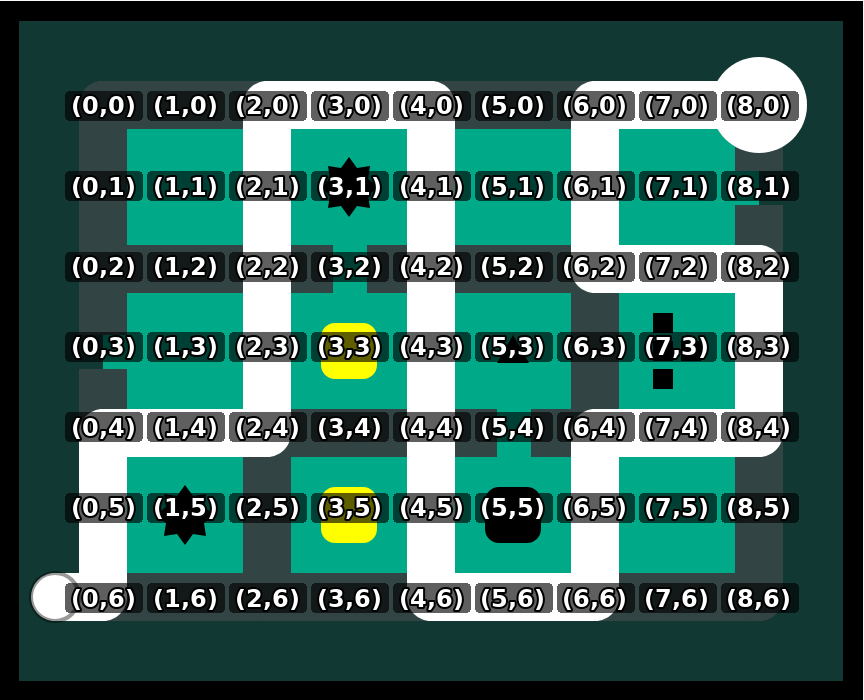}
    \caption{Puzzle grid from \Cref{fig:example-puzzle} with all grid cells annotated with their coordinates.}
    \label{fig:app_puzzle_index}
\end{figure}

\Cref{fig:app_puzzle_index} shows a puzzle grid with all its coordinates according to the prompts in \Cref{app:prompts}.
\label{app:indexing}

\section{Licenses and Code Acknowledgments}
\label{sec:licenses-and-acknowledgments}

The source code developed and used in this work is provided under the BSD 3-Clause License. This licensing choice is required due to dependencies on code from the following repositories, which are partially distributed under the BSD 3-Clause License:

\begin{itemize}
    \item \href{https://github.com/jbzdarkid/jbzdarkid.github.io}{jbzdarkid/jbzdarkid.github.io}
    \item \href{https://github.com/NewSoupVi/The-Witness-Randomizer-for-Archipelago}{NewSoupVi/The-Witness-Randomizer-for-Archipelago}
\end{itemize}

We gratefully acknowledge the authors of these repositories for making their implementations publicly available.

The datasets used in this study are provided under the Creative Commons Attribution 4.0 International (CC-BY-4.0) license.

\section{Difficulty Metric Calculation}
\label{app:metric_details} 

This section provides the details for calculating the difficulty metric used to rate \dataset\ puzzles in this paper. 
The metric aims to capture multiple aspects of complexity. 
The calculation involves determining individual component scores, combining them via a weighted sum, and normalizing the result.
The score function is described in \Cref{app:scoredifficulty} and its components in \Cref{app:componentscore}.

\subsection{Combination and Normalization}\label{app:scoredifficulty}
The individual component scores (\Cref{app:componentscore}) are combined using a weighted sum to produce a raw difficulty score ($S_{\text{raw}}$). 
The specific weights reflect the empirically determined relative importance of each component:
\begin{dmath*}
S_{\text{raw}} = w_{\text{mech}} S_{\text{mech}} + w_{\text{interact}} S_{\text{interact}} + w_{\text{grid}} S_{\text{grid}} + w_{\text{density}} S_{\text{density}} + w_{\text{count}} S_{\text{count}}
\end{dmath*}

where the weights used are: $w_{\text{mech}}=1.2$, $w_{\text{interact}}=1.2$, $w_{\text{grid}}=2.5$, $w_{\text{density}}=1.0$, and $w_{\text{count}}=1.2$. Notably, grid size ($S_{\text{grid}}$) is weighted most heavily.

Finally, to produce a standardized and interpretable difficulty score (typically ranging from 0 to 5), the raw score ($S_{\text{raw}}$) is normalized. 
This is achieved by:
\begin{enumerate}
    \item Calculating the Z-score of $S_{\text{raw}}$ relative to a pre-determined normal distribution, characterized by a mean ($\mu = 12.06$) and standard deviation ($\sigma = 5.27$). These parameters were derived empirically from a large dataset of puzzle scores.
    \[ Z = \frac{S_{\text{raw}} - \mu}{\sigma} \]
    \item Converting the Z-score to a value between 0 and 1 using the standard normal cumulative distribution function (CDF), often denoted as $\Phi(Z)$.
    \[ \text{CDF\_value} = \Phi(Z) \]
    \item Linearly scaling this CDF value to the target range [0, 5].
    \[ \text{Scaled\_Score} = \text{CDF\_value} \times 5 \]
    \item Clamping the result to ensure the final difficulty score strictly falls within the [0, 5] bounds.
    \[ \text{Final Score} = \max(0, \min(5, \text{Scaled\_Score})) \]
\end{enumerate}
This normalization process ensures that scores are comparable across different puzzles and provides a distribution more amenable to interpretation as a rating.

\subsection{Component Scores} \label{app:componentscore}
Five distinct aspects of the puzzle contribute to the overall difficulty score:

\begin{itemize}
    \item \textbf{Mechanics Score ($S_{\text{mech}}$):} This score reflects the cognitive load associated with understanding different rules. It is directly proportional to the number of unique rule types present in the puzzle ($N_{\text{mech}}$).
\item \textbf{Interaction Score ($S_{\text{interact}}$):} This score quantifies complexity from the interplay between different mechanics. It is calculated only when multiple rule types ($N_{\text{mech}} > 1$) are present. It is proportional to both the number of potentially interacting mechanics (approximated as $N_{\text{mech}} - 1$) and the rule density ($\rho_{\text{rules}}$), where rule density is the total number of rule instances ($N_{\text{rules}}$) divided by the grid area ($A = \text{width} \times \text{height}$).
    \item \textbf{Grid Score ($S_{\text{grid}}$):} This component reflects the complexity associated with the search space size. It increases proportionally with the grid area ($A$). Larger grids generally require more path exploration.
    \item \textbf{Density Score ($S_{\text{density}}$):} This score measures constraint concentration. It is directly derived from the rule density ($\rho_{\text{rules}} = N_{\text{rules}} / A$). Higher density can make satisfying all constraints simultaneously more challenging.
    \item \textbf{Rule Count Score ($S_{\text{count}}$):} Independent of density, this score considers the absolute number of constraints. It is proportional to the total number of rule instances ($N_{\text{rules}}$) on the grid. A puzzle with many rules can be complex even if spread over a large grid.
\end{itemize}

\section{Prompting and Representation}
\label{app:prompt_repre}

In addition to the main results, we also evaluated the effect of a different reasoning-oriented fine-tuning method and puzzle grid representation. 
These experiments provide insights into whether prompting style or input format can improve the bad spatial reasoning performance of LLMs observed on SPaRC.

\subsection{Reasoning Fine-Tuned Models}

\begin{figure}[h]
    \centering
    \includegraphics[width=\linewidth]{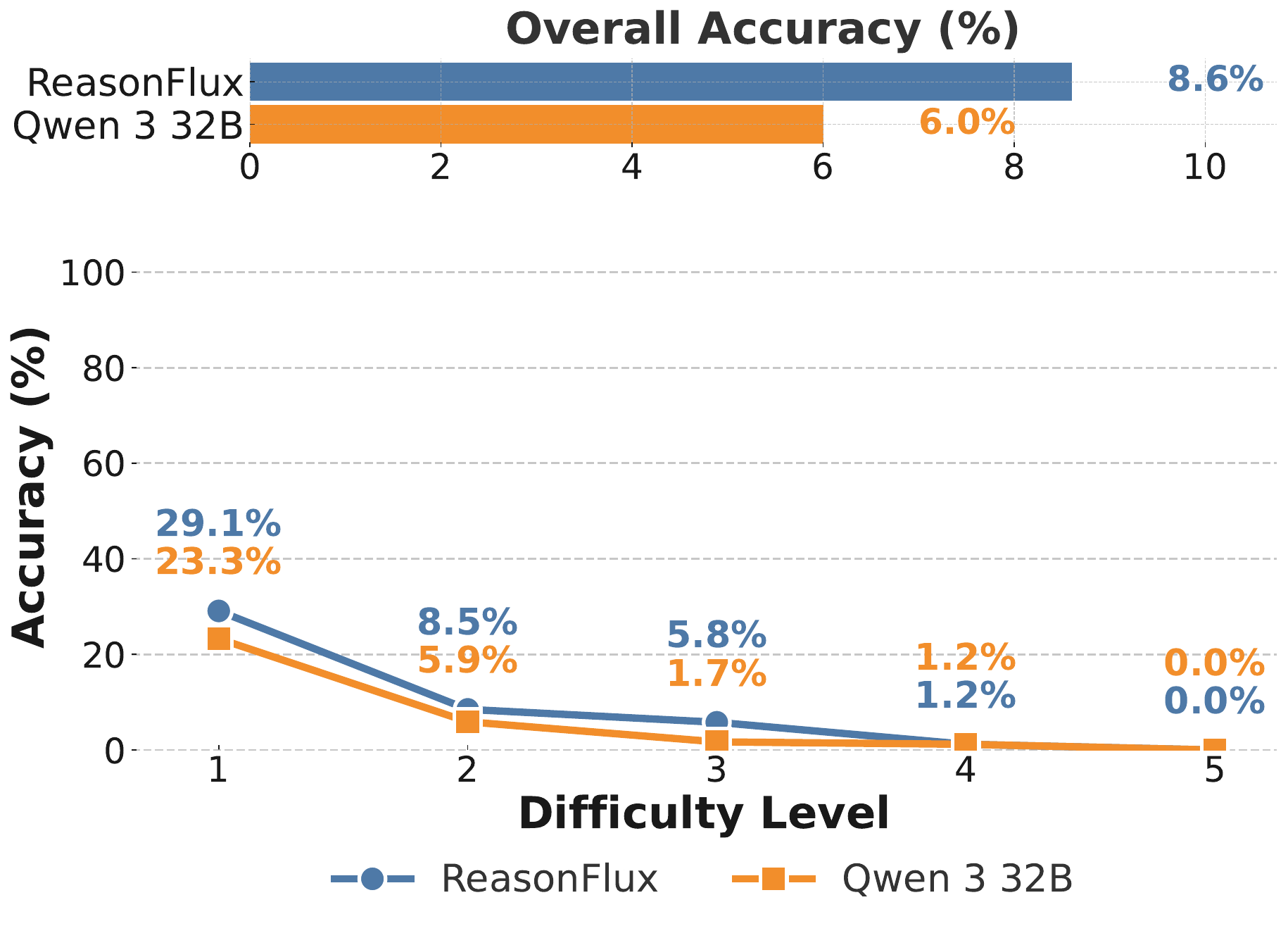}
    \caption{\textbf{Accuracy (\%)} of reasoning-tuned models (Qwen 3 32B vs.\ ReasonFlux variants) across difficulty levels (1–5). Higher is better.}
    \label{fig:prompting}
\end{figure}

While most existing prompting methods are designed for instruction-tuned models, we investigated the ReasonFlux family of reasoning fine-tuned models, which use a different reasoning method in the fine-tuning step compared to the Qwen 3 models and should perform better on long context tasks \citep{zou2025reasonfluxprm}. 
\Cref{fig:prompting} compares Qwen 3 32B against ReasonFlux-F1 32B.

We find that the ReasonFlux-F1 32B model performs comparably to another reasoning-tuned model of the same size (8.6\% vs.\ 6.0\%). 
These results show that the long-context specific reasoning fine-tuning can help, but gains remain small relative to human performance.

\subsection{Puzzle Grid Representations}

We also investigated the influence of puzzle grid representations. 
Besides the default ARC-AGI-inspired array representation (Array), we tested two markdown-based formats: a plain markdown table without headers (Table), and a table including row and column coordinates to assist with spatial referencing (Coords). 

\begin{figure}[h]
    \centering
    \includegraphics[width=\linewidth]{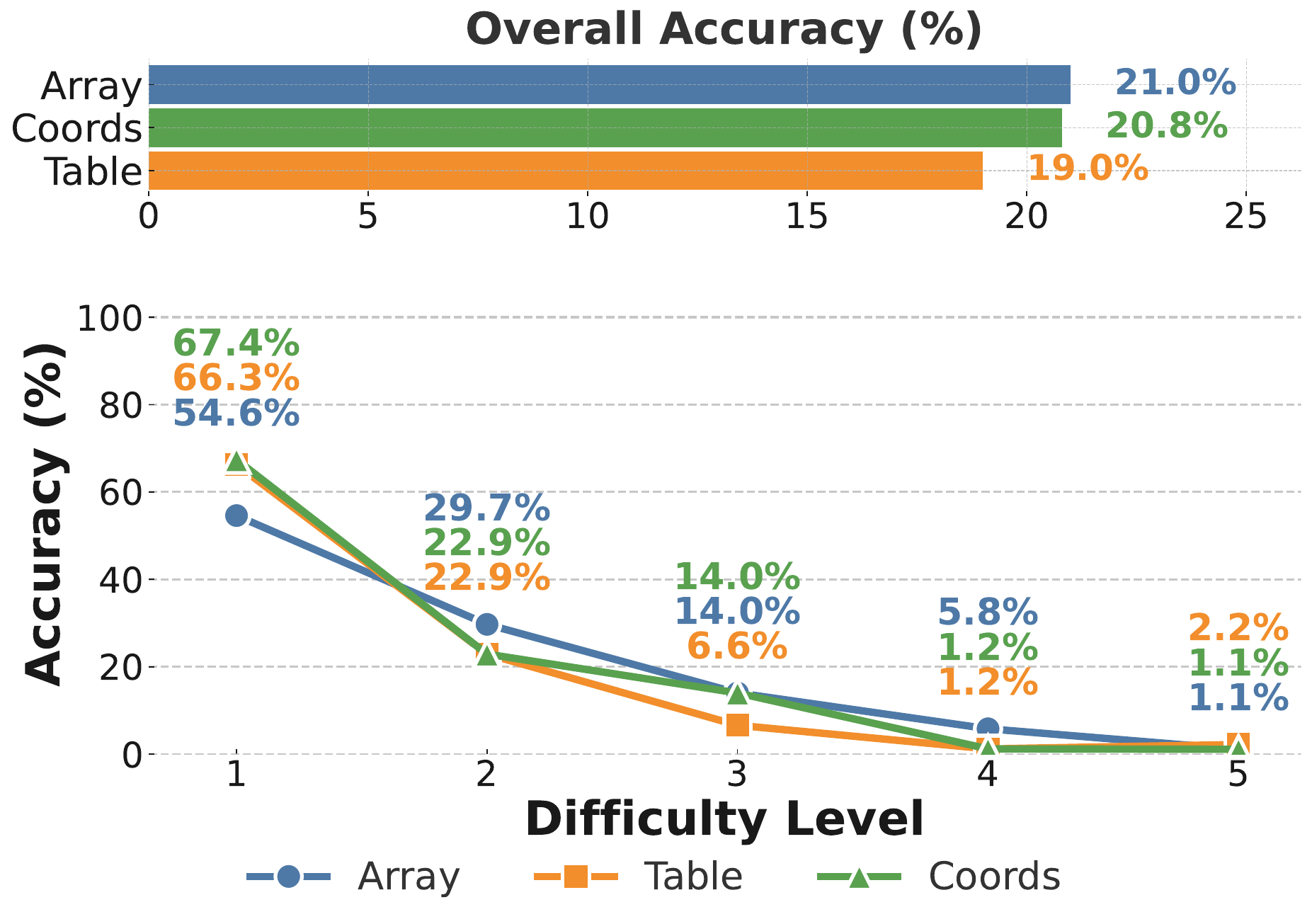}
    \caption{\textbf{Accuracy (\%)} of different puzzle grid representations (array vs.\ markdown table variants) across difficulty levels (1–5). Higher is better.}
    \label{fig:representation}
\end{figure}

As shown in \Cref{fig:representation}, neither markdown variant yielded consistent improvements. 
The baseline array format achieved the highest overall accuracy (21.0\%), slightly outperforming both markdown representations (19.0\% and 20.8\%). 
These results suggest that failures on SPaRC stem from fundamental limitations in spatial reasoning rather than representation format.

\section{Tokens by Puzzle Difficulty}
\label{app:tokens_by_difficulty}
\Cref{tab:tokens_by_difficulty} shows the average tokens produced by the different models, decomposed by puzzle difficulty.

\begin{table}[ht]
\centering

\resizebox{\linewidth}{!}{\begin{tabular}{lccccc}
\toprule
Model & Level 1 & Level 2 & Level 3 & Level 4 & Level 5 \\
\hline
\textbf{Reasoning} \\
\quad QwQ 32B & 14433.3 & 14200.6 & 13983.1 & 14072.8 & 13114.1 \\
\quad R1 70B & 7646.5 & 9119.8 & 9374.6 & 10134.4 & 9989.6 \\
\textbf{Instruction} \\
\quad Qwen 2.5 72B & 790.6 & 888.7 & 953.1 & 1037.7 & 1161.2 \\
\quad Gemma-3 27B & 802.8 & 874.6 & 910.0 & 941.2 & 1033.3 \\
\bottomrule
\end{tabular}}
\caption{Average tokens per puzzle by difficulty level.}
\label{tab:tokens_by_difficulty}
\end{table}

\section{Rule Visualizations}
\label{sec:puzzle_subtypes_visual}

\begin{figure}[H] 
    \centering
    \begin{subfigure}[b]{0.48\textwidth}
        \centering
        \includegraphics[width=0.8\linewidth]{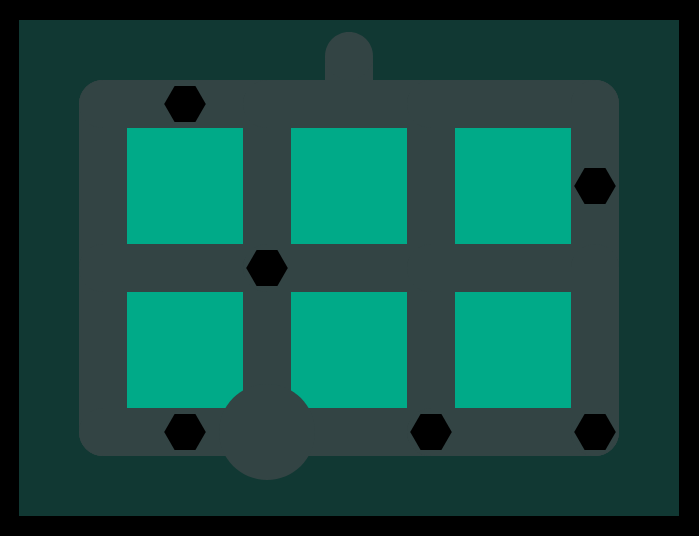} %
        \caption{Unsolved Dots puzzle.}
        \label{fig:unsolved_dots}
    \end{subfigure}
    \hfill %
    \begin{subfigure}[b]{0.48\textwidth}
        \centering
        \includegraphics[width=0.8\linewidth]{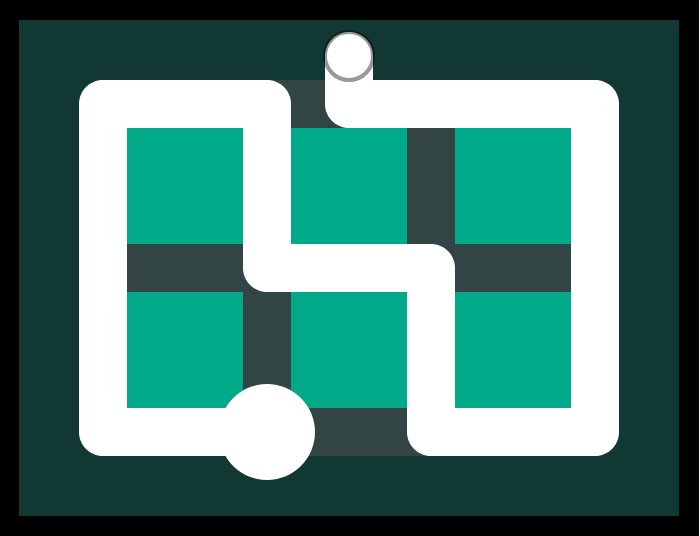} %
        \caption{Solved Dots puzzle.}
        \label{fig:solved_dots}
    \end{subfigure}
    \caption{Example of the \textit{Dots} rule. The solution path must pass through all dots present on its segments.}
    \label{fig:dots_example}
\end{figure}

\begin{figure}[H] 
    \centering
    \begin{subfigure}[b]{0.48\textwidth}
        \centering
        \includegraphics[width=0.8\linewidth]{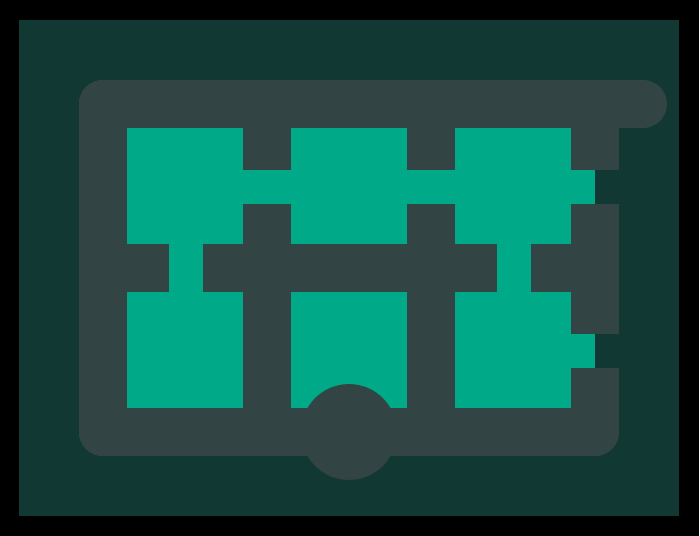} %
        \caption{Unsolved Gaps puzzle.}
        \label{fig:unsolved_gaps}
    \end{subfigure}
    \hfill
    \begin{subfigure}[b]{0.48\textwidth}
        \centering
        \includegraphics[width=0.8\linewidth]{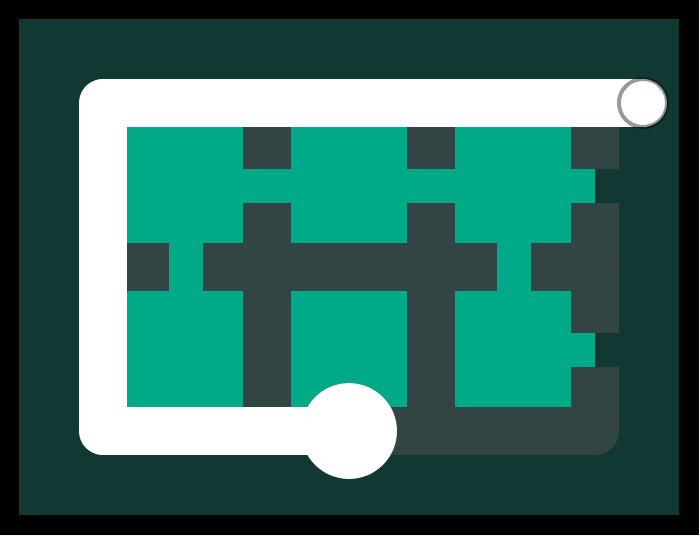} %
        \caption{Solved Gaps puzzle.}
        \label{fig:solved_gaps}
    \end{subfigure}
    \caption{Example of the \textit{Gaps} rule. The solution path cannot cross specific marked edges on the grid.}
    \label{fig:gaps_example}
\end{figure}

\begin{figure}[H] 
    \centering
    \begin{subfigure}[b]{0.48\textwidth}
        \centering
        \includegraphics[width=0.8\linewidth]{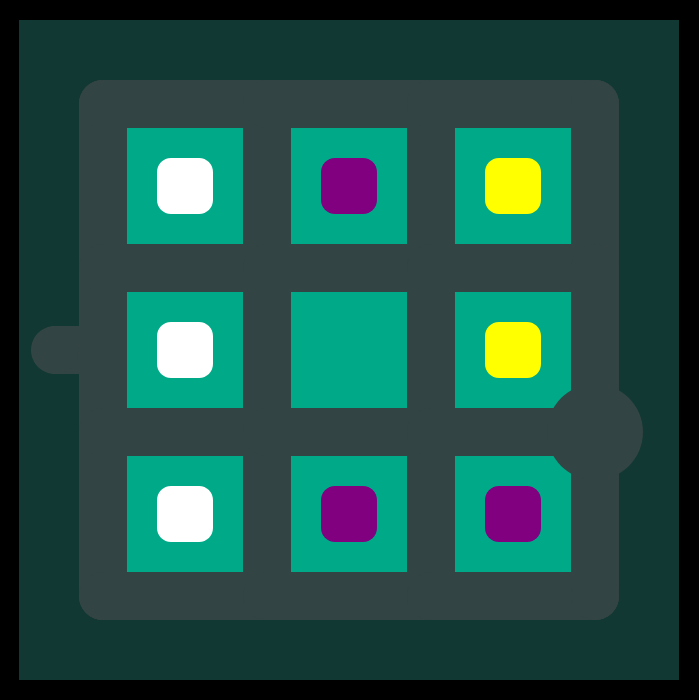} %
        \caption{Unsolved Stones puzzle.}
        \label{fig:unsolved_stones}
    \end{subfigure}
    \hfill
    \begin{subfigure}[b]{0.48\textwidth}
        \centering
        \includegraphics[width=0.8\linewidth]{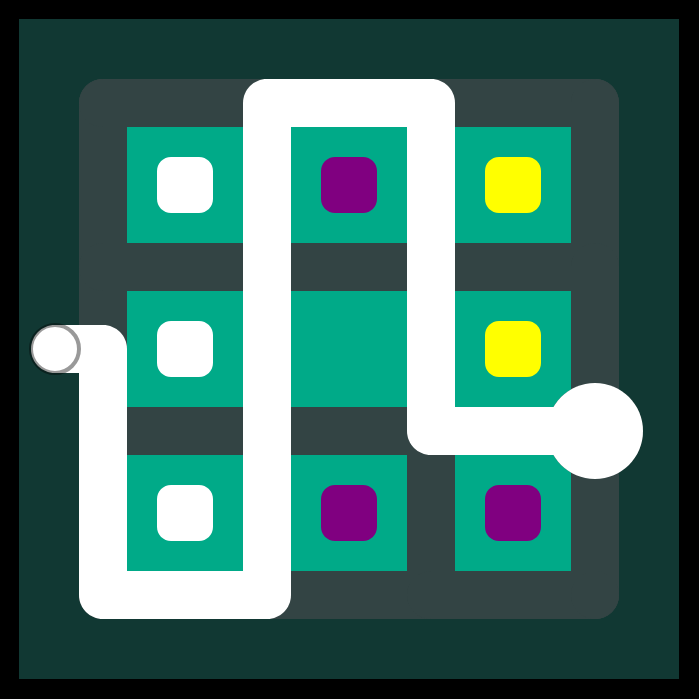} %
        \caption{Solved Stones puzzle.}
        \label{fig:solved_stones}
    \end{subfigure}
    \caption{Example of the \textit{Stones} rule. The solution path must separate grid cells containing different colored stones into distinct regions.}
    \label{fig:stones_example}
\end{figure}

\begin{figure}[H] 
    \centering
    \begin{subfigure}[b]{0.48\textwidth}
        \centering
        \includegraphics[width=0.8\linewidth]{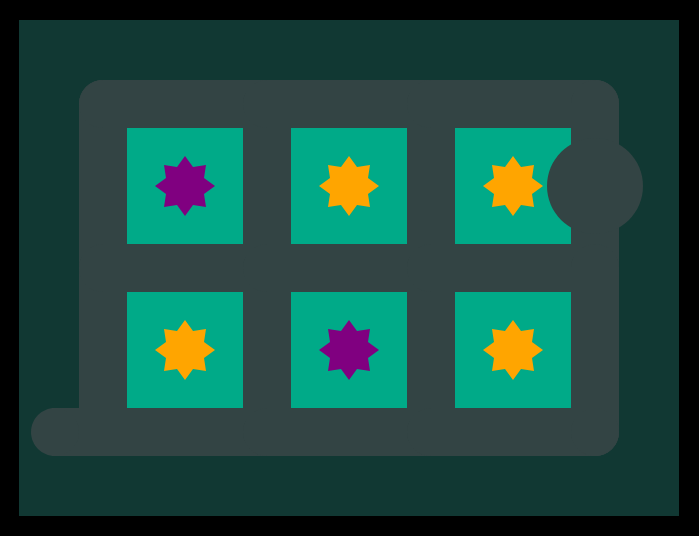} %
        \caption{Unsolved Stars puzzle.}
        \label{fig:unsolved_stars}
    \end{subfigure}
    \hfill
    \begin{subfigure}[b]{0.48\textwidth}
        \centering
        \includegraphics[width=0.8\linewidth]{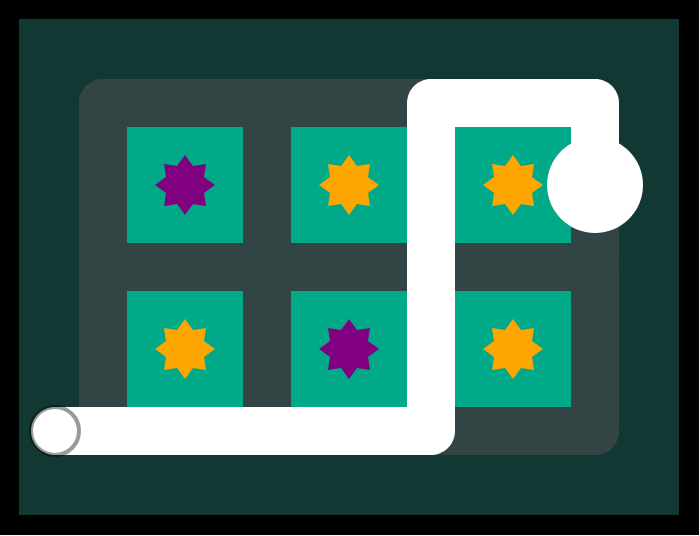} %
        \caption{Solved Stars puzzle.}
        \label{fig:solved_stars}
    \end{subfigure}
    \caption{Example of the \textit{Stars} rule. Each region with a star must contain exactly one other rule of the same color.}
    \label{fig:stars_example}
\end{figure}

\begin{figure}[H] 
    \centering
    \begin{subfigure}[b]{0.48\textwidth}
        \centering
        \includegraphics[width=0.8\linewidth]{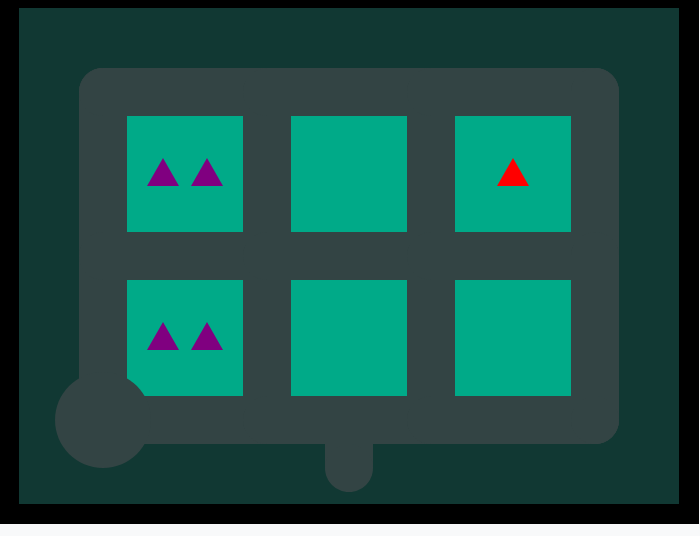} %
        \caption{Unsolved Triangles puzzle.}
        \label{fig:unsolved_triangles}
    \end{subfigure}
    \hfill
    \begin{subfigure}[b]{0.48\textwidth}
        \centering
        \includegraphics[width=0.8\linewidth]{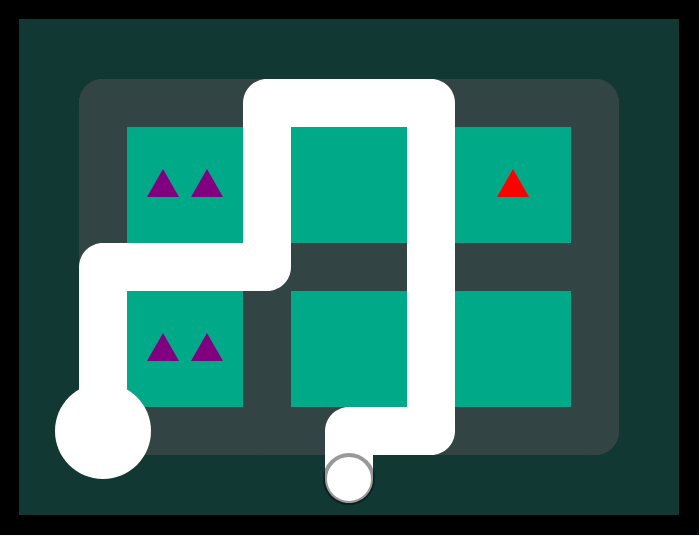} %
        \caption{Solved Triangles puzzle.}
        \label{fig:solved_triangles}
    \end{subfigure}
    \caption{Example of the \textit{Triangles} rule. The solution path must touch the number of grid edges equal to the number of triangles in the adjacent cell.}
    \label{fig:triangles_example}
\end{figure}

\begin{figure}[H] 
    \centering
    \begin{subfigure}[b]{0.48\textwidth}
        \centering
        \includegraphics[width=0.8\linewidth]{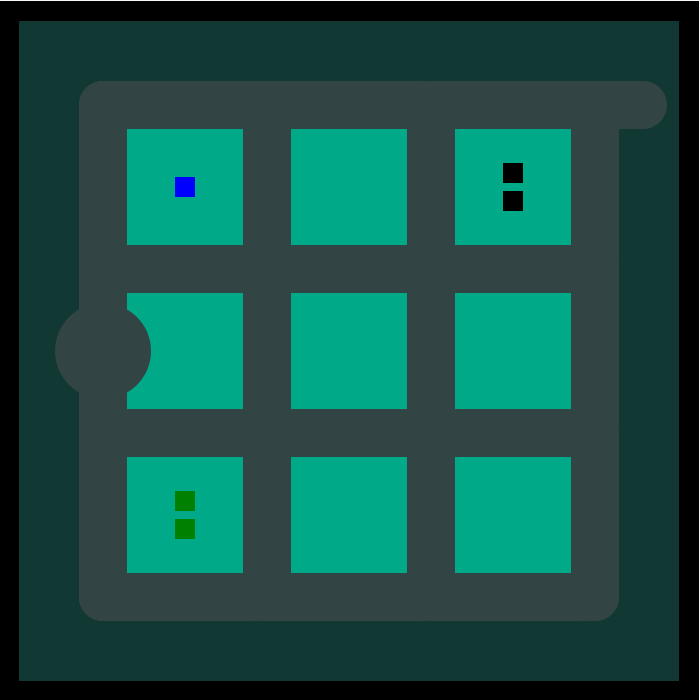} %
        \caption{Unsolved Polys puzzle.}
        \label{fig:unsolved_polys}
    \end{subfigure}
    \hfill
    \begin{subfigure}[b]{0.48\textwidth}
        \centering
        \includegraphics[width=0.8\linewidth]{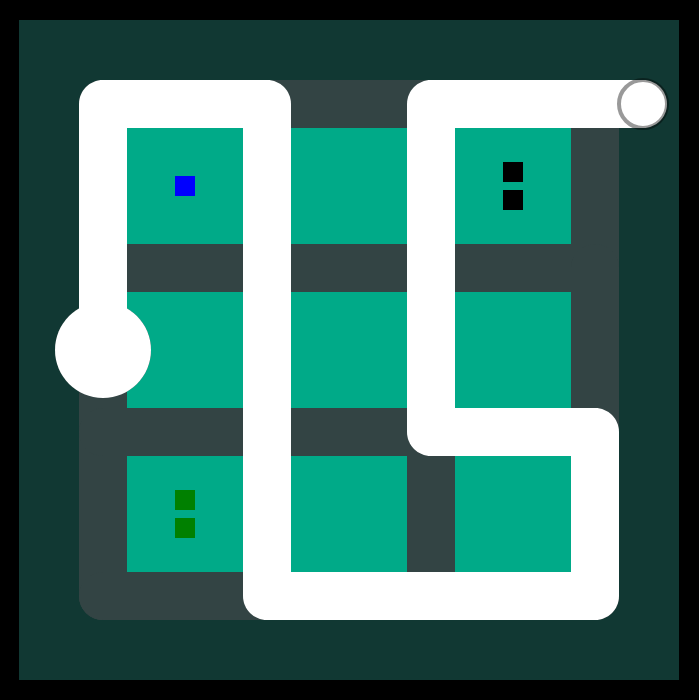} %
        \caption{Solved Polys puzzle.}
        \label{fig:solved_polys}
    \end{subfigure}
    \caption{Example of the \textit{Polys} rule (Polyominoes). The solution path must outline a region that perfectly contains the depicted poly shape. Multiple polys in one region can be combined.}
    \label{fig:polys_example}
\end{figure}

\begin{figure}[H] 
    \centering
    \begin{subfigure}[b]{0.48\textwidth}
        \centering
        \includegraphics[width=0.8\linewidth]{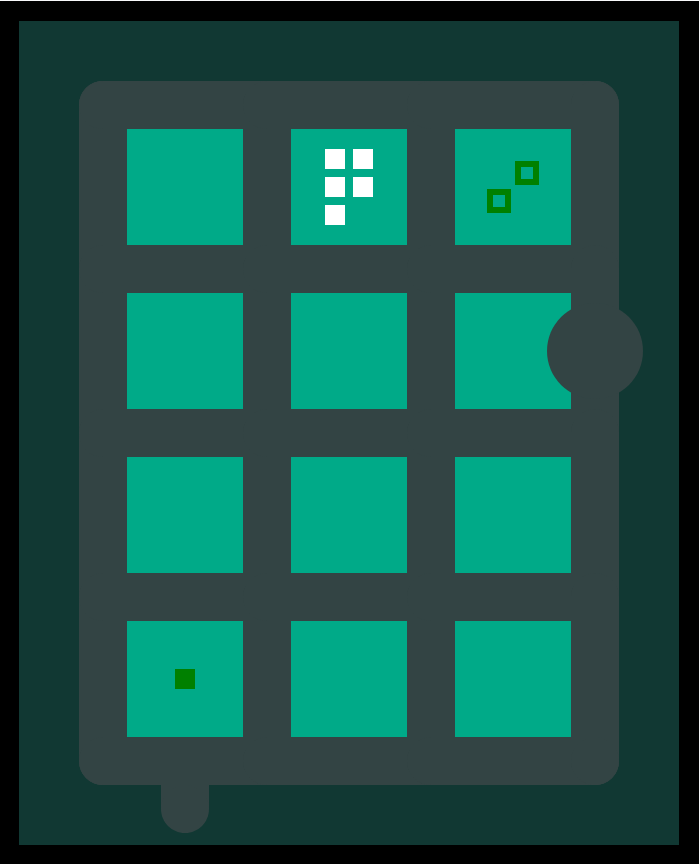} %
        \caption{Unsolved Polys \& Inverse Polys puzzle.}
        \label{fig:unsolved_polys_ylops}
    \end{subfigure}
    \hfill
    \begin{subfigure}[b]{0.48\textwidth}
        \centering
        \includegraphics[width=0.8\linewidth]{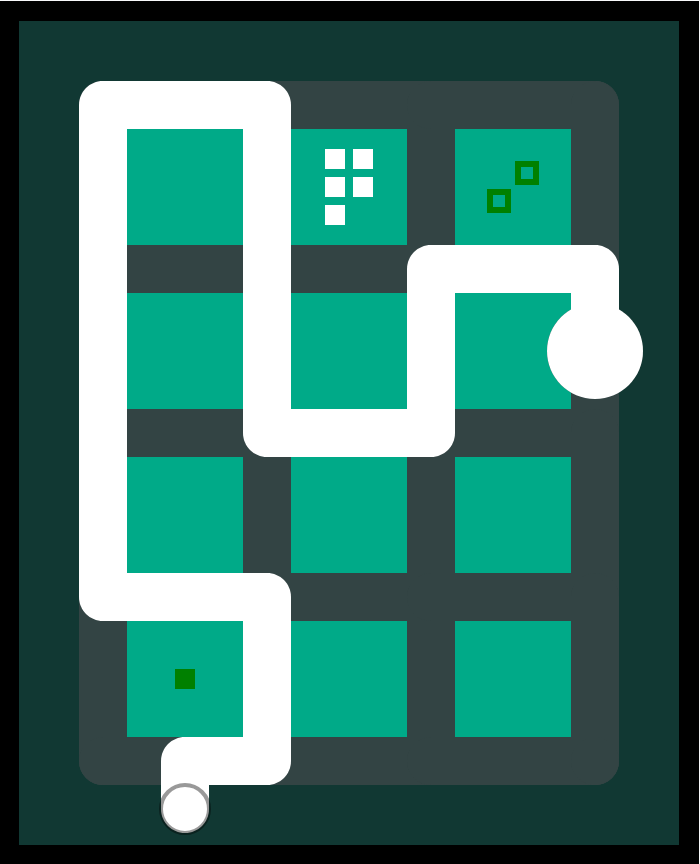} %
        \caption{Solved Polys \& Inverse Polys puzzle.}
        \label{fig:solved_polys_ylops}
    \end{subfigure}
    \caption{Example of the \textit{Polys \& Ylops} (Inverse Polys) rule combination. The solution path must outline a region satisfying both polyomino shape inclusion and subtraction constraints.}
    \label{fig:polys_ylops_example}
\end{figure}

\Cref{fig:dots_example,fig:gaps_example,fig:stones_example,fig:stars_example,fig:triangles_example,fig:polys_example,fig:polys_ylops_example} provide visual context for the different puzzle rule types discussed in our evaluation (\Cref{sec:evaluation}), this section presents examples of each core subtype. 
For each rule, we show the unsolved puzzle grid (a) with a valid solution path (b).
\onecolumn

\section{Additional Dataset Statistics}
\Cref{tab:dataset_summary_sections_subsets} provides the rule distributions of the full set of \dataset\ and all of its splits.

\begin{table}[H]
\centering
\resizebox{0.99\textwidth}{!}{%
\begin{tabular}{lrrrrrrrrrrr}
\toprule
 Statistics & Full Set & Gaps & Dots & Stones & Stars & Tri & Polys & P-Y & St-S & G-D-T & D-S-P \\
\midrule
Train Set Size & 500 & 50 & 50 & 50 & 50 & 50 & 50 & 50 & 50 & 50 & 50 \\
Test Set Size & 500 & 50 & 50 & 50 & 50 & 50 & 50 & 50 & 50 & 50 & 50 \\
\multicolumn{12}{l}{\textbf{Count per Difficulty Level}} \\
\quad Puzzles (Level 1) & 86 & 34 & 29 & 0 & 0 & 13 & 0 & 3 & 0 & 21 & 7 \\
\quad Puzzles (Level 2) & 118 & 6 & 9 & 5 & 29 & 17 & 24 & 4 & 12 & 13 & 9 \\
\quad Puzzles (Level 3) & 121 & 7 & 7 & 13 & 15 & 12 & 22 & 9 & 13 & 6 & 7 \\
\quad Puzzles (Level 4) & 86 & 3 & 3 & 18 & 5 & 4 & 4 & 18 & 12 & 4 & 9 \\
\quad Puzzles (Level 5) & 89 & 0 & 2 & 14 & 1 & 4 & 0 & 16 & 13 & 6 & 18 \\
\multicolumn{12}{l}{\textbf{Count per Rule Type}} \\
\quad Puzzles with Gaps & 313 & 50 & 0 & 0 & 0 & 0 & 0 & 0 & 0 & 40 & 0 \\
\quad Puzzles with Dots & 292 & 0 & 50 & 0 & 0 & 0 & 0 & 0 & 0 & 47 & 46 \\
\quad Puzzles with Stones & 355 & 0 & 0 & 50 & 0 & 0 & 0 & 0 & 49 & 0 & 0 \\
\quad Puzzles with Stars & 210 & 0 & 0 & 0 & 50 & 0 & 0 & 0 & 32 & 0 & 39 \\
\quad Puzzles with Triangles & 233 & 0 & 0 & 0 & 0 & 50 & 0 & 0 & 0 & 37 & 0 \\
\quad Puzzles with Polygons & 305 & 0 & 0 & 0 & 0 & 0 & 50 & 50 & 0 & 0 & 43 \\
\quad Puzzles with Ylops & 25 & 0 & 0 & 0 & 0 & 0 & 0 & 43 & 0 & 0 & 0 \\
\bottomrule
\end{tabular}}
\caption{Statistics for all splits of \dataset. Difficulty and rule statistics are only based on the test set, as only these are used for evaluation.}
\label{tab:dataset_summary_sections_subsets}
\end{table}

\section{Prompting}
\label{app:prompts}
\Cref{lst:normal_prompt,lst:alt_prompt,lst:vision_prompt,lst:one_shot,lst:two_shot} in \Cref{app:prompts_normal,app:prompts_alt,app:prompts_vision,app:prompts_examples} provide the prompts and few-shot examples used for the experiments in \Cref{sec:evaluation}.

\definecolor{promptbackground}{HTML}{F5F5F5} %
\definecolor{promptframe}{HTML}{CCCCCC}       %
\definecolor{prompttext}{HTML}{333333}        %

\lstdefinestyle{Prompt}{
    backgroundcolor=\color{promptbackground}, %
    basicstyle=\small\ttfamily\color{prompttext}, %
    keywordstyle=\color{blue},         %
    stringstyle=\color{purple},       %
    commentstyle=\color{green!60!black}, %
    numbers=none,                      %
    numberstyle=\tiny\color{promptframe}, %
    numbersep=0pt,                    %
    breaklines=true,                   %
    tabsize=4,                         %
    frame=single,                      %
    rulecolor=\color{promptframe},     %
    captionpos=b,                      %
    showstringspaces=false,            %
    extendedchars=true                %
}

\subsection{Default Prompt}
\label{app:prompts_normal}
\begin{lstlisting}[style=Prompt, caption={The LLM prompt used for generating the results discussed in \Cref{sec:main_results}.}, label={lst:normal_prompt}]
You are an expert spatial reasoning AI specializing in solving puzzles from the game 'The Witness'. Your task is to solve the following puzzle by finding a valid line from the Start Node to the End Node.

GRID DEFINITION:
- The puzzle involves a grid of {grid_size['width']}x{grid_size['height']} cells.
- COORDINATE SYSTEM: Nodes are indexed (x, y). Node (0,0) is the top-left node. x increases to the right, y increases downward.
- Line: The solution line travels along grid edges, connecting adjacent nodes horizontally or vertically. The line cannot visit the same node twice.
- RULE PLACEMENT: Rule symbols (squares, stars, polyshapes, negative polyshapes, triangles) are located at cells with all odd coordinates. The line goes AROUND cells containing rules, forming boundaries.

SOLVING RULES:
1.  Draw a continuous line from the START NODE to the END NODE by connecting adjacent nodes (horizontally or vertically) without visiting the same node twice.
2.  The line can only be placed on (+) and (.) cells. These cells have at least one even coordinate. The line can NEVER be placed on a rule cell (all odd coordinates).
3.  The line acts as a boundary, potentially dividing the grid cells into one or more distinct regions.
4.  All rules associated with symbols on the grid must be satisfied:
    - Gaps ('G'): The line CANNOT traverse a cell marked by a Gap.
    - Dots ('.'): The line MUST pass through a cell marked by a Dot.
    - Squares ('o-X'): All squares within a single region created by the line must be the same color. Different colored squares MUST be separated into different regions by the line.
    - Stars ('*-X'): Each star must be paired with EXACTLY one other element of the same color in a region. Other colors are ignored.
    - Triangles ('A-X (1)', 'B-X (2)', 'C-X (3)', 'D-X (4)'): The line must touch EXACTLY the number of edges specified by the triangle count (edges are top, right, bottom, left of the cell).
    - Polyshapes ('P-X-Y'): The region containing this symbol must be shaped EXACTLY like the defined polyshape Y. The shape must fit entirely within the region's boundaries. If multiple positive polyshapes are in one region, the region's shape must accommodate their combined, non-overlapping forms (like Tetris pieces).
    - Negative Polyshapes ('Y-X-Y'): The negative polyshape can only be placed on top of already placed normal polyshapes. The negative polyshapes must fit on the grid, but can allow overlap between normal polyshapes or placement of polyshapes that extend beyond the area defined by the line. If the negative polyshapes exactly cancel the normal polyshapes, there is no restriction on the grid shape anymore. A negative polyshape only counts as valid if it is used.


START POSITION: {start_pos}
END POSITION: {end_pos}

GRID NOTATION:
- 'S': Start point
- 'E': End point
- '+': Cell on which the line can be drawn
- 'N': Empty rule cell
- 'G': Gap (cannot be crossed)
- '.': Dot line must cross this cell
- 'o-X': Stone of color X
- '*-X': Star of color X
- 'A-X' Triangle with count 1
- 'B-X' Triangle with count 2
- 'C-X' Triangle with count 3
- 'D-X' Triangle with count 4
- 'P-X-Y': Positive polyshape of color X and shape ID Y
- 'Y-X-Y': Negative polyshape (ylop) of color X and shape ID Y

COLOR CODES:
R=red, B=blue, G=green, Y=yellow, W=white, O=orange, P=purple, K=black

{example_section}

PUZZLE GRID:
{grid_str}

POLYSHAPE DEFINITIONS:
Defines the shapes referenced by P-X-Y and Y-X-Y symbols in the grid.
In the 2D array, 1 indicates a cell occupied by the shape, 0 indicates an empty cell.
{polyshapes_str}

Please solve this puzzle.
First, explain your reasoning step-by-step, including key deductions and constraint checks made along the way.
Then, provide the final solution as a sequence of node coordinates in (x, y) format (dont skip any intermediate nodes), starting with the start node and ending with the end node, after this string: "####".
Example coordinate list: [(0,0), (1,0), (2,0), (2,1), ...]
\end{lstlisting}
\subsection{Alternative Prompt}
\label{app:prompts_alt}
\begin{lstlisting}[style=Prompt, caption={The LLM prompt used for generating the results discussed in prompt ablation in \Cref{sec:ablations}.}, label={lst:alt_prompt}]
## Objective
You are a specialized AI proficient in spatial reasoning and solving puzzles from the game 'The Witness'. Your goal is to find a valid path (a continuous line) from the specified Start Node to the End Node on the provided grid, adhering to all puzzle rules.

## Core Concepts & Grid Basics
*   **Grid Dimensions:** The puzzle grid has {grid_size['width']} columns and {grid_size['height']} rows.
*   **Coordinate System:** Nodes are identified by `(x, y)` coordinates. `(0,0)` is the top-left node. `x` increases to the right, `y` increases downwards.
*   **Path:** The solution is a single, continuous line connecting adjacent nodes either horizontally or vertically.
*   **No Revisits:** The path **CANNOT** visit the same node more than once.
*   **Valid Path Cells:** The path travels along the grid lines (edges between nodes). It can only occupy positions marked `+` or `.` in the grid layout (these correspond to positions with at least one even coordinate).
*   **Rule Cells:** Cells containing rule symbols (squares, stars, etc.) have coordinates where both `x` and `y` are odd. The path goes *around* these rule cells, never *on* them.
*   **Regions:** The drawn path divides the grid cells into one or more distinct enclosed areas (regions). Many rules apply based on the contents of these regions.

## Puzzle Input Data
*   **Start Node:** {start_pos}
*   **End Node:** {end_pos}
*   **Grid Layout:**
    ```
    {grid_str}
    ```
*   **Polyshape Definitions (if applicable):**
    *   Shapes are defined by 2D arrays where '1' indicates an occupied cell and '0' indicates an empty cell.
    ```
    {polyshapes_str}
    ```

## Symbol Legend (Grid Notation)
*   `S`: **Start Node** (Path begins here)
*   `E`: **End Node** (Path ends here)
*   `+`: Valid cell for the path to occupy
*   `N`: Empty rule cell (no rule)
*   `G`: **Gap** (Path **CANNOT** cross this cell)
*   `.`: **Dot** (Path **MUST** pass through this cell)
*   `o-X`: **Square** of color X
*   `*-X`: **Star** of color X
*   `A-X`: **Triangle** (touch 1 edge)
*   `B-X`: **Triangle** (touch 2 edges)
*   `C-X`: **Triangle** (touch 3 edges)
*   `D-X`: **Triangle** (touch 4 edges)
*   `P-X-Y`: **Polyshape** (positive) of color X and shape ID Y
*   `Y-X-Y`: **Negative Polyshape** (ylop) of color X and shape ID Y

**Color Codes:** R=Red, B=Blue, G=Green, Y=Yellow, W=White, O=Orange, P=Purple, K=Black

## Detailed Solving Rules
The drawn path must satisfy **ALL** applicable constraints:

1.  **Path Constraints:**
    *   Path **MUST** start at `S` and end at `E`.
    *   Path connects adjacent nodes (horizontal/vertical moves only).
    *   Nodes **CANNOT** be revisited.
    *   Path **MUST** pass through all Dot (`.`) cells.
    *   Path **CANNOT** pass through any Gap (`G`) cells.

2.  **Region-Based Rules** (Apply to areas enclosed by the path):
    *   **Squares (`o-X`):** All squares within a single region **MUST** be the same color. Squares of different colors **MUST** be separated into different regions by the path.
    *   **Stars (`*-X`):** Within a single region, each star symbol **MUST** be paired with exactly **ONE** other element (star or square) *of the same color*. Other colors within the region are irrelevant to this specific star's rule.
    *   **Polyshapes (`P-X-Y`):** The region containing this symbol **MUST** be able to contain the specified shape (defined in Polyshape Definitions). The shape must fit entirely within the region's boundaries. If multiple positive polyshapes are in one region, the region must accommodate their combined, non-overlapping forms. Rotation of polyshapes is generally allowed unless context implies otherwise.
    *   **Negative Polyshapes (`Y-X-Y`):** These "subtract" shape requirements, typically within the same region as corresponding positive polyshapes. A negative polyshape cancels out a positive polyshape of the exact same shape and color within that region. If all positive shapes are canceled, the region has no shape constraint. A negative shape is only considered 'used' if it cancels a positive one. Negative shapes can sometimes rationalize apparent overlaps or boundary violations of positive shapes if interpreted as cancellations.

3.  **Path-Based Rules (Edge Touching):**
    *   **Triangles (`A-X`, `B-X`, `C-X`, `D-X`):** The path **MUST** touch a specific number of edges of the cell containing the triangle symbol.
        *   `A-X` (1): Path touches **EXACTLY 1** edge of the triangle's cell.
        *   `B-X` (2): Path touches **EXACTLY 2** edges of the triangle's cell.
        *   `C-X` (3): Path touches **EXACTLY 3** edges of the triangle's cell.
        *   `D-X` (4): Path touches **EXACTLY 4** edges (fully surrounds) the triangle's cell.

{example_section}

## Task & Output Format
1.  **Solve the Puzzle:** Determine the valid path from the Start Node to the End Node that satisfies all rules.
2.  **Explain Reasoning:** Provide a step-by-step explanation of your thought process. Detail key deductions, how constraints were applied, and any backtracking or choices made.
3.  **Provide Solution Path:** After the reasoning, output the exact marker string `####` followed immediately by the solution path as a list of node coordinates `(x, y)`. Include all intermediate nodes from start to end.

**Example Solution Path Format:**
####
[(0, 0), (1, 0), (2, 0), (2, 1), ...]
\end{lstlisting}
\subsection{Vision Prompt}
\label{app:prompts_vision}
\begin{lstlisting}[style=Prompt, caption={The LLM prompt used for generating the results discussed in vision ablation in \Cref{sec:ablations}.}, label={lst:vision_prompt}]
You are an expert spatial reasoning AI specializing in solving puzzles from the game 'The Witness'. 
Your task is to solve the puzzle in the image by finding a valid line from the Start Node to the End Node.

The image shows a Witness puzzle grid of size {grid_size['width']*2}x{grid_size['height']*2}. In this puzzle:
- The solution is a continuous line from the start circle to the end marker
- The line travels along grid edges, connecting adjacent nodes horizontally or vertically
- The line cannot visit the same node twice
- The line must satisfy all constraints represented by the symbols on the grid
- The line can not be placed on rule cells
- The line can only travel 1 cell per step (no diagonal moves and provide each step as a separate coordinate)

COORDINATE SYSTEM: 
- Nodes are indexed (x, y) where (0,0) is the top-left node
- x increases to the right, y increases downward
- The grid cells have rule symbols located at cells with all odd coordinates
- The line goes AROUND cells containing rules, forming boundaries
- Both line and rule cells are on the same grid. Therefore each intersection has a distance of 2 to the next intersection.

SOLVING RULES:
1. Draw a continuous line from the START NODE (big circle on the line) to the END NODE (rounded end) without visiting the same node twice.
2. The line can only be placed on valid path cells.
3. The line acts as a boundary, potentially dividing the grid cells into one or more distinct regions.
4. All rules associated with symbols on the grid must be satisfied:
   - Dots: The line MUST pass through each dot.
   - Colored squares: All squares within a single region created by the line must be the same color. Different colored squares MUST be separated into different regions by the line.
   - Colored stars: Each star must be paired with EXACTLY one other element of the same color in a region. Other colors are ignored.
   - Triangles: The line must touch EXACTLY the number of edges specified by the number of triangles in that cell (edges are top, right, bottom, left of the cell).
   - Tetris-like polyomino shapes: The region containing this symbol must be shaped EXACTLY like the defined polyshape.
   - Negative polyshapes: These cancel out regular polyshapes if they overlap.

Text description of the puzzle:
{puzzle_data.get("text_visualization", "")}

Analyze the puzzle image carefully and determine the solution path.
First, explain your reasoning step-by-step, including key deductions and constraint checks made along the way.
Then, provide the final solution as a sequence of node coordinates in (x, y) format, starting with the start node and ending with the end node, after this string: "####".. DON'T SKIP ANY intermediate nodes (the distance between each node must be 1).
Example coordinate list: [(0,0), (1,0), (2,0), (2,1), ...]
\end{lstlisting}
\subsection{Few-Shot Example}
\label{app:prompts_examples}
\begin{lstlisting}[style=Prompt, caption={The examples used for generating the results discussed in few-shot ablation in \Cref{sec:ablations}.}, label={lst:one_shot}]
EXAMPLE PUZZLE GRID:

["+",".","+","+","+","E","+"]
["+","C-R","+","o-K","+","o-K","+"]
["S","+","+","+","+","+","+"]
["+","P-G-112","+","*-G","+","P-B-624","+"]
["+","+","+","+","+","+","+"]
["+","*-G","+","*-G","+","o-K","+"]
["+","+","+",".","+","+","+"]

EXAMPLE POLYSHAPE DEFINITIONS:
Shape 112:
[0,1,0,0]
[0,1,0,0]
[0,1,0,0]
[0,0,0,0]

Shape 624:
[0,1,0,0]
[0,1,1,0]
[0,1,0,0]
[0,0,0,0]

EXAMPLE SOLUTION:

We start at (0,2) and draw a line to (0,0).
We then draw a line to (2,0) to reach the dot at (1,0) and surround the 3 count triangle.
We then draw a line to (2,2) here we go down to touch the third side of the triangle cell and therefore validate the 3 count triangle.
We continue down to (2,6) to validate the polyshape 112 and also the green star with the green polyshape
After this we draw a line to (4,6) to start validating the polyshape 624 by surrounding it.
Therefore we have to draw a line to (6,4) over (4,4) which creates a region for the stone at (5,5) which validates the stone.
We continue up to (6,2) for the polyshape 624 and then go to (4,2) and after this to (4,0) to finaly validate the polyshape 624.
This also validates the two green stars at (3,3) and (3,5) with each other and the black stone at (3,1) because its the only stone in its region.
This line also creates a region for the black stone at (5,1) because its the only stone in its region.
Now we can draw a line to (5,0) to reach the end node.

#### (0,2),(0,1),(0,0),(1,0),(2,0),(2,1),(2,2),(2,3),(2,4),(2,5),(2,6),(3,6),(4,6),(4,5),(4,4),(5,4),(6,4),(6,3),(6,2),(5,2),(4,2),(4,1),(4,0),(5,0)

\end{lstlisting}
\begin{lstlisting}[style=Prompt, caption={The examples used for generating the results discussed in few-shot ablation in \Cref{sec:ablations}.}, label={lst:two_shot}]
SECOND EXAMPLE PUZZLE GRID:
["+","E","+","+","+","+","+","+","+"]
["+","N","+","N","+","o-B","+","N","S"]
["+","+","+","+","+","+","+","+","+"]
["+","P-W-8992","G","Y-W-18","+","P-W-48","+","P-W-48","+"]
["+","+","+","G","+","+","+","+","+"]

SECOND EXAMPLE POLYSHAPE DEFINITIONS:
Shape 18:
[0,1,0,0]
[1,0,0,0]
[0,0,0,0]
[0,0,0,0]

Shape 48:
[0,1,0,0]
[0,1,0,0]
[0,0,0,0]
[0,0,0,0]

Shape 8992:
[0,0,1,0]
[0,1,1,1]
[0,0,0,0]
[0,0,0,0]

SECOND EXAMPLE SOLUTION:

We start at (8,1) and draw a line to (8,2).
Then we draw a straight line to (4,2).
From here we go up to (4,0).
This creates one region with only a blue stone at (5,1) which makes it valid.
The other region contains numerus polyshapes and ylops. But the region already has a valid shape.
The P-W-8992 gets placed on the bottom left and combined with the Y-W-18 to form a 2x1 region.
The other part of the region can exactly be formed by the two P-W-48 polyshapes.
Now we can draw a line to (1,0) to reach the end node.

#### (8,1),(8,2),(7,2),(6,2),(5,2),(4,2),(4,1),(4,0),(3,0),(2,0),(1,0)
\end{lstlisting}

\section{Full Tabular Main Results}
\Cref{tab:llm_solve_rate_difficulty_breakdown,tab:llm_path_violation_rates,tab:llm_subset_solve_rates} provide the detailed and complete results for the experiments in \Cref{sec:main_results}.

\label{app:detailed_results}
\subsection{Difficulty per Level}
\begin{table}[H]
\centering
\small
\begin{tabular}{lllllll}
\toprule
Model & All & Level 1 & Level 2 & Level 3 & Level 4 & Level 5 \\
\midrule
\textbf{Reasoning} &  &  &  &  &  &  \\
\quad \includegraphics[width=8px]{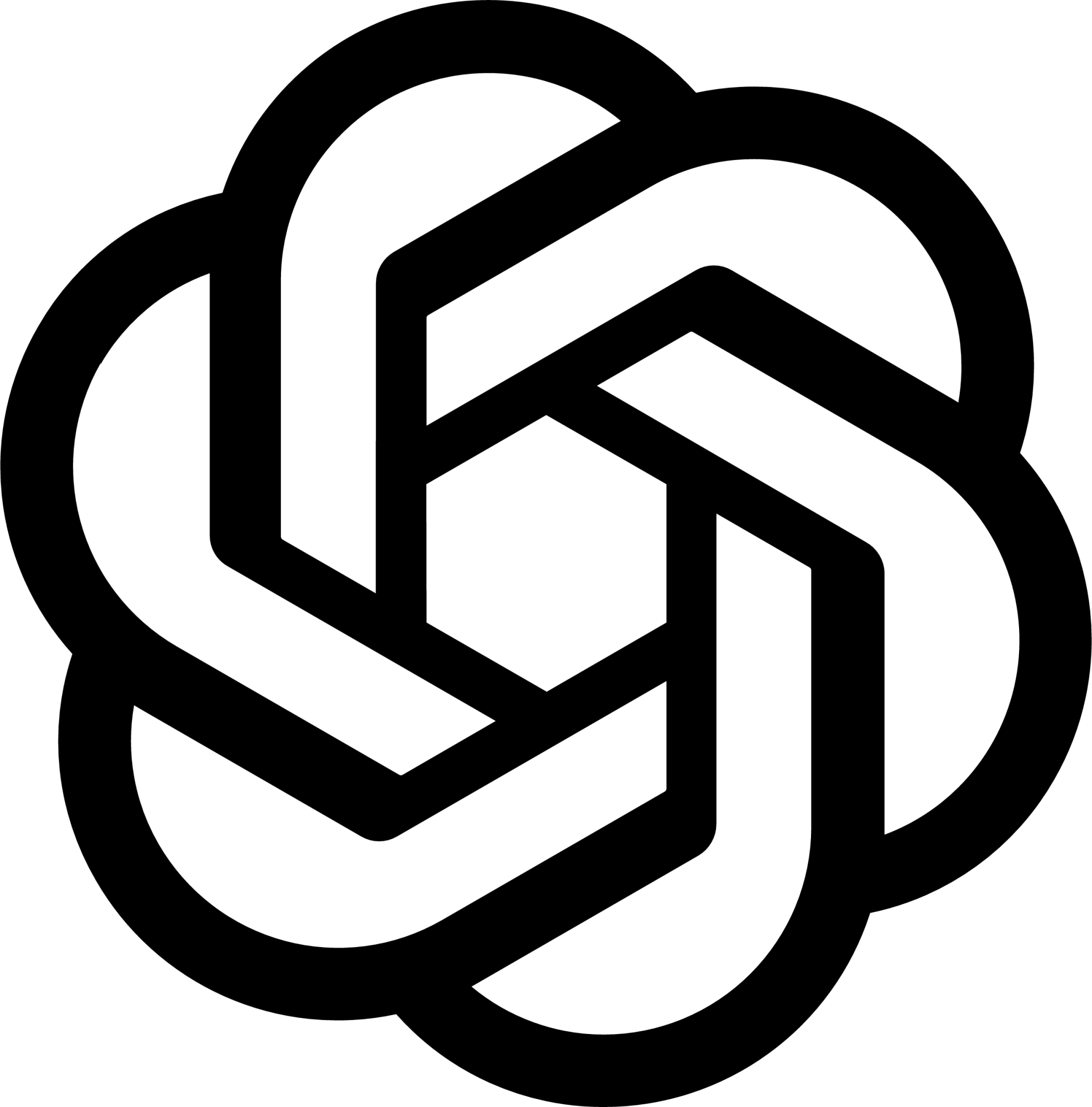} o4-mini & \textbf{15.8\%} & \textbf{47.7\%} & \textbf{19.5\%} & \textbf{10.7\%} & \textbf{1.2\%} & \textbf{1.1\%} \\
\quad \includegraphics[width=8px]{images/logos/openai.png}  o3-mini & 8.2\% & 29.1\% & 10.2\% & 2.5\% & \textbf{1.2\%} & 0.0\% \\
\quad \includegraphics[width=8px]{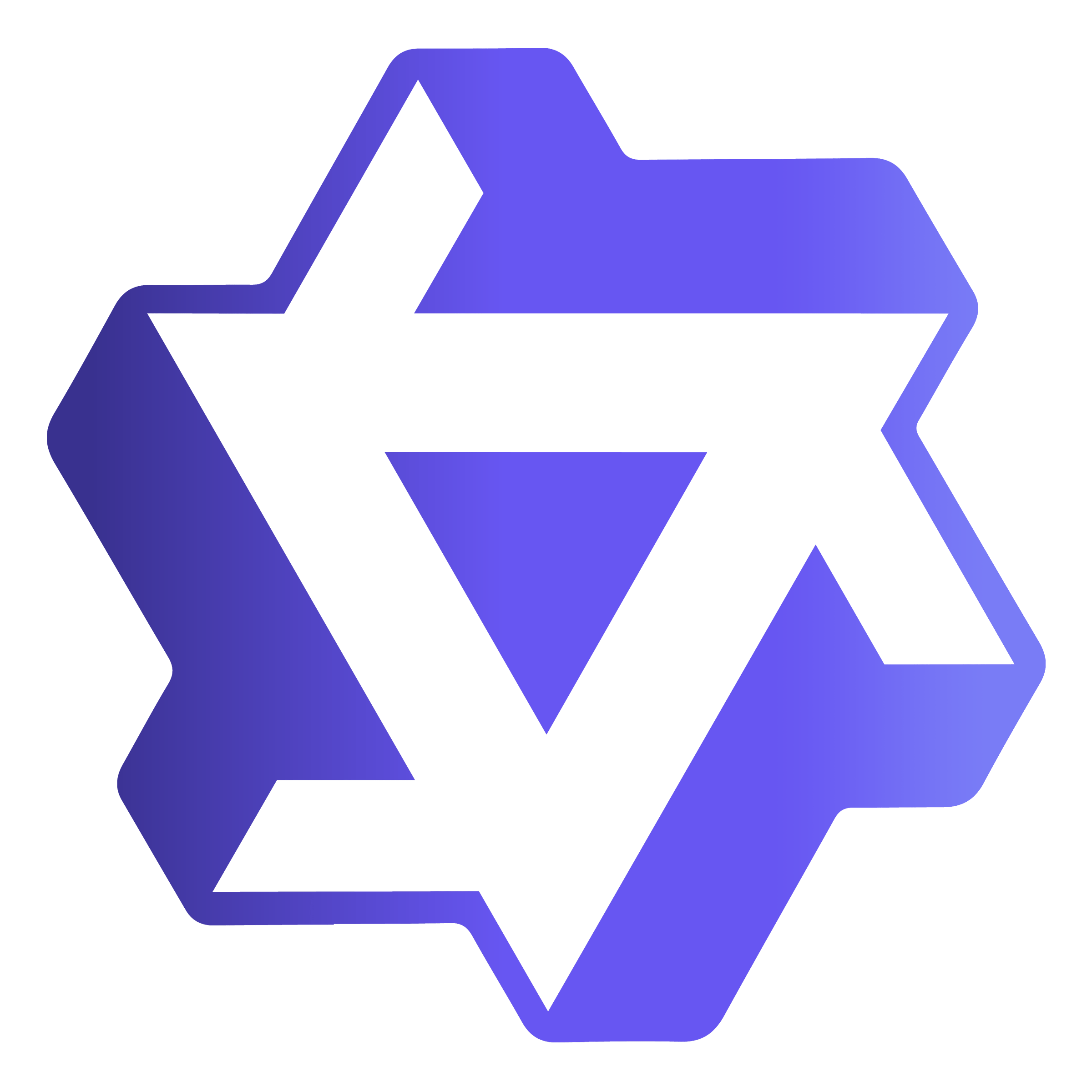} QwQ 32B & 5.8\% & 20.9\% & 5.9\% & 2.5\% & \textbf{1.2\%} & 0.0\% \\
\quad \includegraphics[width=8px]{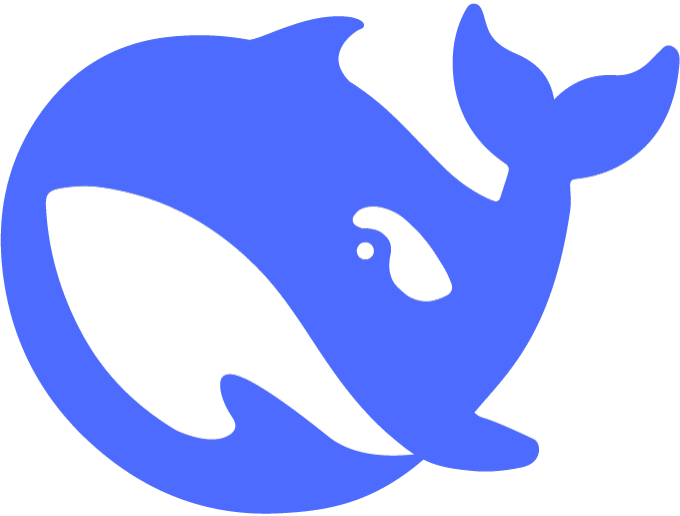} R1 70B & 4.0\% & 17.4\% & 2.5\% & 1.7\% & 0.0\% & 0.0\% \\
\textbf{Instruction} &  &  &  &  &  &  \\
\quad \includegraphics[width=8px]{images/logos/openai.png} GPT-4.1 & 1.6\% & 7.0\% & 0.8\% & 0.8\% & 0.0\% & 0.0\% \\
\quad \includegraphics[width=8px]{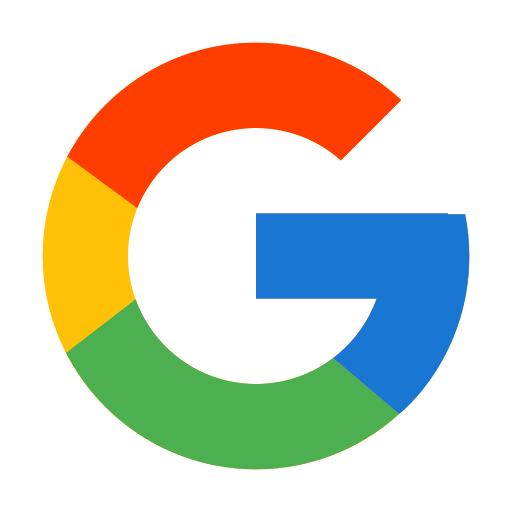} Gemma-3 27B & 1.2\% & 3.5\% & 0.8\% & 0.8\% & 0.0\% & \textbf{1.1\%} \\
\quad \includegraphics[width=8px]{images/logos/qwen.png} Qwen 2.5 72B & 0.4\% & 0.0\% & 1.7\% & 0.0\% & 0.0\% & 0.0\% \\
\bottomrule
\end{tabular}
\caption{
  \textbf{Accuracy (\%)} for \dataset\ puzzles achieved by various LLMs, categorized as \textbf{Reasoning} or \textbf{Instruction} models. 
  The table displays the overall accuracy (\textit{All}) and the breakdown by puzzle \textbf{Difficulty Level} (1-5) for each model. 
  Performance generally decreases sharply as the difficulty level increases. The highest overall performance is achieved by \textit{o4-mini} (\textbf{15.8\%}). Values are shown in percent (\%).
}
\label{tab:llm_solve_rate_difficulty_breakdown}
\end{table}

\subsection{Path Metrics}
\begin{table}[H]
\centering
\resizebox{\textwidth}{!}{\begin{tabular}{lccccc}
\toprule
Model & Incorrect Start/End & Disconnected Line & Intersecting Line & Rule Cell Crossing & Invalid Path \\
\midrule
\textbf{Reasoning} &  &  &  &  &  \\
\quad \includegraphics[width=8px]{images/logos/openai.png} o4-mini & 3.8\% & 27.6\% & 31.2\% & \textbf{51.2\%} & \textbf{59.2\%} \\
\quad \includegraphics[width=8px]{images/logos/openai.png} o3-mini & 3.0\% & \textbf{13.2\%} & \textbf{8.0\%} & 56.2\% & 63.2\% \\
\quad \includegraphics[width=8px]{images/logos/qwen.png} QwQ 32B & \textbf{1.6\%} & 26.2\% & 30.8\% & 70.0\% & 76.4\% \\
\quad \includegraphics[width=8px]{images/logos/deepseek.png} R1 70B & 10.2\% & 52.4\% & 35.8\% & 57.6\% & 82.2\% \\
\textbf{Instruction} &  &  &  &  &  \\
\quad \includegraphics[width=8px]{images/logos/openai.png} GPT-4.1 & 53.8\% & 87.0\% & 51.0\% & 55.0\% & 93.6\% \\
\quad \includegraphics[width=8px]{images/logos/google.png} Gemma-3 27B & 40.8\% & 37.6\% & 42.0\% & 84.6\% & 88.0\% \\
\quad \includegraphics[width=8px]{images/logos/qwen.png} Qwen 2.5 72B & 8.0\% & 41.0\% & 20.2\% & 59.0\% & 90.6\% \\
\bottomrule
\end{tabular}}
\caption{
  Percentage of generated solutions with path violations for \dataset\ puzzles across different LLMs. 
  Models are grouped into \textbf{Instruction} and \textbf{Reasoning} categories. 
  Columns show the rate (\%) for specific violation types. 
}
\label{tab:llm_path_violation_rates}
\end{table}

\subsection{Rule Specific Analysis}
\begin{table}[H]
\centering
\resizebox{\textwidth}{!}{\begin{tabular}{llllllllllll}
\toprule
Model & Full Set & Gaps & Dots & Stones & Stars & Tri & Polys & St-S & P-Y & G-D-T & D-S-P \\
\midrule
\textbf{Reasoning} &  &  &  &  &  &  &  &  &  &  &  \\
\quad \includegraphics[width=8px]{images/logos/openai.png} o4-mini & \textbf{15.8\%} & \textbf{84.0\%} & \textbf{22.0\%} & \textbf{16.0\%} & \textbf{34.0\%} & \textbf{14.0\%} & 16.0\% & \textbf{20.0\%} & \textbf{4.0\%} & \textbf{18.0\%} & 8.0\% \\
\quad \includegraphics[width=8px]{images/logos/openai.png} o3-mini & 8.2\% & 48.0\% & 10.0\% & 6.0\% & 8.0\% & 4.0\% & 2.0\% & 6.0\% & 2.0\% & 8.0\% & \textbf{10.0\%} \\
\quad \includegraphics[width=8px]{images/logos/qwen.png} QwQ 32B & 5.8\% & 52.0\% & 6.0\% & 8.0\% & 28.0\% & 2.0\% & \textbf{20.0\%} & 8.0\% & 2.0\% & 0.0\% & 6.0\% \\
\quad \includegraphics[width=8px]{images/logos/deepseek.png} R1 70B & 4.0\% & 32.0\% & 4.0\% & 2.0\% & 2.0\% & 4.0\% & 8.0\% & 6.0\% & 0.0\% & 4.0\% & 2.0\% \\
\textbf{Instruction} &  &  &  &  &  &  &  &  &  &  &  \\
\quad \includegraphics[width=8px]{images/logos/openai.png} GPT-4.1 & 1.6\% & 10.0\% & 0.0\% & 2.0\% & 4.0\% & 0.0\% & 4.0\% & 6.0\% & 0.0\% & 0.0\% & 2.0\% \\
\quad \includegraphics[width=8px]{images/logos/google.png} Gemma-3 27B & 1.2\% & 6.0\% & 0.0\% & 0.0\% & 2.0\% & 0.0\% & 14.0\% & 2.0\% & 2.0\% & 0.0\% & 0.0\% \\
\quad \includegraphics[width=8px]{images/logos/qwen.png} Qwen 2.5 72B & 0.4\% & 2.0\% & 0.0\% & 0.0\% & 0.0\% & 0.0\% & 8.0\% & 0.0\% & 0.0\% & 0.0\% & 2.0\% \\
\bottomrule
\end{tabular}}
\caption{
  \textbf{Accuracy (\%)} for various LLMs on \dataset\ puzzles, broken down by puzzle split type. 
  Models are categorized as \textbf{Reasoning} or \textbf{Instruction}. 
  Columns display the overall accuracy (\textit{Full Set}) and the accuracy (\%) on splits featuring specific single rules (\textit{Gaps}, \textit{Dots}, \textit{Stones}, \textit{Stars}, \textit{Tri}, \textit{Polys}) or rule combinations (\textit{St-S}: Stones \& Stars, \textit{P-Y}: Polys \& Ylops, \textit{G-D-T}: Gaps \& Dots \& Triangles, \textit{D-S-P}: Dots \& Stars \& Polys). 
  Values are shown in percent (\%).
}
\label{tab:llm_subset_solve_rates}
\end{table}

\section{Details on Path Errors}
\label{sec:appendix_visual_examples}
\begin{figure}[H]
  \centering
  \begin{tabular}{@{}ccc@{}}
    \begin{subfigure}[b]{0.30\textwidth}
      \includegraphics[width=\linewidth]{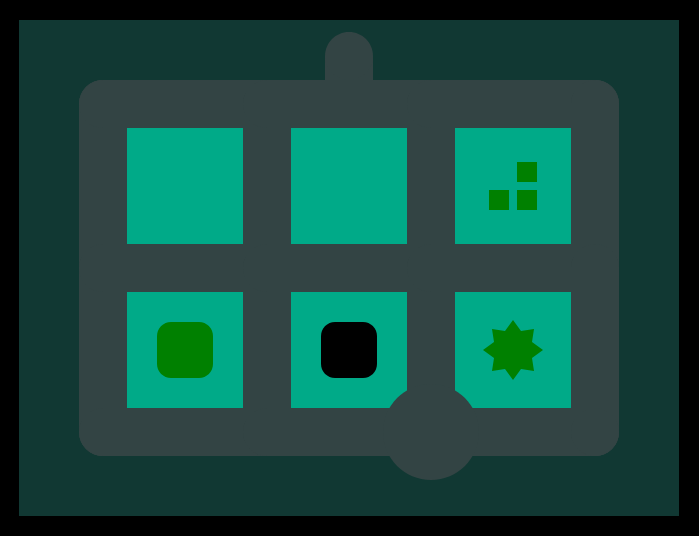}
      \caption{Empty puzzle grid.}
      \label{fig:appendix_empty_puzzle}
    \end{subfigure} &
    \begin{subfigure}[b]{0.30\textwidth}
      \includegraphics[width=\linewidth]{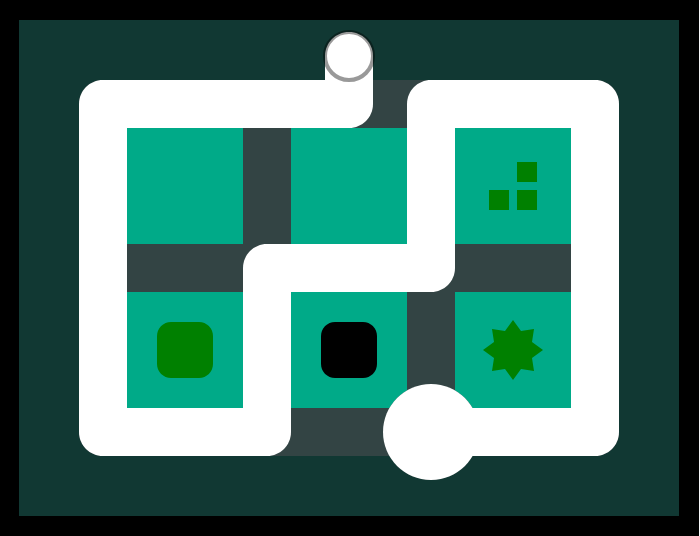}
      \caption{Solved puzzle (correct).}
      \label{fig:appendix_solved_puzzle}
    \end{subfigure} &
    \begin{subfigure}[b]{0.30\textwidth}
      \includegraphics[width=\linewidth]{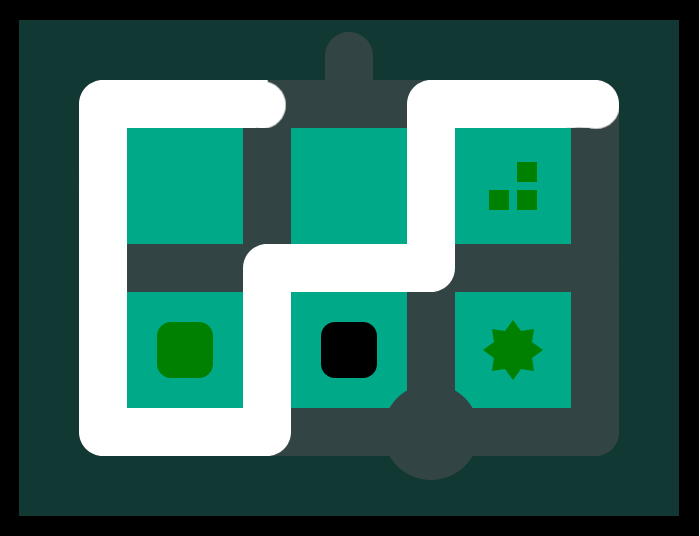}
      \caption{Incorrect start/end (incorrect).}
      \label{fig:appendix_error_incorrect_start_end}
    \end{subfigure} \\[1em]
    \begin{subfigure}[b]{0.30\textwidth}
      \includegraphics[width=\linewidth]{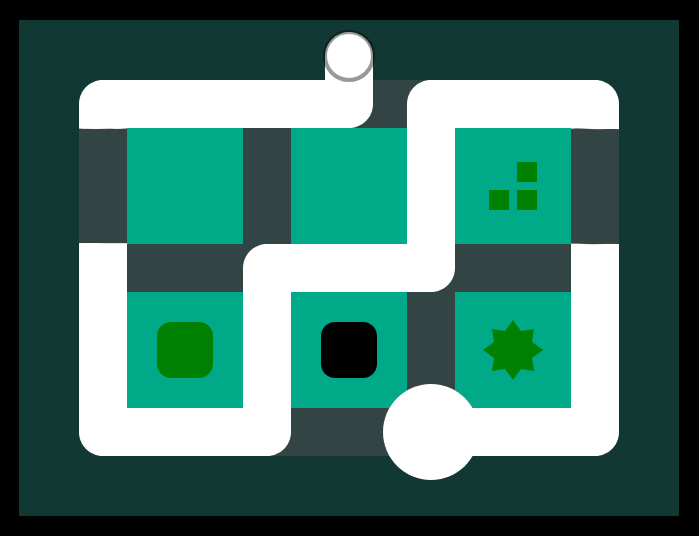}
      \caption{Disconnected line (incorrect).}
      \label{fig:appendix_error_disconnected_line}
    \end{subfigure} &
    \begin{subfigure}[b]{0.30\textwidth}
      \includegraphics[width=\linewidth]{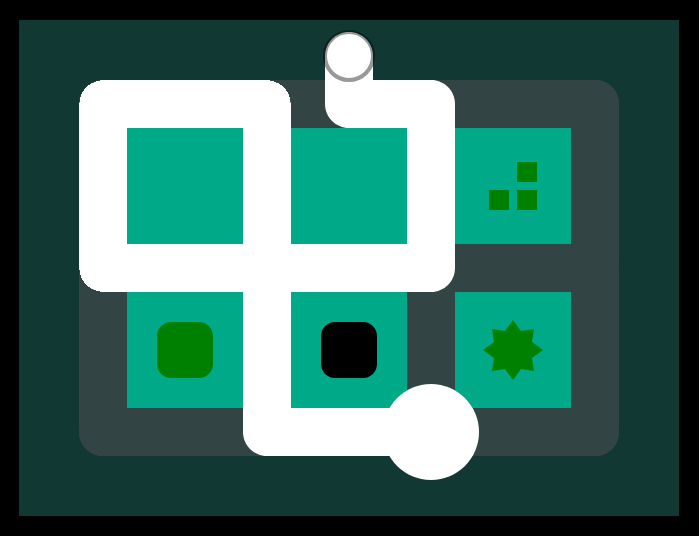}
      \caption{Self-intersecting path (incorrect).}
      \label{fig:appendix_error_self_intersecting_path}
    \end{subfigure} &
    \begin{subfigure}[b]{0.30\textwidth}
      \includegraphics[width=\linewidth]{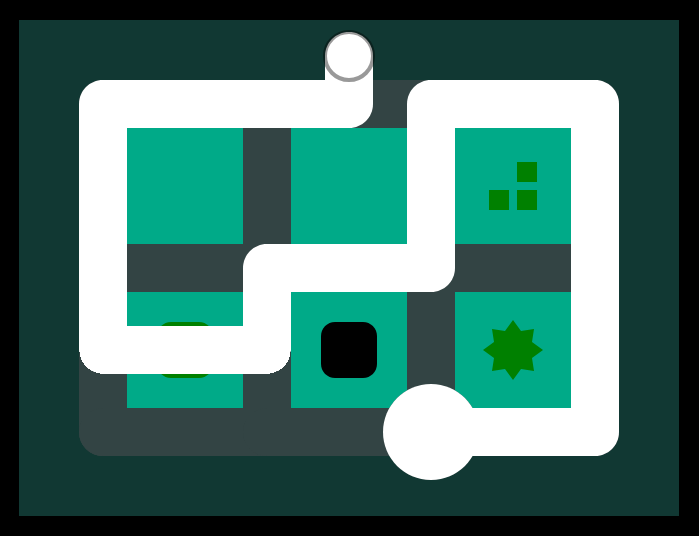}
      \caption{Rule-cell crossing (incorrect).}
      \label{fig:appendix_error_rule_cell_crossing}
    \end{subfigure}
  \end{tabular}
  \caption{%
    (a)–(b): An empty puzzle and its solution.  
    (c)–(f): The four kinds of path-generation errors.
  }
  \label{fig:appendix_error_types}
\end{figure}

\Cref{fig:appendix_error_types} shows the examples of \dataset\ puzzle structure and the fundamental path violation types discussed in Section~\ref{sec:evaluation}.

\clearpage
\section{Details on Reasoning Mistakes}
\label{ap:reasoning_mistakes}

\begin{figure}[H]
    \begin{AIbox}{Puzzle: 80a59619e323acba. Model: DeepSeek R1 Distill Llama 70B.}
    \parbox[t]{\linewidth}{
        \begin{center}
            \begin{tabular}{ccc}
                \includegraphics[width=0.3\linewidth]{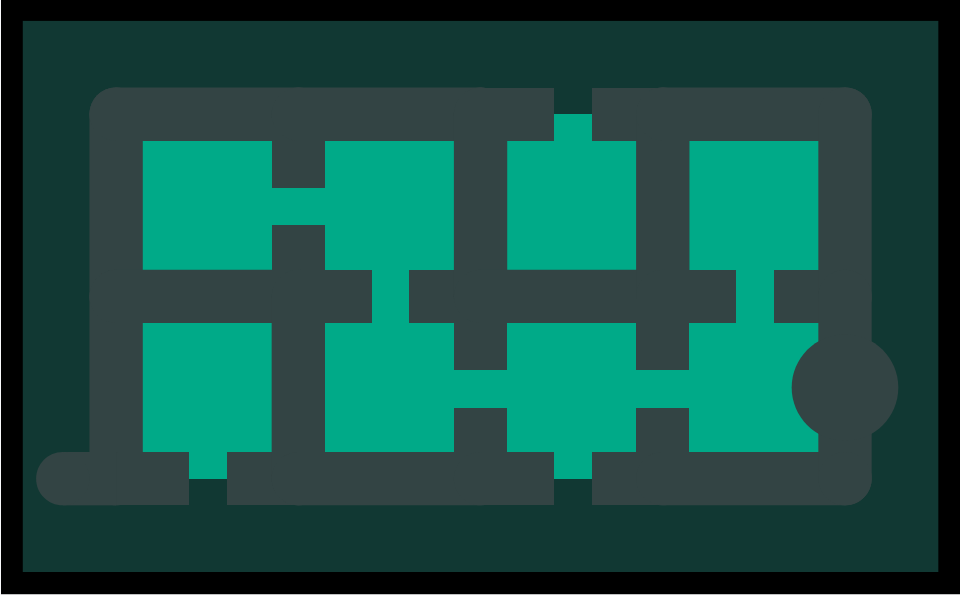} &
                \includegraphics[width=0.3\linewidth]{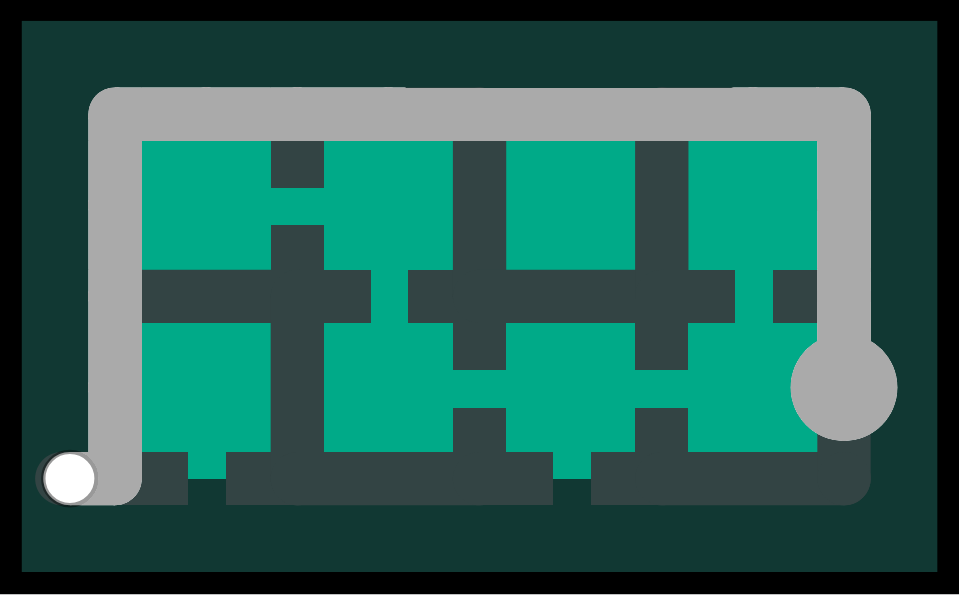} &
                \includegraphics[width=0.3\linewidth]{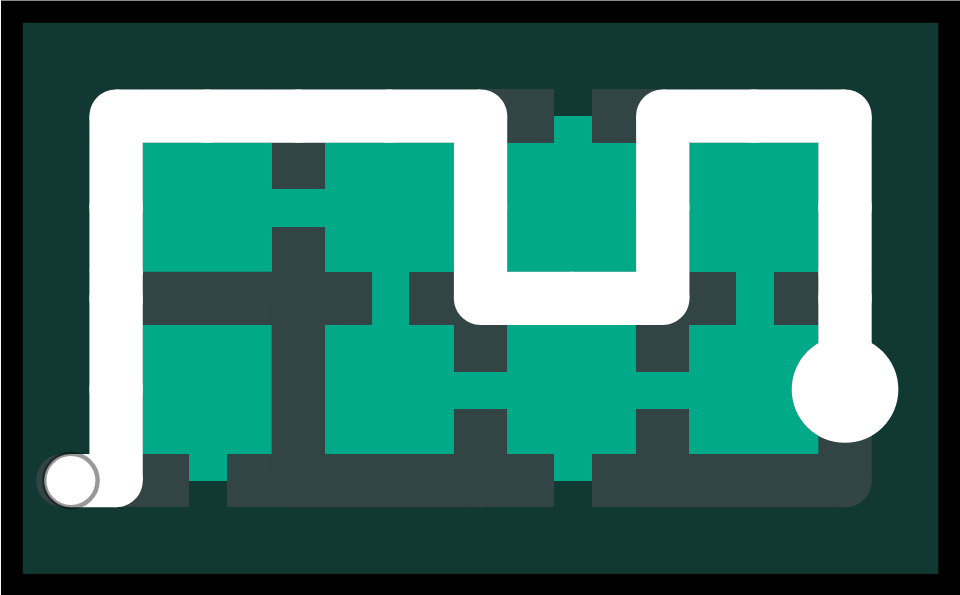} \\
                \small Puzzle Grid & \small Model's Path & \small Correct Path
            \end{tabular}
        \end{center}
        \tcbline %

        \noindent{\small {\bf Reasoning Excerpts:}}
        
        \noindent{\small ...}

        {\small Alright, so I've got this Witness puzzle to solve. Let me try to figure it out step by step.}
        
        \noindent{\small ...}

        \noindent{\small \hl{From (8,3) -> (8,2) -> (8,1)}. From there, perhaps moving left to (7,1), then (6,1), etc., but I need to avoid 'G's and 'N's.}

        \noindent{\small ...}

        \noindent{\small Wait, but \hl{from (8,3) to (8,2) to (8,1) to (8,0)}}

        \noindent{\small ...}

        \noindent{\small From \hl{from (8,0) to (7,0) to (6,0)}}

        \noindent{\small ...}

        \noindent{\small \hl{From (6,0), can go left to (5,0) or down to (6,1). Let's go left to (5,0). (6,0) -> (5,0). From (5,0), can go left to (4,0) or down to (5,1). Let's go left to (4,0). (5,0) -> (4,0).}}

        \noindent{\small ...}

        {\small\hl{So:}}\\
	\hspace*{0.5cm}{\small ...}\\
        \hspace*{0.5cm}{\small- (8,0) -> (7,0}\\
        \hspace*{0.5cm}{\small\hl{- (7,0) -> (6,0)}}\\
        \hspace*{0.5cm}{\small\hl{- (6,0) -> (5,0)}}\\
        \hspace*{0.5cm}{\small\hl{- (5,0) -> (4,0)}}\\
        \hspace*{0.5cm}{\small- (4,0) -> (3,0)}\\
	\hspace*{0.5cm}{\small ...}\\

        \noindent{\small ...}
        
        \noindent{\small \hl{But let me double-check if I missed any 'G's along the way.} For example, in row 0, column 5 is 'G', but the path doesn't go near there. Similarly, row 1 has 'G's at 2 and 5, but the path goes through row 1, columns 0 and 1, which are '+', so that's okay.}

        \noindent{\small ...}
        
        {\small \hl{Yes, this path should work.}}
    }
    \end{AIbox}
    \caption{Analysis of puzzle 80a59619e323acba with an incorrect solution attempt by DeepSeek R1 Llama 70B Distill. The model tends to commit to a sequence of reasoning steps without thoroughly validating each one, leading to careless mistakes.}
    \label{fig:reasoning_mistakes_puzzle_gaps}
\end{figure}

\begin{figure}[H]
    \begin{AIbox}{Puzzle: cf36a0865a1ed242. Model: DeepSeek R1 70B.}
    \parbox[t]{\linewidth}{
        \begin{center}
            \begin{tabular}{ccc}
                \includegraphics[width=0.3\linewidth]{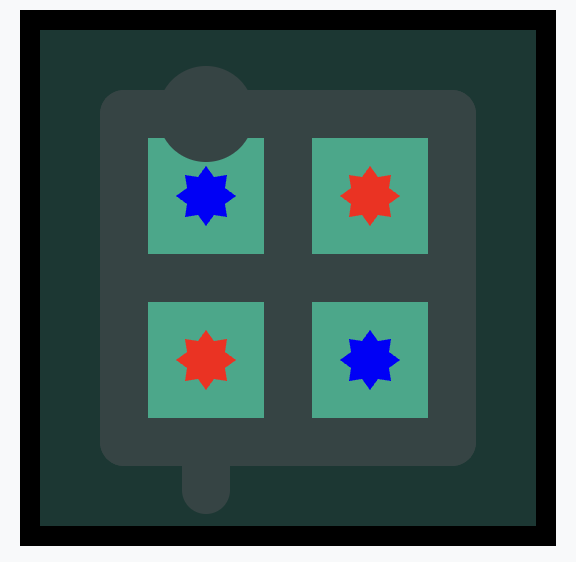} &
                \includegraphics[width=0.3\linewidth]{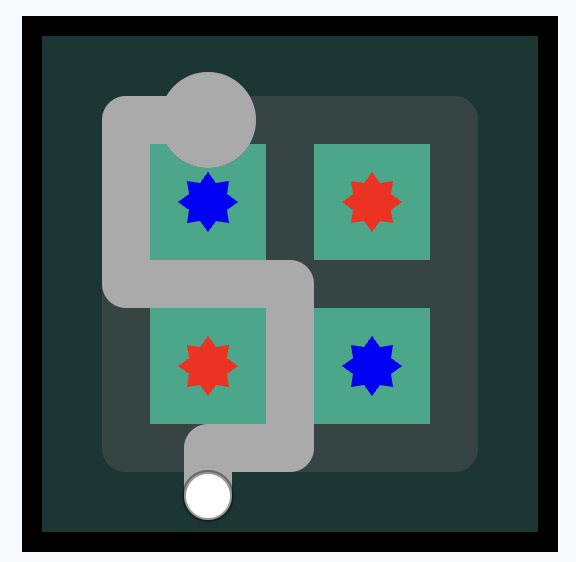} &
                \includegraphics[width=0.3\linewidth]{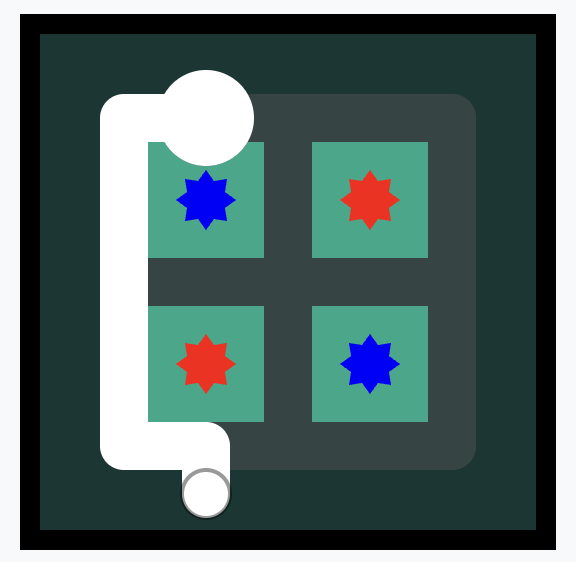} \\
                \small Puzzle Grid & \small Model's Path & \small Correct Path
            \end{tabular}
        \end{center}

        \tcbline

        \noindent{\small {\bf Reasoning Excerpts:}}\\
        \noindent{\small ...}
        
        \noindent{\small Since the line starts at (1,0) and ends at (1,4), it's a vertical line in the middle. \hl{So maybe the line goes straight down the middle, but that would split the grid into two halves}. But then the blue stars are on the top-left and bottom-right, and the red stars are on the top-right and bottom-left. \hl{So if the line goes straight down the middle, each blue star would be in a separate region, which might not satisfy the rule because each needs to be paired with exactly one other}. Similarly for the red stars.}

        \noindent{\small ...}

        \noindent{\small But how can the line enclose both blue stars in one region and both red stars in another, given the line starts and ends in the middle?}

        \noindent{\small ...}

        \noindent{\small Wait, perhaps the line can loop around the blue stars. Let me try to imagine \hl{a path that starts at (1,0), goes around the top-left blue star, then around the bottom-right blue star}, and then proceeds to the end.}

        \noindent{\small ...}

        \noindent{\small But the line can't cross itself, so it's challenging. Alternatively, \hl{maybe the line goes from (1,0) down to (1,4)}, but that's a straight line, which might not satisfy the star rules.}

        \noindent{\small ...}

        \noindent{\small Wait, maybe I need to create a loop that goes around the top-left and bottom-right blue stars. Let's try:}

        \noindent{\small (1,0) $\rightarrow$ (0,0) $\rightarrow$ (0,1) $\rightarrow$ (0,2) $\rightarrow$ (1,2) $\rightarrow$ (2,2) $\rightarrow$ (2,3) \hl{$\rightarrow$ (3,3) is blocked}. So can't. So from (2,3), go down to (2,4). Then left to (1,4).}

        \noindent{\small ...}

        \noindent{\small Wait, maybe the line can go from (1,0) to (0,0), then down to (0,1), \hl{then right to (1,1) is blocked, so go down to (0,2)}, then right to (1,2), \hl{then down to (1,3) is blocked}, so go right to (2,2), then down to (2,3), \hl{then right to (3,3) is blocked}, so go down to (2,4), then left to (1,4).}

        \noindent{\small ...}

        \noindent{\small But this doesn't enclose the blue stars. I'm stuck.}

        \noindent{\small ...}

        \noindent{\small \hl{I think I'm stuck} and need to conclude that I can't find a solution right now.}
    }
    \end{AIbox}
    \caption{Analysis of puzzle cf36a0865a1ed242 from the stars split with incorrect solution by DeepSeek R1 70B. The model fails to realize that the grid can form one large region with two stars of each color, leading to an incorrect path. It often tries to loop around stars and run into grid cells, which is impossible. The model notices it got stuck and provides the latest attempt to a valid solution.}
    \label{fig:reasoning_mistakes_puzzle_stars}
\end{figure}

\begin{figure}[H]
    \begin{AIbox}{Puzzle: a49a784ae7d91339. Model: DeepSeek R1 Llama 70B Distill.}
    \parbox[t]{\linewidth}{
        \begin{center}
            \begin{tabular}{ccc}
                \includegraphics[width=0.3\linewidth]{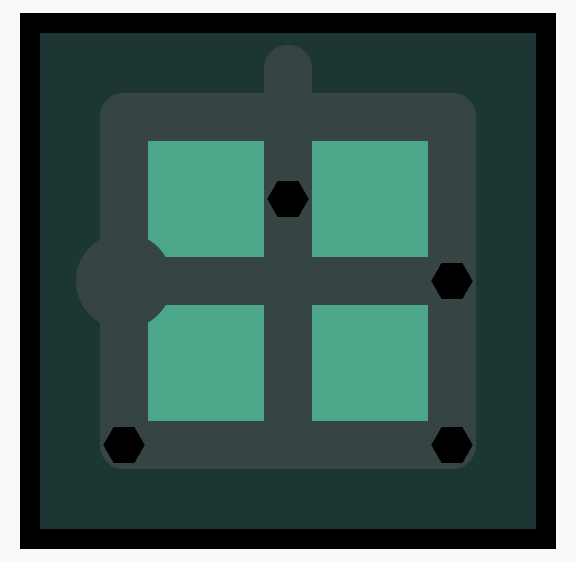} &
                \includegraphics[width=0.3\linewidth]{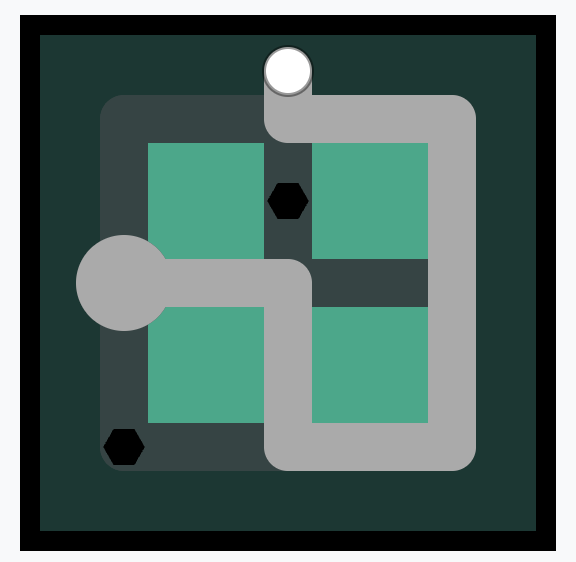} &
                \includegraphics[width=0.3\linewidth]{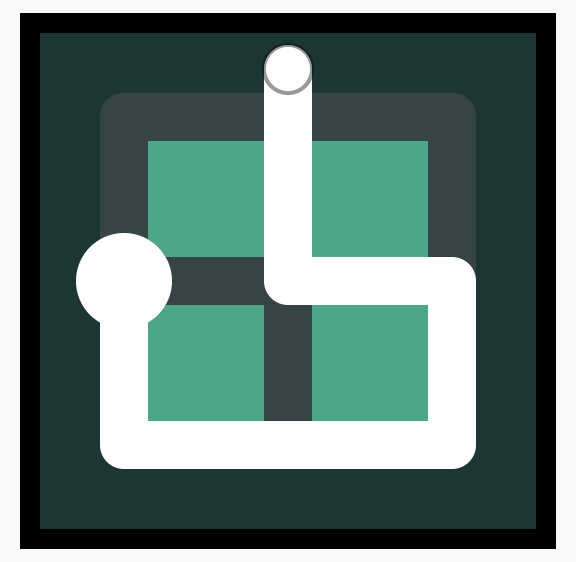} \\
                \small Puzzle Grid & \small Model's Path & \small Correct Path
            \end{tabular}
        \end{center}

        \tcbline

        \noindent{\small {\bf Reasoning Excerpts:}}\\
        \noindent{\small ...}
        
        \noindent{\small Wait, that \hl{seems like a 5x5 grid of nodes, but the actual cells are 2x2} because each cell is four nodes. Hmm, maybe I'm misunderstanding.}

        \noindent{\small ...}

        \noindent{\small Wait, I'm getting confused. Let me clarify. \hl{The grid is 2x2 cells, each cell being a square.} So the nodes are the intersections, making a 3x3 grid of nodes. \hl{So the nodes go from (0,0) to (2,2). The start is at (0,2) which is the bottom-left node, and the end is at (2,0), the top-right node.}}

        \noindent{\small ...}

        \noindent{\small Start at (0,2). From there, go up to (0,1), then right to (1,1) — \hl{but (1,1) is a rule cell, so can't go there.} So instead, go right to (1,2), which is a '.' that must be passed through. So from (0,2) to (1,2).}

        \noindent{\small ...}

        \noindent{\small \hl{Wait, but the line must pass through (2,4).} So perhaps from (1,2), go down to (2,2), then right to (2,3), then down to (2,4), which is a '.'.}

        \noindent{\small ...}

        \noindent{\small So path so far: (0,2) $\rightarrow$ (1,2) $\rightarrow$ (2,2) $\rightarrow$ (2,3) $\rightarrow$ (2,4).}

        \noindent{\small ...}

        \noindent{\small From (2,4), \hl{need to get to (4,4), which is another '.'.} So go right to (3,4), then to (4,4).}

        \noindent{\small ...}

        \noindent{\small Now, from (4,4), \hl{need to get to (4,0), another '.'.} So go down to (4,3), (4,2), (4,1), (4,0).}

        \noindent{\small ...}

        \noindent{\small \hl{But wait, (4,0) is a '.' and the end is at (2,0).} So from (4,0), go left to (3,0), then to (2,0). But (2,0) is the end point, so that's acceptable.}

        \noindent{\small ...}

        \noindent{\small \hl{But does this path cover all the '.'s?} Let's see:}

        \hspace*{0.5cm} {\small- \hl{(1,2)}: yes}\\
        \hspace*{0.5cm} {\small- (2,4): yes}\\
        \hspace*{0.5cm} {\small- (4,4): yes}\\
        \hspace*{0.5cm} {\small- \hl{(4,0)}: yes}\\

        \noindent{\small \hl{I think this path satisfies all the constraints.}}
    }
    \end{AIbox}
    \caption{Analysis of puzzle a49a784ae7d91339 from the dots split with incorrect solution by DeepSeek R1 Llama 70B Distill. The model has issues interpreting the grid sizes as 5x5 initially, but converges to the correct dimensions after some reasoning steps. The model tries to draw paths on rule cells but corrects itself quickly. Finally, the model mixes up $x$ and $y$ coordinates, thinking it had drawn the correct path but drawing an inverse indexed path.}
    \label{fig:reasoning_mistakes_puzzle_dots}
\end{figure}

\begin{figure}[H]
    \begin{AIbox}{Puzzle: d9e6e97a86d583d9. Model: DeepSeek R1 70B.}
    \parbox[t]{\linewidth}{
        \begin{center}
            \begin{tabular}{ccc}
                \includegraphics[width=0.3\linewidth]{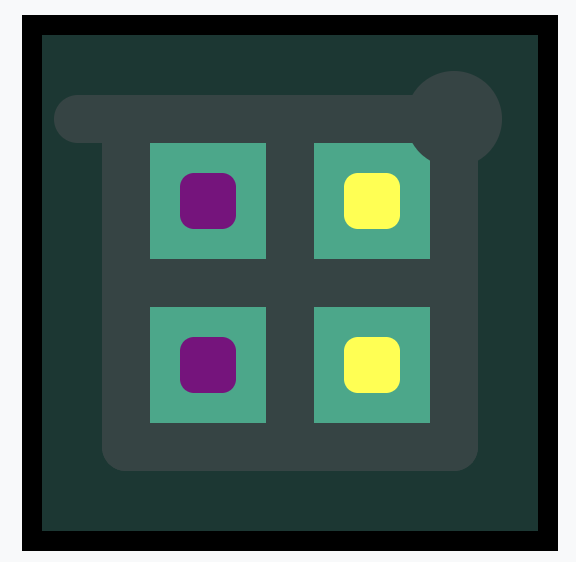} &
                \includegraphics[width=0.3\linewidth]{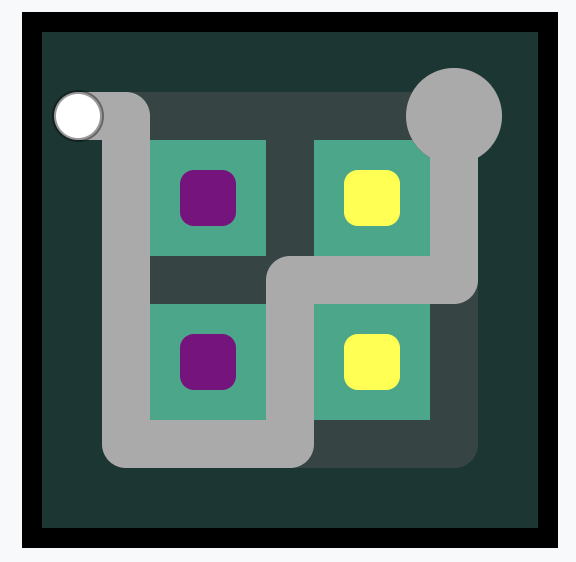} &
                \includegraphics[width=0.3\linewidth]{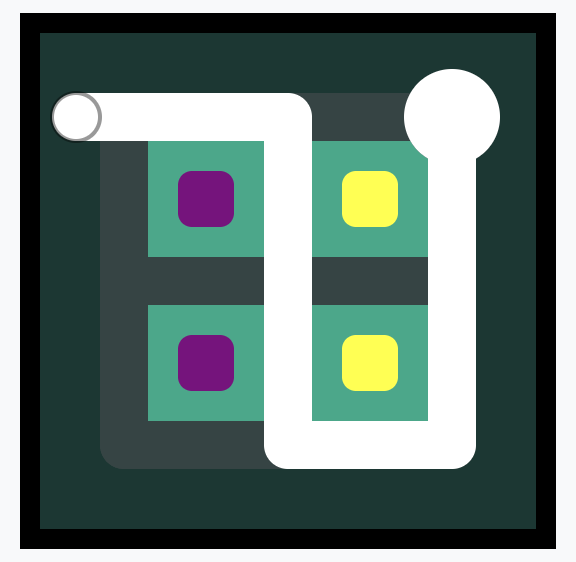} \\
                \small Puzzle Grid & \small Model's Path & \small Correct Path
            \end{tabular}
        \end{center}

        \tcbline

        \noindent{\small {\bf Reasoning Excerpts:}}\\
        \noindent{\small ...}
        
        \noindent{\small Okay, so I'm trying to solve this puzzle from The Witness. It's a bit tricky, but I'll take it step by step. Let me first understand the problem.}

        \noindent{\small ...}

        \noindent{\small \hl{Starting at (4,0), which is the bottom-right corner.} The end is at (0,0), top-left. So the line needs to move up and left.}

        \noindent{\small ...}

        \noindent{\small From (4,2), move left to (3,2). From (3,2), move down to \hl{(3,3) is blocked by a rule cell}, so instead move left to (2,2).}

        \noindent{\small ...}

        \noindent{\small This path would create a region on the left that includes (1,1) and (3,1), and a region on the right that includes (1,3) and (3,3).}

        \noindent{\small ...}

        \noindent{\small \hl{Yes, this should satisfy the rules} because each region has only one color of stones.}
    }
    \end{AIbox}
    \caption{Analysis of puzzle d9e6e97a86d583d9 of the stones split with incorrect solution by DeepSeek R1 70B. The model misinterprets the coordinate system, assuming (4,0) is the bottom-right corner, which is incorrect, as (4,4) is the bottom-right. The model also often attempts to draw a line over rule cells. This leads to an incorrect path that fails to satisfy the puzzle's rules.}
    \label{fig:reasoning_mistakes_puzzle_stones}
\end{figure}

\Cref{fig:reasoning_mistakes_puzzle_stars,fig:reasoning_mistakes_puzzle_stones,fig:reasoning_mistakes_puzzle_dots,fig:reasoning_mistakes_puzzle_gaps} provide four examples of common reasoning mistakes for DeepSeek R1 70B and highlights the steps that lead to the mistake.

\clearpage
\section{Details on Ablations}
\label{ap:ablations}
\begin{figure}[H] 
    \centering
    \includegraphics[width=\linewidth]{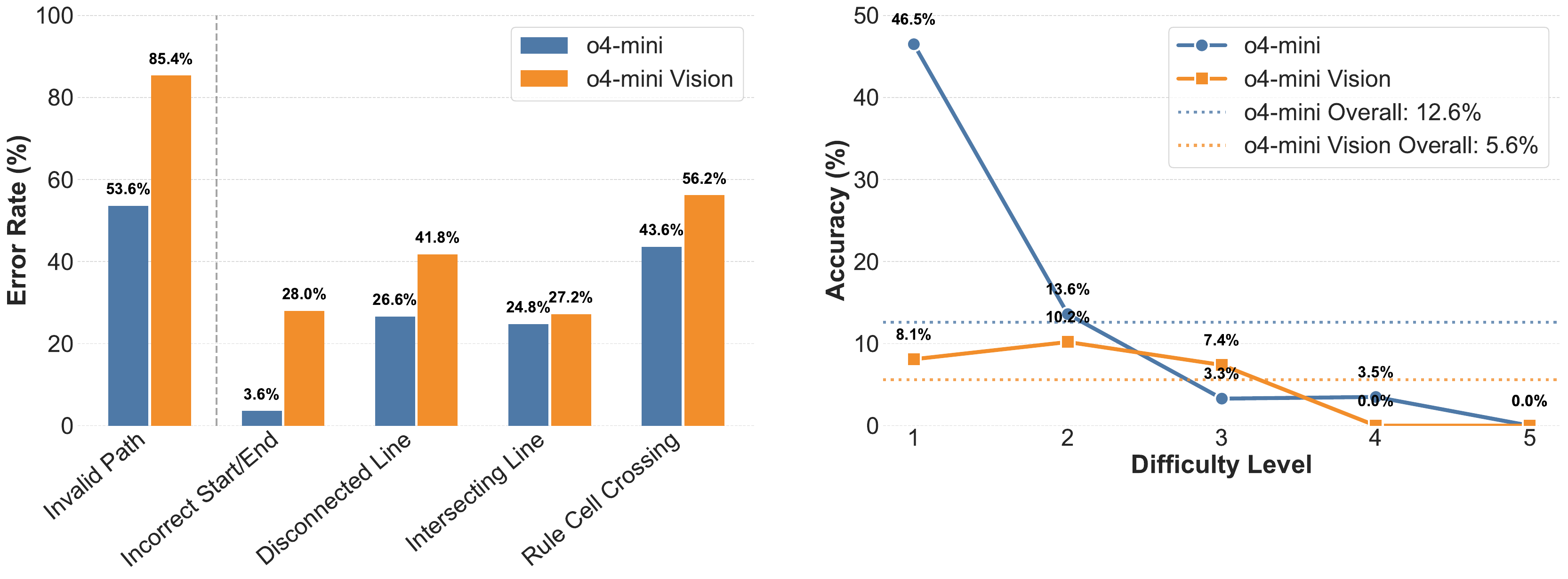}
\caption{
  Comparison of the standard \textit{o4-mini Zero-Shot} (blue) and its vision-enabled counterpart \textit{o4-mini Vision} (orange) on \dataset\ puzzles. 
  \textbf{Left Panel:} Bar chart showing the \textbf{Error Rate (\%)} for different types of path violations across all generated solutions. \textit{o4-mini Vision} generally exhibits higher rates of these structural errors. 
  \textbf{Right Panel:} Line chart displaying the \textbf{Accuracy (\%)} against puzzle \textbf{Difficulty Level} (1-5). The standard \textit{o4-mini Zero-Shot} achieves a significantly higher overall accuracy (\textbf{12.6\%}, blue dotted line) compared to \textit{o4-mini Vision} (\textbf{5.6\%}, orange dotted line), outperforming it at nearly all difficulty levels.
}
\label{fig:o4mini_vs_vision_comparison}
\end{figure}

\begin{figure}[H] 
    \centering
    \includegraphics[width=\linewidth]{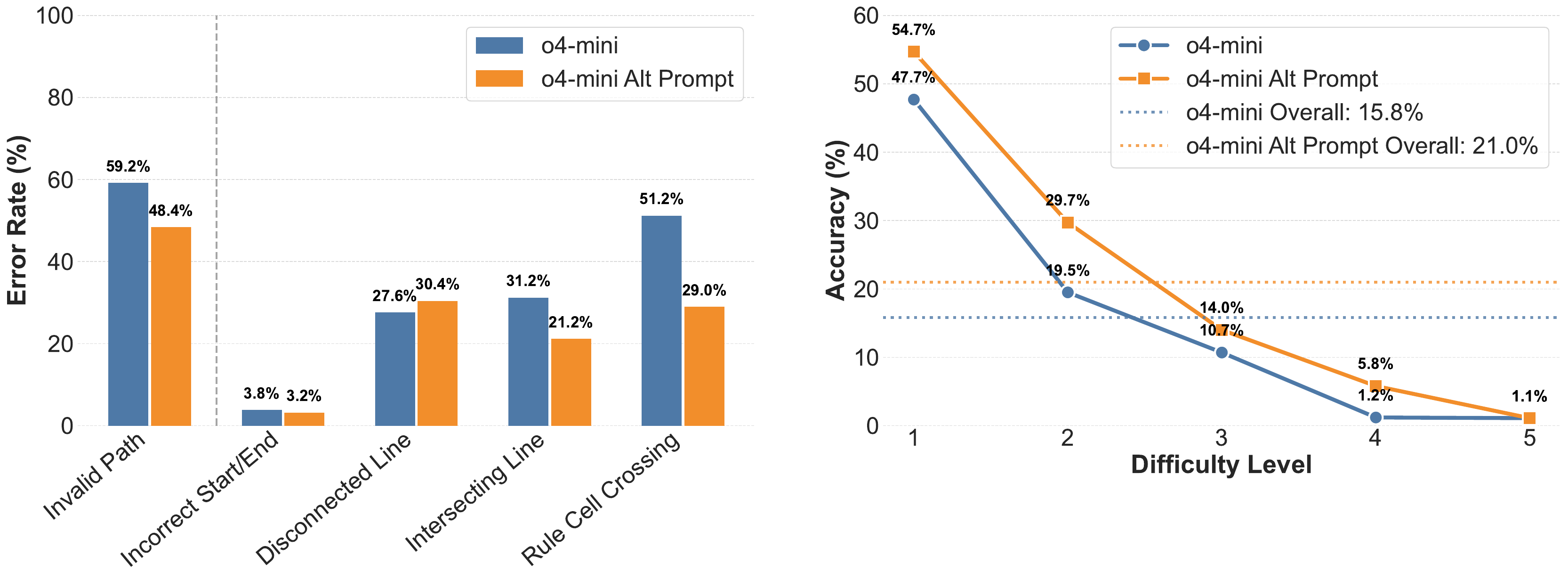}
\caption{
  Performance comparison of \textit{o4-mini} using its standard prompt (blue) versus an \textit{alternative prompt} (orange) on \dataset\ puzzles. 
  \textbf{Left Panel:} Bar chart showing the \textbf{Error Rate (\%)} for different types of path violations across all generated solutions. The alternative prompt generally reduces the frequency of these structural errors. 
  \textbf{Right Panel:} Line chart displaying the \textbf{Accuracy (\%)} against puzzle \textbf{Difficulty Level} (1-5). The alternative prompt results in a higher accuracy across all difficulties, improving the overall success from \textbf{15.8\%} (standard, blue dotted line) to \textbf{21.0\%} (alternative, orange dotted line).
}
\label{fig:o4mini_prompt_comparison}
\end{figure}

\begin{figure}[H] 
    \centering
    \includegraphics[width=\linewidth]{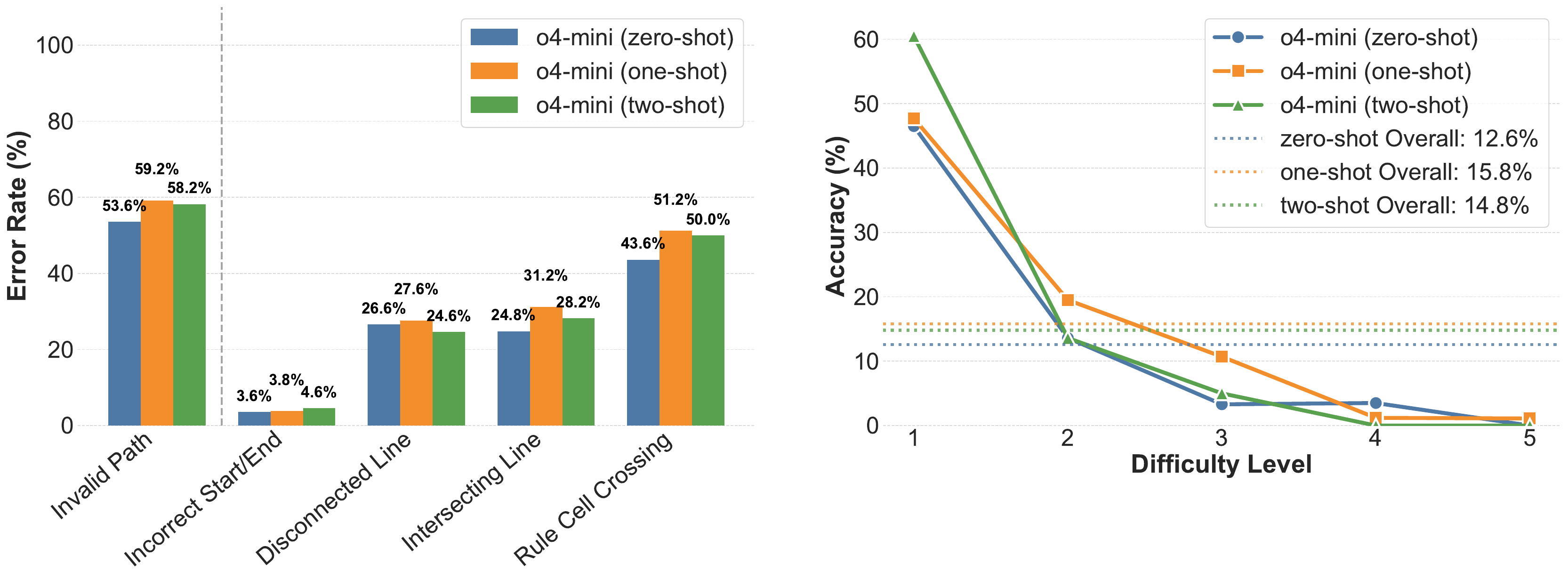}
\caption{
  Impact of few-shot prompting on \textit{o4-mini}'s performance and error profile for \dataset\ puzzles. 
  Compares \textit{zero-shot} (blue), \textit{one-shot} (orange), and \textit{two-shot} (green) prompting strategies.
  \textbf{Left Panel:} Bar chart showing the \textbf{Error Rate (\%)} for different types of fundamental path violations across all generated solutions. Few significant differences emerge in the error profiles across prompting strategies.
  \textbf{Right Panel:} Line chart displaying the \textbf{Accuracy (\%)} against puzzle \textbf{Difficulty Level} (1-5). While \textit{one-shot} prompting achieves the highest overall success rate (\textbf{15.8\%}, orange dotted line) compared to \textit{zero-shot} (\textbf{12.6\%}, blue dotted line) and \textit{two-shot} (\textbf{14.8\%}, green dotted line), all strategies show a sharp decline in performance as puzzle difficulty increases.
}
\label{fig:o4mini_fewshot_analysis}
\end{figure}

\Cref{fig:o4mini_vs_vision_comparison,fig:o4mini_prompt_comparison,fig:o4mini_fewshot_analysis} provide more details for the ablation experiments in \Cref{sec:ablations}, considering vision models, alternative prompts, and few-shot examples.
\subsection{Vision Mode}
\label{ap:vision-details}

\subsection{Alternative Prompt}

\subsection{Few-Shot}
\label{ap:few-shot-details}

\begin{figure}[H]
    \centering
    \includegraphics[width=0.7\textwidth]{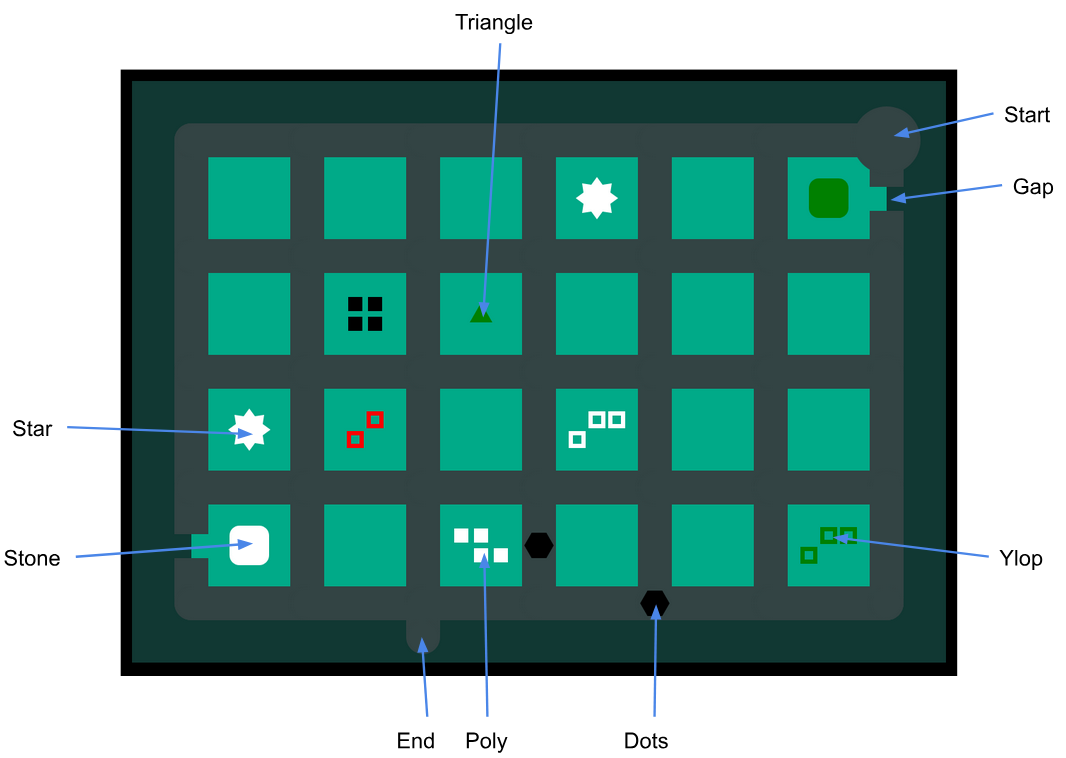}
    \caption{Visual explanation of how each rule looks on the puzzle grid for the annotators of the dataset.}
    \label{fig:app_rule_explanation}
\end{figure}

\twocolumn
\section{Details on Human Annotation}
\label{app:human_annotator_instructions}
\subsection{Annotators}
The annotators are two Ph.D. students and four research assistants (two bachelor students and two master students) in Computer Science and Data Science.
They are all male and between 22 and 27 years old.
As part of their research job, they receive at least the minimum wage in Germany.

\subsection{Annotation Instructions}
\paragraph{Introduction.}
This guide provides the rules and instructions for annotating \dataset\ puzzles. We want to compare whether there are patterns/similarities that make the puzzles difficult for humans or LLMs to solve. Therefore, we have created 6 test sets of 50 puzzles each. These sets contain puzzles with all possible combinations of rules and grid sizes.

\paragraph{Task.}

Annotate all the samples in the dataset in as little time as possible per puzzle. Each puzzle is solvable, but if you can't think of a solution after a reasonable amount of time (5-10 minutes), you can click Show Solution or Skip Puzzle to continue. Once you have completed all 50 puzzles, please e-mail the annotated file back to us. By sending the file back to us, you agree that we can publish your annotations anonymously. This includes solve time, required attempts, and solution path.

\paragraph{Rules.}

The line must connect Start with End with a continuous path without using the same cell twice. It must also follow all rules defined by the puzzle. A visual explanation of the rules can be seen in \Cref{fig:app_rule_explanation}.
\begin{itemize}
    \item \textbf{Gaps:} The line CANNOT traverse a cell marked by a Gap.
    \item \textbf{Dots:} The line MUST pass through a cell marked by a Dot.
    \item \textbf{Stone:} All stones within a single region created by the line must be the same color. Different colored squares MUST be separated into different regions by the line.
    \item \textbf{Stars:} Each star must be paired with EXACTLY one other element of the same color in a region. Other colors are ignored.
    \item \textbf{Triangles:} The line must touch EXACTLY the number of edges specified by the triangle count (edges are top, right, bottom, left of the cell).
    \item \textbf{Polyshapes (Poly):} The region containing this symbol must be shaped EXACTLY like the defined polyshape. The shape must fit entirely within the region's boundaries. If multiple positive polyshapes are in one region, the region's shape must accommodate their combined, non-overlapping forms (like Tetris pieces).
    \item \textbf{Negative Polyshapes (Ylop):} The negative polyshape can only be placed on top of already placed normal polyshapes. The negative polyshapes must fit on the grid, but can allow overlap between normal polyshapes or placement of polyshapes that extend beyond the area defined by the line. If the negative polyshapes exactly cancel the normal polyshapes, there is no restriction on the grid shape anymore. A negative polyshape only counts as valid if it is used.
\end{itemize}

\paragraph{Example Dataset.}
You can use the following dataset to experiment and get familiar with the puzzles and all rules:

\textbf{Link redacted for anonymity.}

\paragraph{Important Hints}
\begin{itemize}
    \item The annotation state gets saved even when closing the window, but to be safe, also always download the current annotated dataset when you stop annotating.
    \item If you reload the page, don't overwrite the existing data.
\end{itemize}
\onecolumn
\section{Acknowledgment of AI Usage}

AI Usage card based on \citet{wahle2023aiusagecardsresponsibly}.

{\sffamily
    \centering
    \tcbset{colback=white!10!white}
    \begin{tcolorbox}[
        title={\large \textbf{AI Usage Card} \hfill \makebox{\qrcode[height=1cm]{https://ai-cards.org}}},
        breakable,
        boxrule=0.7pt,
        width=\textwidth,
        center,
        before lower={\footnotesize{AI Usage Card v2.0 \hfill \url{https://ai-cards.org} \hfill \href{https://jpwahle.com/ai-cards-preprint}{PDF} | \href{https://jpwahle.com/cite/jcdl2023wahle.bib}{BibTeX}}},
        segmentation empty,
        halign lower=center,
        collower=black,
        coltitle=black, %
        colbacklower=gray!20, %
        colbacktitle=gray!20  %
        ]
        \vspace{-10pt}
        \footnotesize{
            \begin{longtable}{>
            {\raggedright\arraybackslash}p{.15\textwidth}>{\raggedright\arraybackslash}p{.25\textwidth}>{\raggedright\arraybackslash}p{.25\textwidth}>{\raggedright\arraybackslash}p{.25\textwidth}}
              
                {\color{LightBlue} \MakeUppercase{Project Details}} \newline 
                & {\color{LightBlue} \MakeUppercase{Project Name}} \newline SPaRC: A Spatial Pathfinding Reasoning Challenge
                & {\color{LightBlue} \MakeUppercase{Domain}} \newline Paper
                & {\color{LightBlue} \MakeUppercase{Key Application}} \newline Dataset
                
                \\
                
                {\color{LightBlue} \MakeUppercase{Contact(s)}}  
                & {\color{LightBlue} \MakeUppercase{Name(s)}} 
                & {\color{LightBlue} \MakeUppercase{Email(s)}} 
                & {\color{LightBlue} \MakeUppercase{Affiliation(s)}} 
                \\ & Lars Benedikt Kaesberg & \href{mailto:l.kaesberg@uni-goettingen.de}{l.kaesberg@uni-goettingen.de} & University Göttingen
                
                \\
                
                {\color{LightBlue} \MakeUppercase{Model(s)}} 
                & {\color{LightBlue} \MakeUppercase{Model Name(s)}} 
                & {\color{LightBlue} \MakeUppercase{Version(s)}} 
                \\ & ChatGPT & 4o, 4.5, o3
                \\ & Gemini & 2.5 pro
                \\ & Claude & 3.7 sonnet

                \\
                \cmidrule{1-4}
                \\
                {\color{LightBlue} \MakeUppercase{Literature Review}} \newline 
                & {\color{LightBlue} \MakeUppercase{Finding literature}} \newline ChatGPT \newline Gemini 
                & {\color{gray} \MakeUppercase{Finding examples from known literature or adding literature for existing statements}}  
                & {\color{gray} \MakeUppercase{Comparing literature}}  
                \\
                \cmidrule{2-4}
                \\        
                {\color{LightBlue} \MakeUppercase{Writing}} \newline    
                & {\color{gray} \MakeUppercase{Generating new text based on instructions}} 
                & {\color{LightBlue} \MakeUppercase{Assisting in improving own content or Paraphrasing related work}} \newline ChatGPT \newline Gemini 
                & {\color{gray} \MakeUppercase{Putting other works in perspective}}  
                \\
                \cmidrule{2-4}
                \\
                {\color{LightBlue} \MakeUppercase{Coding}} \newline 
                & {\color{LightBlue} \MakeUppercase{Generating new code based on descriptions or existing code}} \newline ChatGPT \newline Gemini \newline Claude 
                & {\color{LightBlue} \MakeUppercase{Refactoring and optimizing existing code}} \newline ChatGPT \newline Gemini \newline Claude 
                & {\color{gray} \MakeUppercase{Comparing aspects of existing code}}  
                \\
                \cmidrule{1-4}
                \\
        
                {\color{LightBlue} \MakeUppercase{Ethics}} \newline    
                & {\color{LightBlue} \MakeUppercase{Why did we use AI for this project?}} \newline Efficiency / Speed \newline Expertise Access  
                & {\color{LightBlue} \MakeUppercase{What steps are we taking to mitigate errors of AI?}} \newline -
                & {\color{LightBlue} \MakeUppercase{What steps are we taking to minimize the chance of harm or inappropriate use of AI?}} \newline -

                \\
                \cmidrule{1-4}
                \\
            \end{longtable}

        \medskip
        
        \textbf{\color{LightBlue} \MakeUppercase{The corresponding authors verify and agree with the modifications or generations of their  used AI-generated content}}
        }
        
        \tcblower
    \end{tcolorbox}
}

\clearpage
\hypertarget{annotation}{}
\pagestyle{empty}
\lstset{
  basicstyle=\footnotesize\ttfamily,
  breaklines=true,
  breakatwhitespace=false,
  columns=flexible,
  numbers=none
}

\definecolor{Primary}{RGB}{59, 130, 246}    %
\definecolor{PrimaryDark}{RGB}{30, 64, 175} %
\definecolor{LightBg}{RGB}{239, 246, 255}   %
\definecolor{TextDark}{RGB}{31, 41, 55}     %
\definecolor{TextMuted}{RGB}{107, 114, 128} %

\begin{tikzpicture}[remember picture, overlay]
  \fill[Primary] ([xshift=0cm,yshift=0cm]current page.north west) rectangle ([xshift=\paperwidth,yshift=-0.4cm]current page.north west);
\end{tikzpicture}

\vspace{0.8cm}
\begin{center}
  {\fontsize{22}{26}\selectfont\sffamily\bfseries \textcolor{PrimaryDark}{CiteAssist}}\\[0.2em]
  {\Large\sffamily\scshape \textcolor{TextMuted}{Citation Sheet}}\\[0.8em]
  {\small\sffamily Generated with \href{https://citeassist.uni-goettingen.de/}{\textcolor{Primary}{\texttt{citeassist.uni-goettingen.de}}}\\
  \citep{kaesberg-etal-2024-citeassist}}
\end{center}

\begin{center}
\vspace{1em}
\begin{tikzpicture}
\draw[Primary, line width=0.6pt] (0,0) -- (\textwidth,0);
\end{tikzpicture}
\vspace{1.2em}
\end{center}

\begin{tcolorbox}[enhanced,
                 frame hidden,
                 boxrule=0pt,
                 borderline west={2pt}{0pt}{Primary},
                 colback=LightBg,
                 sharp corners,
                 breakable,
                 fonttitle=\sffamily\bfseries\large,
                 coltitle=Primary,
                 title=BibTeX Entry,
                 attach title to upper={\vspace{0.2em}\par},
                 left=12pt]
\begin{lstlisting}
@techreport{kaesberg2025,
  author={Kaesberg, Lars Benedikt and Wahle, Jan Philip and Ruas, Terry and Gipp, Bela},
  title={SPaRC: A Spatial Pathfinding Reasoning Challenge},
  year={2025},
  month={05}
}
\end{lstlisting}
\end{tcolorbox}

\vspace{0.8em}
\begin{tcolorbox}[enhanced,
                 frame hidden,
                 boxrule=0pt,
                 borderline west={2pt}{0pt}{Primary},
                 colback=LightBg,
                 sharp corners,
                 breakable,
                 fonttitle=\sffamily\bfseries\large,
                 coltitle=Primary,
                 title=Online Access,
                 attach title to upper={\vspace{0.2em}\par},
                 left=12pt]

\renewcommand{\arraystretch}{1.5}
\begin{tabular}{@{}p{0.25\textwidth}@{}p{0.75\textwidth}@{}}

\textbf{\sffamily CiteAssist} & 
\begin{minipage}[t]{0.72\textwidth}
\href{https://citeassist.uni-goettingen.de/preprint/1c30ada3-ead0-44a5-97f9-6a7dfe39ba22}{\color{Primary}https://citeassist.uni-goettingen.de/preprint/1c30ada3-ead0-44a5-97f9-6a7dfe39ba22}
\end{minipage}\\
\end{tabular}

\end{tcolorbox}

\vfill
\begin{tikzpicture}
\draw[Primary!40, line width=0.4pt] (0,0) -- (\textwidth,0);
\end{tikzpicture}
\begin{center}
\small\sffamily\textcolor{TextMuted}{Generated \today}
\end{center}

\end{document}